\documentclass{article}

\PassOptionsToPackage{numbers, compress}{natbib}
\usepackage[preprint]{neurips_2026}

\usepackage[utf8]{inputenc} 
\usepackage[T1]{fontenc}    
\usepackage{hyperref}       
\usepackage{url}            
\usepackage{booktabs}       
\usepackage{amsfonts}       
\usepackage{nicefrac}       
\usepackage{microtype}      
\usepackage{xcolor}         

\usepackage{microtype}
\usepackage{graphicx}
\usepackage{subfigure}
\usepackage{multirow}
\usepackage{makecell}
\usepackage{xurl}
\usepackage{appendix}
\usepackage{caption}
\usepackage{tabularx}
\usepackage{tcolorbox}
\usepackage{amsmath, amssymb, amsthm}
\usepackage{lipsum}
\usepackage{algorithm}
\usepackage{algorithmicx}
\usepackage{algpseudocode}
\tcbuselibrary{breakable} 
\tcbuselibrary{skins} 
\usepackage{colortbl}

\definecolor{myred}{RGB}{255, 102, 102}
\definecolor{mybest}{RGB}{255, 127, 14}
\definecolor{mysecond}{RGB}{31, 119, 180}

\usepackage{wrapfig}
\usepackage{geometry}
\newcommand{\best}[1]{\textcolor{mybest}{\textbf{#1}}}
\newcommand{\second}[1]{\textcolor{mysecond}{\underline{#1}}}

\usepackage{mathtools}

\usepackage{cleveref}
\crefname{figure}{Figure}{Figures}
\Crefname{figure}{Figure}{Figures}

\title{Rethinking Time Series Tokenization from a Frequency Perspective}

\author{
  Zelong Tian\textsuperscript{1,2} \quad Ming Jin\textsuperscript{1*} \quad Bo Du\textsuperscript{2} \quad  Shirui Pan\textsuperscript{1*}\\
  \textsuperscript{1} School of Information and Communication Technology,\\
  \textsuperscript{2} Griffith Business School\\
  Griffith University, Brisbane, Australia\\
  \href{mailto:zelongtian006@gmail.com}{\texttt{zelongtian006@gmail.com}},\ \ \href{mailto:ming.jin@griffith.edu.au,bo.du@griffith.edu.au,s.pan@griffith.edu.au}{\texttt{\{ming.jin, bo.du, s.pan\}@griffith.edu.au}}\\
}

\begin{document}

\maketitle

\begin{abstract}
Tokenization that partitions time series into subsequences has become a foundational paradigm in modern time series modeling. However, we find that current tokenization strategies inherently introduce spectral distortions, forcing the learned representations to diverge from the true underlying patterns such as trend, periodicity, and seasonality, thereby impairing model performance. Through theoretical analysis from a frequency perspective, we derive the boundary conditions under which such distortions occur. To overcome this distortion, we propose \textbf{TimeFT}, a nearly distortion-free frequency-based tokenizer for time series data. TimeFT obtains tokens through frequency-domain partitioning, frequency shifting, and Nyquist sampling. The method does not have additional learnable parameters, incurs negligible computational complexity, and can serve as a drop-in replacement for existing tokenizers. Our extensive experiments on diverse tasks such as forecasting, classification, and anomaly detection demonstrate that TimeFT consistently and significantly improves performance. The code has been made available at~\url{https://github.com/ME1682/TimeFT}.
\end{abstract}

\section{Introduction}
\label{sec:Introduction}

\begingroup
\renewcommand{\thefootnote}{}
\footnotetext{\textsuperscript{*}Corresponding authors.}
\endgroup

\begin{wrapfigure}{r}{0.5\textwidth}
    \centering
    \includegraphics[width=\linewidth]{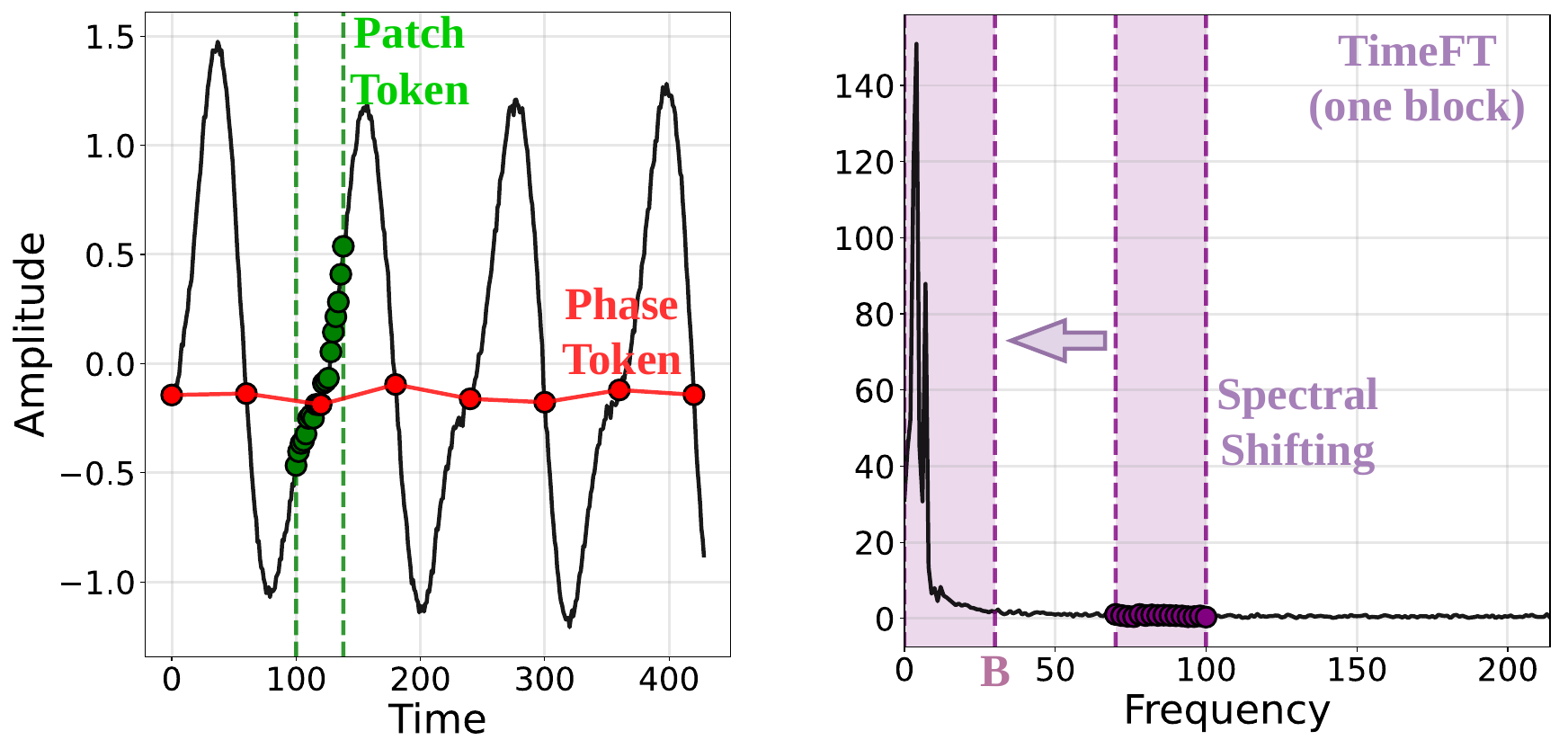}
    \vspace{-13pt}
    \caption{\textbf{The key distinction between existing tokenization and TimeFT.} \emph{Left:} The processes of patch and phase tokenization in the time domain. \emph{Right:} Spectral shifting moves the frequency spectrum, preventing aliasing by ensuring all components satisfy the Nyquist criterion.}
    \label{fig: figure1}
    \vspace{-50pt}
\end{wrapfigure}

Models~\cite{survey_ts} such as PatchTST~\cite{PatchTST}, PatchMLP~\cite{PatchMLP} and PhaseFormer~\cite{PhaseFormer} have established a tokenization paradigm~\cite{TimeCat, PatchCat} for time series data by partitioning the raw sequence into a set of non-overlapping (or overlapping) segments:

\begin{equation}
\begin{split}
\label{eq:patch token}
    \mathbf{Z}_{\mathrm{patch}\ i} :=& X\!\left[\, \sum_{j=1}^{i-1} S_j \,:\, \sum_{j=1}^{i-1} S_j + L_i \,\right]\\
    &i \in \{1,2, \dots, I\}
\end{split}
\end{equation}

\begin{equation}
\begin{split}
\label{eq:phase token}
    \mathbf{Z}_{\mathrm{phase}\ i} &:= \left[\, X[\;j\;] \;\text{ for }\; j = i, i+S, \dots \right]\\
    &j\leq N-1 , \ i \in \{0,1,2, \dots, I\}
\end{split}
\end{equation}

where $\mathbf{Z}_{\mathrm{patch}\ i}$ is the patch token~\cite{PatchTST} and $\mathbf{Z}_{\mathrm{phase}\ i}$ is the phase token~\cite{PhaseFormer}, which represent two main classes of tokenization methods. $X$ denotes the original input sequence of length $N$, $L_i$ is the length of each patch, $S_i$ or $S$ is the stride, and $I$ is the number of tokens. As shown in Figure~\ref{fig: figure1}, the patch tokenizer can be seen as segmenting the sequence in the time domain, and the phase tokenizer can be regarded as downsampling in the time domain. Extensive existing work shows that the tokenization paradigm has become a fundamental component of modern time series modeling~(\S\ref{sec:Related Work}). However, the frequency domain effects of tokenization remain largely unexplored.

\begin{figure*}[t]
    \centering
    \includegraphics[width=0.9\textwidth]{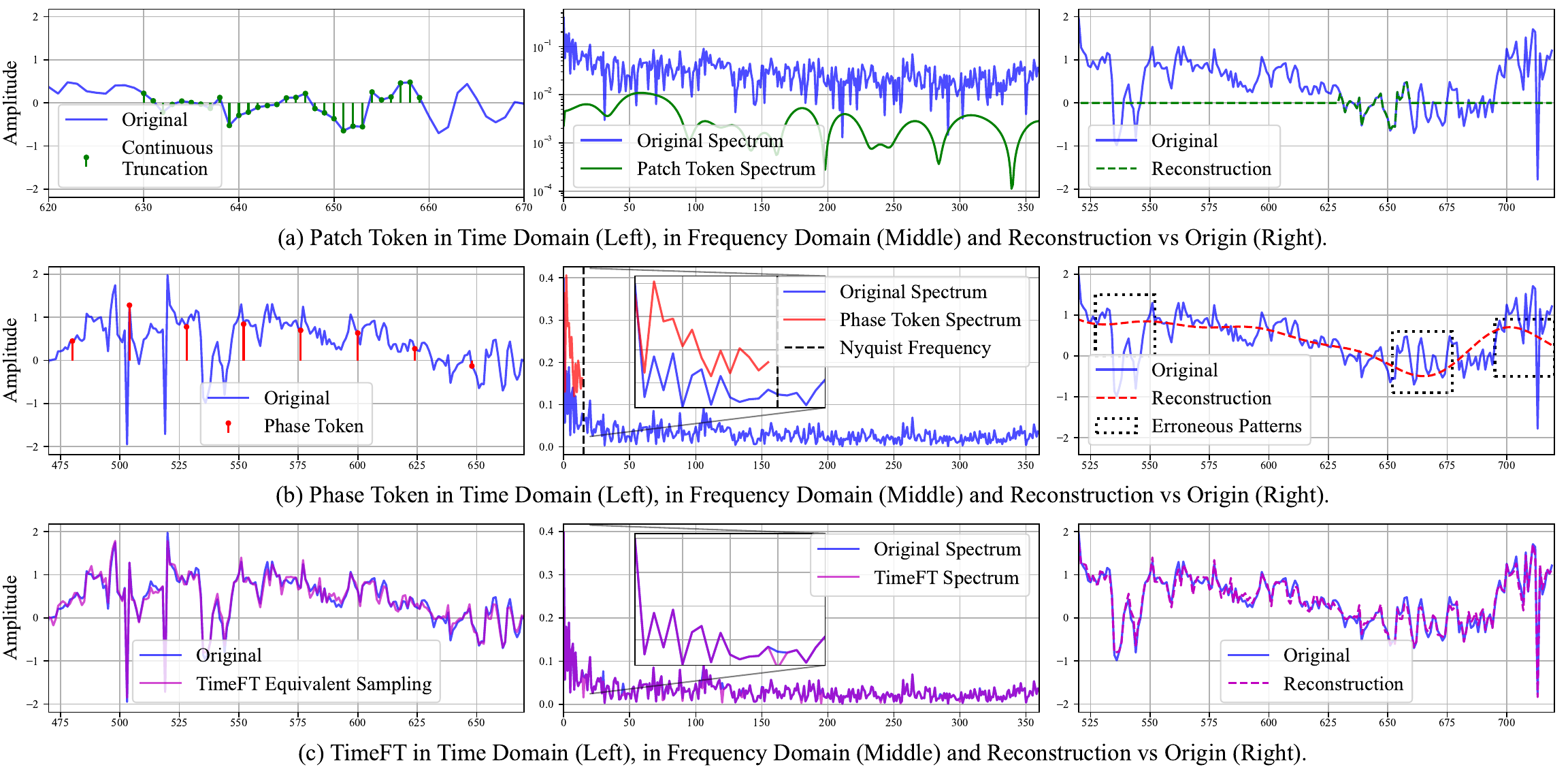} 
    \caption{\textbf{The comparison of tokenization methods across the time domain, frequency domain and reconstruction versus origin.} (a) Patch tokens lead to a completely different frequency spectrum and cause the loss of all information outside the rectangular window. (b) Phase tokens cause frequency components above the Nyquist frequency to completely alias into the low frequencies, thereby exhibiting erroneous patterns. (c) Our TimeFT effectively solves the above issues and achieves almost distortion‑free tokenization. The further explanation of Figure~\ref{fig: figure2} and the details of reconstruction methods are provided in the Appendix~\ref{app: Reconstruction_method}.}
    \label{fig: figure2}
    \vspace{-10pt}
\end{figure*}

For patch tokens, rectangular window truncation in the time domain leads to issues such as information loss and spectral leakage in the frequency domain~(\S\ref{app: patch_analysis}). For phase tokens, downsampling in the time domain leads to frequency aliasing in the frequency domain~(\S\ref{app: phase_analysis}). These distortions in tokenization cause the patterns of tokens, such as trend, periodicity, and seasonality, to diverge significantly from those of the original time series, thereby leading the model to learn erroneous patterns and degrading model performance. The fundamental reason for this phenomenon lies in distortions caused by tokenization methods and the constraints that the sequence length imposes on the frequency. Towards this, we propose \textbf{TimeFT}, a nearly distortion-free frequency-based tokenizer for time series data.

\begin{equation}
\label{eq:TimeFT}
    \mathbf{Z}_{\mathrm{TimeFT}} := \operatorname{Reshape}\Big( \operatorname{Nyquist\ Sampling}_{S}\Big( \mathcal{F}_{\mathbb{R}}^{-1}\big( \operatorname{Shift}\big( \operatorname{Block}_{K}\big( \mathcal{F}_{\mathbb{R}}(X) \big) \big) \big) \Big) \Big)
\end{equation}

where $\mathbf{Z}_{\mathrm{TimeFT}}$ is the set of TimeFT tokens, $S$ is the sampling stride, and $K$ is the number of blocks. $\operatorname{Block}_{K}(\cdot)$, $\operatorname{Shift}(\cdot)$, $\operatorname{Nyquist\ Sampling}_{S}(\cdot)$ and $\operatorname{Reshape}(\cdot)$ denote the operations~\ref{eq:band_extraction}, \ref{eq:shifting_truncation}, \ref{eq:D_s} and \ref{eq:token_reorganization} of TimeFT, which split the time series spectrum into multiple sub-bands that satisfy the alias-free condition, shift all of them to baseband, and then apply Nyquist sampling. $\mathcal{F}_{\mathbb{R}}$ and $\mathcal{F}_{\mathbb{R}}^{-1}$ denote the real-valued fast Fourier transform (RFFT) and its inverse transform. The comparison of three tokenization methods is illustrated in Figure~\ref{fig: figure2}.

As mentioned, TimeFT is a method without additional learnable parameters that can serve as a fundamental time series tokenization module for a wide range of models, and its time complexity is negligible~(\S\ref{subsec: time_complexity}). We prove that our method solves the above data distortion issue~(\S\ref{app: Analysis of TimeFT}). As a result, TimeFT can consistently and effectively improve model performance in various tasks~(\S\ref{subsec: main_experiments}).  More motivation explanations and theoretical insights are provided in \S\ref{app: Motivation and Insights}. The main contributions of our work are as follows.

\begin{itemize}
\item We find that the existing time series tokenization paradigm causes distortions in the frequency domain, which leads the data to exhibit erroneous temporal patterns. We then derive the transformation relationships in the time domain and the frequency domain during tokenization and further establish the boundary conditions of such distortions.

\item We propose TimeFT based on the above boundary conditions, which generates tokens while preserving frequency information by employing frequency-domain partitioning, spectrum shifting, and Nyquist sampling. Our method effectively solves the frequency distortion issue during tokenization.

\item Extensive experiments demonstrate that TimeFT achieves state-of-the-art performance compared to existing tokenization methods in forecasting, classification, and anomaly detection tasks. As a plug-in module, TimeFT results in an improvement of nearly 10\% in many cases, and the time cost is consistently less than 1.4\% of the training time per epoch.

\end{itemize}
\section{Preliminaries}
\label{sec: Preliminary}

In this section, we first define the tokenization problem. Then, from a frequency-domain perspective, we theoretically analyze the causes of the distortions shown in Figure~\ref{fig: figure2} for patch and phase tokens. Detailed theoretical analysis and illustrations are provided in the Appendix~\ref{app: Theory}.

\paragraph{Problem Definition.}

Consider a multivariate time series input denoted as $\mathbf{X} \in \mathbb{R}^{V \times N}$, where $V$ is the number of variates (channels) and $N$ is the length of the temporal sequence. The fundamental objective of our tokenization strategy is to discretize time series $\mathbf{X} \in \mathbb{R}^{V \times N}$ into a finite collection of tokens $\mathbf{Z} \in \mathbb{R}^{V \times S \times *}$ that can be learned by models, where $S$ and $*$ are the shapes of the tokens.

\paragraph{Patch Tokenization in Frequency Domain.}

Let $x[n]$, $n \in \{0, 1, \dots, N-1\}$, be a discrete sequence of finite length $N$. The patch tokenization in Equation \ref{eq:patch token} can be equivalent to the operation of rectangular windowing. The subsequence of length $L$ is obtained by extracting the segment in the rectangular window and re-indexing it from zero, i.e., $y[m] = x[n_0 + m]$ for $m = 0, \dots, L-1$. This is equivalent to convolving with a shifted Dirichlet kernel in the frequency domain.

Intuitively, it is easy to assume that patch tokens distort high frequency information~\cite{AaPE, Hi_WaveTST, Protohpe} due to reduced data length, but the actual situation is much more complex. Applying a rectangular window of length $L$ to a sequence of length $N$ induces three main distortions. First, the kernel's slowly decaying sidelobes produce \textbf{spectral leakage}, spreading energy from each frequency component across the entire spectrum and potentially obscuring weak signals. Second, the width of the mainlobe establishes an absolute \textbf{resolution limit}: components spaced closer than half this interval merge into a single broadened peak, irrespective of the original sequence length $N$. Moreover, the windowed sequence is further truncated to length $L$, so it \textbf{discards information} beyond length $L$ while inheriting identical leakage and resolution penalties.

\paragraph{Phase Tokenization in Frequency Domain.}

The phase tokenization in Equation \ref{eq:phase token} is equivalent to the operation of downsampling, i.e., $y[k] = x[kS]$ for $k = 0, 1, \dots, L-1$, where $L = \lceil N/S \rceil$. If the original frequency spectrum contains nonzero energy for frequencies higher than the bound $B$, the adjacent replicas will overlap when downsampling, causing the higher frequency components to fold into lower frequencies. This phenomenon is known as \textbf{spectral aliasing} and constitutes an irreversible distortion. To avoid aliasing, downsampling must satisfy the \textbf{Nyquist sampling theorem~\cite{DSP}}.

\begin{equation}
\label{eq:nyquist_physical1}
2\pi \frac{B}{f_s} \leq \frac{\pi}{S} \quad \Longrightarrow \quad B \leq \frac{f_s}{2S}
\end{equation}

where $f_s$ represents the sampling frequency of the original discrete time sequence. The downsampled signal will be free from aliasing if and only if the original continuous-time signal is bandlimited to less than or equal to the new \textbf{Nyquist frequency} $f_s/2S$. Sampling that satisfies this condition is referred to as \textbf{Nyquist sampling}.
\section{Methodology}
\label{sec: Methodology}

As shown in Figure~\ref{fig: main_timeft}, our TimeFT consists of three principal stages: (1) frequency-domain decomposition based on boundary conditions that extracts different sub-bands, (2) frequency shifting translates high-frequency components into the low-frequency band, and (3) information-preserving token extraction that transforms these representations into final tokens.

\subsection{Boundary Conditions for Distortion-Free Tokenization}
\label{subsec: Conditions for distortion-free tokenization}

Section~\ref{sec: Preliminary} shows that for any finite discrete sequence $\mathbf{X}$, there exists a simple method (such as Digital-to-Analog Conversion) that can reconstruct it into a continuous process $\mathcal{X}$ (an infinite sequence) over a period of time. $\mathbf{X}$ can be regarded as being obtained from $\mathcal{X}$. When the boundary conditions are satisfied, both the sampling and reconstruction processes are invertible and lossless, and in this case $\mathbf{X}$ is equivalent to $\mathcal{X}$. However, due to the finite length constraint on the information contained in $\mathbf{X}$, an incorrect $\mathcal{X}$ is often reconstructed, leading to tokens of erroneous patterns. We next present the boundary conditions for this conversion.

\paragraph{Boundary Condition for Patch Tokenization.}
\label{subsubsec: patch_boundary}

Patch tokenization multiplies $x[n]$ by a rectangular window of length $L$ and extracts the non-zero segment. The necessary and sufficient condition for the token to retain all the information required for an exact reconstruction of the original signal is that the entire valid sequence of $x[n]$ is contained within the windowed interval. For a finite-duration sequence of length $N$, this condition is satisfied if and only if the window covers all non-zero values. If $x[n]$ is periodic with period $T$, the condition becomes $L = mT$ ($m \in \mathbb{N}^+$), which means that any contiguous $L$ sample block then suffices for perfect periodic extension, completely eliminating information loss. However, these situations rarely occur for general time series.

\paragraph{Boundary Condition for Phase Tokenization.}
\label{subsubsec: phase_boundary}

Compared to patch tokenization, the boundary condition of phase tokenization is much simpler. Due to the Nyquist sampling theorem, the alias-free condition requires $x[n]$ to satisfy Equation~\ref{eq:nyquist_physical1}. Therefore, for any discrete $x[n]$, we define:

\begin{equation}
\label{eq:bound}
B_N = \lfloor \frac{N}{2S} \rfloor
\end{equation}

As long as the maximum frequency component of $x[n]$ is less than or equal to $B_N$, it satisfies the boundary condition for phase tokenization. In this case, the original sequence can be perfectly recovered from the tokens, making the tokenization free from information loss.

\subsection{Frequency Domain Decomposition Based on Boundary Conditions}
\label{subsec: Freq_decomposition}

\begin{figure*}[t]
    \centering
    \includegraphics[width=0.99\textwidth]{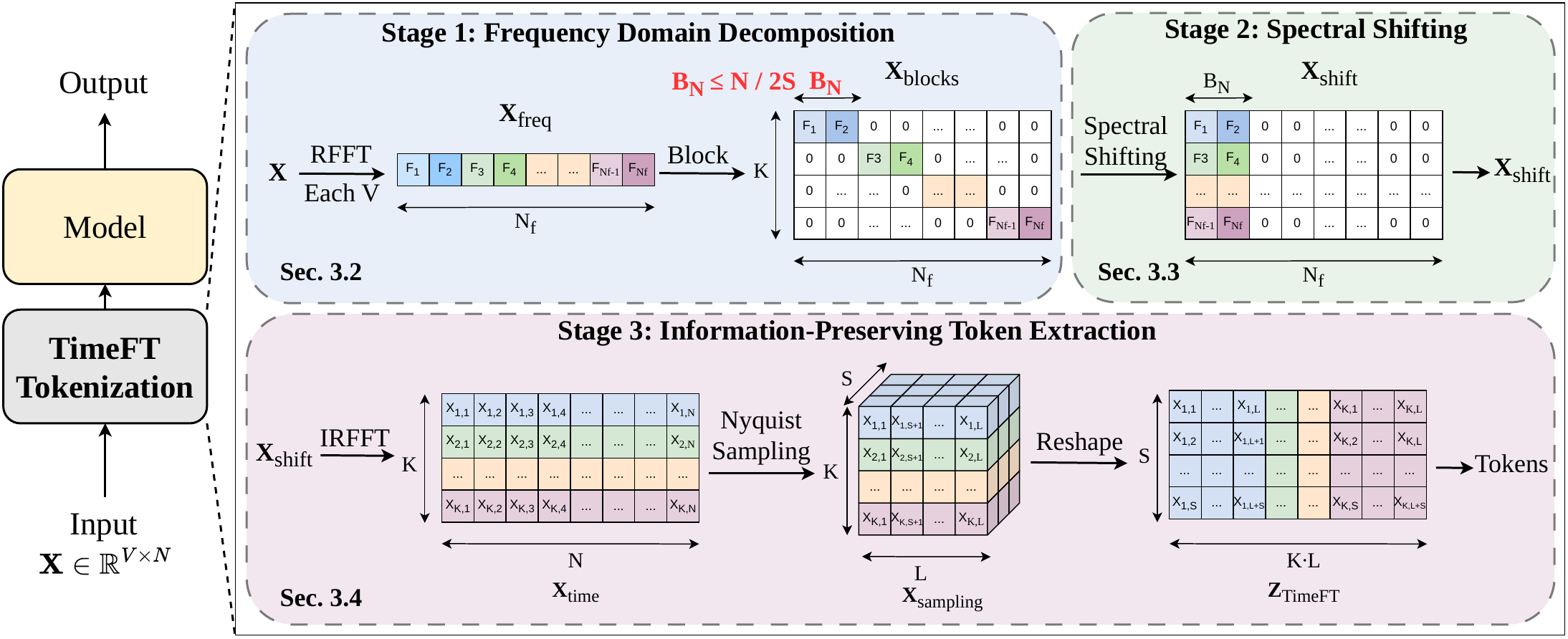} 
    \caption{\textbf{The pipeline of TimeFT} is implemented by (1) frequency domain decomposition, (2) spectral shifting and (3) information-preserving token extraction. Different colors and markers are used to denote the processes of operations. Based on the frequency threshold $B_N$ derived from downsampling without information loss, TimeFT shifts all spectra of bandwidth $B_N$ to the baseband, and then obtains almost distortion-free tokens by Nyquist sampling.}
    \label{fig: main_timeft}
    \vspace{-10pt}
\end{figure*}

In order to achieve lossless tokenization in the frequency domain, we first project the input time series into the frequency domain via the real-valued fast Fourier transform applied along the temporal axis, as shown in Stage 1 in Figure~\ref{fig: main_timeft}. Let $\mathcal{F}_{\mathbb{R}}$ denote the RFFT operator. The resulting complex-valued tensor is as follows:

\begin{equation}
\label{eq: rFFT}
\mathbf{X}_{\mathrm{freq}} = \mathcal{F}_{\mathbb{R}}(\mathbf{X}) \in \mathbb{C}^{V \times N_f}
\end{equation}

where $N_f = \lfloor N/2 \rfloor + 1$ is the number of frequency bins corresponding to the non-negative frequencies of a real signal of length $N$. We adopt a channel-independent approach for different variables to avoid frequency domain mixing. For a univariate real sequence $\mathbf{x} = (x_0, \dots, x_{N-1})$, its $k$-th RFFT coefficient is given by $[\mathcal{F}_{\mathbb{R}}(\mathbf{x})]_k = \sum_{n=0}^{N-1} x_n \cdot e^{-i 2\pi k n / N}, \  k = 0, 1, \dots, N_f-1$. To retain multi-frequency information, we partition the frequency representation along the spectral dimension into $K$ contiguous blocks, each of size $B_N$. This makes the spectral bandwidth satisfy Equation~\ref{eq:bound} and the boundary condition in~\S\ref{subsec: Conditions for distortion-free tokenization}. For the $k$-th band ($k = 0,\dots,K-1$), we define a full extraction that retains only the frequency components within the interval $[kB_N, kB_N+B_N-1]$ and zeros out the rest:

\begin{equation}
\label{eq:band_extraction}
\bigl[\mathbf{X}_{\mathrm{block}}^{(k)}\bigr]_{v, f} =
\begin{cases}
[\mathbf{X}_{\mathrm{freq}}]_{v, f}, & \text{if }\ kB_N \le f < (k+1)B_N, \\[4pt]
0, & \text{otherwise},
\end{cases}
\quad f = 0,\dots,N_f-1
\end{equation}

\begin{equation}
\label{eq:stacked_bands}
\mathbf{X}_{\mathrm{blocks}} = \bigl[ \mathbf{X}_{\mathrm{block}}^{(0)}; \mathbf{X}_{\mathrm{block}}^{(1)}; \dots; \mathbf{X}_{\mathrm{block}}^{(K-1)} \bigr] \in \mathbb{C}^{V \times K \times N_f}
\end{equation}

where $K = \lceil N_f / B_N \rceil$ denotes the total number of spectral blocks, and the last portion of $\mathbf{X}_{\mathrm{freq}}$ is filled by zero-padding. In this step, we separate the frequency components that correspond to characteristics such as trend, periodicity, and seasonality. Through this partitioning, we obtain a series of sequences in the frequency-domain that satisfy the required frequency band length under the boundary conditions, which is the first step in avoiding aliasing.

\subsection{Spectral Shifting}
\label{subsec: spectral_shifting}

However, existing effective frequency bands are located at high frequencies, causing high frequency components to irreversibly alias into low frequencies during downsampling. Therefore, we need to shift the high frequency information down to the low frequency range. As shown in Stage 2 in Figure~\ref{fig: main_timeft}, we align all bands to the baseband and multiply by a factor that compensates for the band starting frequency:

\begin{equation}
\label{eq:shifting_truncation}
\bigl[\mathbf{X}_{\mathrm{shift}} \bigr]_{v,k,b} = \bigl[\mathbf{X}_{\mathrm{blocks}}\bigr]_{v,k,\, kB_N + b} \cdot e^{-i 2\pi (k B_N) b / N},
\quad b = 0,1,\dots,B_N-1
\end{equation}

The frequency portion of $\mathbf{X}_{\mathrm{shift}}$ from $B_N$ to $N_f$ is zero-padded. Equivalently, one may view this as applying the shifting operator $\mathcal{S}$ to each band, which is $\mathbf{X}_{\mathrm{shift}} = \mathcal{S}\bigl(\mathbf{X}_{\mathrm{blocks}}\bigr) \in \mathbb{C}^{V \times K \times N_f}$. The resulting sequences $\mathbf{X}_{\mathrm{shift}}$ satisfy the Nyquist sampling condition when sampled with a step size of $S$, ensuring that no aliasing occurs. In Equation \ref{eq:shifting_truncation}, the factor is applied to align the phase reference with the time origin after shifting the frequency band. It compensates for the phase shift introduced by the translation without altering the magnitude of the complex signal in the frequency domain. Since TimeFT does not compress the temporal dimension, the phase factor in Equation \ref{eq:shifting_truncation} is ignored for models (\S\ref{app:linear_phase_rfft}). Therefore, Equations \ref{eq:band_extraction}-\ref{eq:shifting_truncation} can be simplified to the following form to avoid computing the above complex steps:

\begin{equation}
\label{eq: torch_shift}
\mathbf{X}_{\mathrm{shift}} = \operatorname{ZeroPad}_{B_N \to N_f}\Bigl( \operatorname{Reshape}\bigl(\mathbf{X}_{\mathrm{freq}}, (V, K, B_N)\bigr) \Bigr) \in \mathbb{C}^{V \times K \times N_f}
\end{equation}

where the $\operatorname{ZeroPad}_{B_N \to N_f}$ operation denotes appending zeros to the end of each frequency-domain sequence of length $B_N$ to extend its dimension to $N_f$. The $\operatorname{Reshape(\cdot)}$ operation zero-pads at the end if the dimension is insufficient. Here, the spectrum will slightly lose the imaginary part at zero frequency. In this way, we obtain a frequency-domain sequence that satisfies the Nyquist sampling theorem.

\subsection{Information-Preserving Token Extraction}
\label{subsec: token_extraction}

However, the tokenization extraction operation should be performed in the time domain. As shown in Stage 3 in Figure~\ref{fig: main_timeft}, to recover the corresponding time domain sequence, we apply the inverse real fast Fourier transform (IRFFT) $\mathcal{F}_{\mathbb{R}}^{-1}$ to every frequency block independently.

\begin{equation}
\label{eq:IRFFT}
\mathbf{X}_{\mathrm{time}} = \mathcal{F}_{\mathbb{R}}^{-1}\bigl(\mathbf{X}_{\mathrm{shift}}\bigr) \in \mathbb{R}^{ V \times K \times N}
\end{equation}

where $\mathbf{X}_{\mathrm{time}}$ denotes the time-domain representation of the sequences of $K$ different frequency bands of width $B_N$ for each variable $V$. Then we perform \emph{dense sampling}: for every possible offset $\lambda \in \{0,1,\dots,S-1\}$, we sample the signal starting at the index $\lambda$ with step $S$. Formally, for the $k$-th band of the variable $v$, let $\mathbf{x}_{v,k} = \mathbf{X}_{\mathrm{time}}[v,k,\cdot] \in \mathbb{R}^{N}$. The dense sampling operator $\mathcal{D}_S$ produces a matrix of size $S \times L$, where $L = \lceil N / S \rceil$:

\begin{equation}
\label{eq:D_s}
\bigl[\mathcal{D}_S(\mathbf{x}_{v,k})\bigr]_{\lambda, \ell} = \mathbf{x}_{v,k}[\lambda + \ell \cdot S],
\quad \lambda = 0,\dots,S-1,\;\; \ell = 0,\dots,L-1
\end{equation}

Applying this operator independently to every sub‑band yields the final token tensor, which is $\mathbf{X}_{\mathrm{sampling}} = \mathcal{D}_S\bigl(\mathbf{X}_{\mathrm{time}}\bigr) \in \mathbb{R}^{V \times K \times S \times L}$. Thus, each variate is represented by $K \times S$ vectors of length $L$, where $K$ indexes the frequency bands and $S$ captures different temporal phases. In our framework, before sampling, the $k$-th band has been shifted to the baseband and truncated to the $B$ frequency bins. Due to Equation \ref{eq:bound} the sampling stride $S$ satisfies $S \le  {N}/(2B_N)$. Hence, the operator $\mathcal{D}_S(\cdot)$ corresponds to the \textbf{Nyquist sampling} operation. Consequently, the dense sampling procedure preserves information and is lossless with respect to the sub‑band representation. Finally, to obtain a unified token, we first permute the dimension to group the frequency and sampling dimensions, then flatten the frequency bands into the feature dimension.

\begin{equation}
\label{eq:token_reorganization}
\mathbf{Z}_{\mathrm{TimeFT}} = \operatorname{reshape}\Bigl( \operatorname{permute}\bigl(\mathbf{X}_{\mathrm{sampling}}\bigr), \, (V, S, K \cdot L) \Bigr) \in \mathbb{R}^{V \times S \times (K \cdot L)}
\end{equation}

where $\operatorname{permute}(\cdot)$ reorders the dimension from $(V, K, S, L)$ to $(V, S, K, L)$, and the subsequent reshape collapses the last two dimensions into a single feature dimension of size $K \cdot L$. Since models typically apply embedding to tokens, the $K \cdot L$ dimension here does not lead to the curse of dimensionality. Each step of TimeFT is designed to be approximately invertible. The simplified spectral shifting only introduces a negligible loss at the zero‑frequency component, and we can almost perfectly reconstruct the original spectrum from each token (\S\ref{app:Reconstruction from TimeFT Tokens}).
\section{Experiments}
\label{sec: Experiments}

In this section, we conduct a comprehensive evaluation of TimeFT, including three major time series tasks~(\S\ref{subsec: main_experiments}), the computational efficiency~(\S\ref{subsec: time_complexity}), the plug-in experiment~(\S\ref{subsec: Plug_in_experiment}), the ablation study~(\S\ref{subsec: ablation_study}) and the hyperparameter sensitivity~(\S\ref{subsec: hyperparameter sensitivity analysis}). More details, experiments under noisy conditions and results on different random seeds are provided in \S~\ref{app: Experiments}, and visualizations of TimeFT are provided in \S~\ref{app: visualization}.

\subsection{Experimental Setup}
\label{subsec: Experimental_Setup}

\paragraph{Datasets.}

We evaluate TimeFT on a comprehensive set of benchmarks spanning three core time‑series tasks, and we select commonly used datasets for each task. \textbf{Long‑term forecasting} is performed on seven real‑world datasets: ETTh1, ETTh2, ETTm1, ETTm2, Weather, Electricity and Traffic~\cite{AutoFormer, TimesNet, iTransformer}. \textbf{Classification} uses 10 UEA datasets~\cite{EntroPE, Classification_survey, UEA}: EthanolConcentration, FaceDetection, Handwriting, Heartbeat, JapaneseVowels, PEMS‑SF, SelfRegulationSCP1, SelfRegulationSCP2, SpokenArabicDigits, and UWaveGestureLibrary. \textbf{Anomaly detection} is conducted on five widely used benchmarks~\cite{EntroPE, Anomaly_detection_survey, ADbench}: SMD, MSL, SMAP, SWaT, and PSM.

\paragraph{Baselines.}

We compare TimeFT with state-of-the-art tokenization methods and classic time series models. We consider EntroPE~\cite{EntroPE}, TimeMosaic~\cite{TimeMosaic}, PhaseFormer~\cite{PhaseFormer}, PatchMLP~\cite{PatchMLP} and PatchTST~\cite{PatchTST} as standard baselines. In addition, we also use TimeFilter~\cite{TimeFilter}, xPatch~\cite{xPatch}, iTransformer~\cite{iTransformer}, TimeMixer~\cite{TimeMixer}, TimesNet~\cite{TimesNet}, LightTS~\cite{LightTS}, DLinear~\cite{DLinear}, GPT4TS~\cite{GPT4TS}, ETSformer~\cite{ETSormer}, FEDformer~\cite{FEDormer} as baselines. The specific baselines differ slightly between tasks, due to the choice of baselines in previous papers. Since TimeFT is a general tokenization method rather than a model and due to expensive pre-train costs, we do not use time series foundation models (TSFMs) as baselines or plug-in models. \textbf{We use PatchTST as the backbone model for TimeFT in all experiments except the plug-in experiment.} 

\subsection{Main Results on Various Tasks}
\label{subsec: main_experiments}

For the long-term forecasting task, we evaluate the performance of TimeFT against state-of-the-art tokenization baselines across diverse datasets. The experiments adopt the settings in TimeMosaic~\cite{TimeMosaic}. As demonstrated in Table~\ref{tab:long_term_forecasting}, TimeFT achieves the best performance in the majority of datasets and metrics. By outperforming both classic and recent baselines, TimeFT establishes an excellent paradigm for obtaining tokens of time series. The detailed performance is provided in Table~\ref{tab:long_term_forecasting_all}.

\begin{table*}[t]
    \centering
    \caption{\textbf{For long-term forecasting task}, we evaluate average MSE and MAE across prediction horizons $H \in \{96, 192, 336, 720\}$. All baselines use a fixed input sequence length of 320. Best results are in \best{bold} and second best are \second{underlined}. The detailed performance is provided in Table~\ref{tab:long_term_forecasting_all}.}
    \label{tab:long_term_forecasting}
    
    \renewcommand{\arraystretch}{1.2}
    \small
    \setlength{\tabcolsep}{1.4pt}

    \resizebox{\textwidth}{!}{
    \begin{tabular}{l cc cc cc cc cc cc cc cc cc cc}
        \toprule
        Models & \multicolumn{2}{c}{\textbf{TimeFT}} & \multicolumn{2}{c}{EntroPE} & \multicolumn{2}{c}{TimeMosaic} & \multicolumn{2}{c}{PhaseFormer} & \multicolumn{2}{c}{TimeFilter} & \multicolumn{2}{c}{xPatch} & \multicolumn{2}{c}{PatchMLP} & \multicolumn{2}{c}{iTransformer} & \multicolumn{2}{c}{TimeMixer} & \multicolumn{2}{c}{PatchTST} \\
         & \multicolumn{2}{c}{(\textbf{Ours})} & \multicolumn{2}{c}{(2026)} & \multicolumn{2}{c}{(2026)} & \multicolumn{2}{c}{(2026)} & \multicolumn{2}{c}{(2025)} & \multicolumn{2}{c}{(2025)} & \multicolumn{2}{c}{(2025)} & \multicolumn{2}{c}{(2024)} & \multicolumn{2}{c}{(2024)} & \multicolumn{2}{c}{(2023)} \\
        \cmidrule(r){2-3} \cmidrule(lr){4-5} \cmidrule(lr){6-7} \cmidrule(lr){8-9} \cmidrule(lr){10-11} \cmidrule(lr){12-13} \cmidrule(lr){14-15} \cmidrule(lr){16-17} \cmidrule(lr){18-19} \cmidrule(l){20-21}
        Metric & MSE & MAE & MSE & MAE & MSE & MAE & MSE & MAE & MSE & MAE & MSE & MAE & MSE & MAE & MSE & MAE & MSE & MAE & MSE & MAE \\
        \midrule
        ETTm1 & \best{0.348} & \best{0.373} & 0.369 & 0.383 & 0.360 & \second{0.377} & 0.364 & 0.392 & 0.399 & 0.408 & \second{0.354} & 0.386 & 0.374 & 0.396 & 0.386 & 0.406 & 0.356 & 0.385 & 0.372 & 0.399 \\
        ETTm2 & \best{0.255} & \best{0.307} & 0.264 & 0.320 & \second{0.256} & \second{0.310} & 0.268 & 0.324 & 0.278 & 0.330 & 0.262 & 0.322 & 0.269 & 0.324 & 0.293 & 0.338 & 0.265 & 0.323 & 0.278 & 0.332 \\
        ETTh1 & \second{0.417} & \best{0.423} & 0.424 & \second{0.433} & \best{0.412} & \best{0.423} & 0.489 & 0.475 & 0.511 & 0.482 & 0.446 & 0.445 & 0.462 & 0.460 & 0.487 & 0.475 & 0.430 & 0.438 & 0.503 & 0.485 \\
        ETTh2 & \second{0.358} & \best{0.390} & 0.374 & 0.401 & 0.363 & \second{0.393} & 0.384 & 0.417 & 0.402 & 0.421 & \best{0.354} & 0.397 & 0.380 & 0.416 & 0.411 & 0.425 & \second{0.358} & 0.396 & 0.398 & 0.427 \\
        Weather & 0.232 & \second{0.261} & 0.237 & 0.264 & 0.228 & \best{0.256} & 0.233 & 0.268 & 0.231 & 0.270 & \best{0.224} & 0.262 & 0.230 & 0.267 & 0.237 & 0.272 & \second{0.227} & 0.266 & 0.236 & 0.271 \\
        Electricity & \best{0.165} & \best{0.258} & 0.177 & 0.271 & 0.170 & 0.263 & 0.177 & 0.273 & \second{0.167} & \second{0.260} & 0.170 & 0.264 & 0.200 & 0.302 & 0.180 & 0.277 & 0.173 & 0.267 & 0.173 & 0.272 \\
        Traffic & \best{0.420} & 0.290 & 0.444 & 0.300 & 0.434 & \second{0.287} & 0.441 & 0.305 & \best{0.420} & \best{0.284} & 0.430 & 0.300 & 0.525 & 0.382 & 0.463 & 0.335 & 0.444 & 0.316 & \second{0.428} & 0.297 \\
        \bottomrule
    \end{tabular}
    }
\end{table*}

\begin{table*}[t]
    \centering
    \caption{\textbf{For classification task}, we evaluate accuracy (\%) across all UEA datasets. Best results are in \best{bold} and second best are \second{underlined}. The detailed performance is provided in Table~\ref{tab:classification_all_uea}.}
    \label{tab:classification_results}
    \renewcommand{\arraystretch}{1.1}
    \small
    \setlength{\tabcolsep}{4.5pt}
    \resizebox{\textwidth}{!}{
    \begin{tabular}{l cccccccc c}
        \toprule
        Models & \textbf{TimeFT} & EntroPE & TimeMosaic & PhaseFormer & PatchMLP & GPT4TS & PatchTST & TimesNet & MLP \\
         & (\textbf{Ours}) & (2026) & (2026) & (2026) & (2025) & (2024) & (2023) & (2023) & - \\
        \midrule
        Average & \best{71.31} & \second{70.07} & 68.82 & 67.19 & 69.42 & 58.51 & 69.13 & 66.55 & 65.60 \\
        \bottomrule
    \end{tabular}
    }
\end{table*}

\begin{table*}[t]
    \centering
    \caption{\textbf{For anomaly detection task}, we evaluate F1-score (\%) on five benchmark datasets. Best results are in \best{bold} and second best are \second{underlined}. The detailed performance is provided in Table~\ref{tab:anomaly_detection_all}.}
    \label{tab:anomaly_detection}
    
    \renewcommand{\arraystretch}{1.1}
    \small
    \setlength{\tabcolsep}{5pt}

    \resizebox{\textwidth}{!}{
    \begin{tabular}{l cccccccccc}
        \toprule
        Models & \textbf{TimeFT} & EntroPE & TimeMosaic & PhaseFormer & LightTS & TimesNet & PatchTST & DLinear & ETSformer & FEDformer \\
        & (\textbf{Ours}) & (2026) & (2026) & (2026) & (2024) & (2023) & (2023) & (2023) & (2023) & (2022) \\
        \midrule
        Average & \best{86.12} & \second{85.38} & 82.50 & 83.01 & 84.23 & 85.24 & 82.79 & 82.46 & 82.87 & 84.97 \\
        \bottomrule
    \end{tabular}
    }
\end{table*}

For the classification task, we evaluate the performance on 10 commonly used UEA datasets. As shown in Tables~\ref{tab:classification_results} and \ref{tab:classification_all_uea}, TimeFT achieves the highest average accuracy and achieves the best or second-best result in a majority of datasets. This strong performance improvement is due to TimeFT's almost distortion‑free tokenization ability, which faithfully preserves discriminative temporal patterns that are often corrupted by tokenizers.

For the anomaly detection‌ task, we evaluate the capability on five standard datasets. TimeFT preserves the integrity of anomalous patterns and shows excellent detection performance. As shown in Tables~\ref{tab:anomaly_detection} and \ref{tab:anomaly_detection_all}, experiments on extensive datasets demonstrate that TimeFT exhibits superior performance and achieves the highest average F1‑score of 86.12\%.

\subsection{Time Complexity Analysis}
\label{subsec: time_complexity}

The computational complexity of the proposed TimeFT tokenization is $\mathcal{O}(V N \log N + V K N \log N + V K N)$. Each of the $V \times K$ frequency blocks is transformed back to the time domain via an IRFFT (Equation~\ref{eq:IRFFT}), contributing the dominant term of $\mathcal{O}(V K N \log N)$. As shown in Table~\ref{tab: time_complexity}, the one-time overhead introduced by TimeFT is negligible relative to the aggregate cost of model training and evaluation. Moreover, tokenization can be efficiently batched and parallelized, further amortizing its practical runtime. For multiple epochs, TimeFT produces identical tokens as long as the input sequence is the same, thus eliminating the need for repeated execution.

\begin{table*}[tbp]
    \centering
    \caption{\textbf{Comparison of TimeFT tokenization time and training time} per epoch across different datasets. The percentage represents the proportion of TimeFT tokenization time relative to the training time per epoch.}
    \label{tab: time_complexity}
    
    \renewcommand{\arraystretch}{1.1}
    \small
    \setlength{\tabcolsep}{7pt}
    
    \resizebox{0.8\textwidth}{!}{
    \begin{tabular}{l ccccccc}
        \toprule
        Datasets & ETTm1 & ETTm2 & ETTh1 & ETTh2 & Weather & Electricity & Traffic \\
        \midrule
        \rowcolor{blue!7}
        \textbf{TimeFT} (s)  & \textbf{1.492} & \textbf{1.376} & \textbf{0.352} & \textbf{0.362} & \textbf{3.799} & \textbf{28.50} & \textbf{50.27} \\
        Training (s)   & 299.5 & 177.5 & 80.55 & 127.8 & 833.7 & 2157  & 3629  \\
        \midrule
        Percentage  & 0.50\% & 0.78\% & 0.44\% & 0.28\% & 0.46\% & 1.32\% & 1.38\% \\
        \bottomrule
    \end{tabular}
    }
\end{table*}

\begin{table}[tbp]
    \centering
    \caption{\textbf{Plug-in performance} on long-term forecasting task. Evaluation of TimeFT as a plug-in tokenization module on prediction horizon $H=96$ and input length of 320. Experiments on more architectures are provided in Table~\ref{tab: Experiments on More Architectures}.}
    \label{tab:plug_in}
    
    \renewcommand{\arraystretch}{1.1}
    \small
    \setlength{\tabcolsep}{3pt}

    \resizebox{\textwidth}{!}{
    \begin{tabular}{l cc cc cc cc cc cc cc}
        \toprule
        Datasets & \multicolumn{2}{c}{ETTm1} & \multicolumn{2}{c}{ETTm2} & \multicolumn{2}{c}{ETTh1} & \multicolumn{2}{c}{ETTh2} & \multicolumn{2}{c}{Weather} & \multicolumn{2}{c}{Electricity} & \multicolumn{2}{c}{Traffic} \\
        \cmidrule(r){2-3} \cmidrule(lr){4-5} \cmidrule(lr){6-7} \cmidrule(lr){8-9} \cmidrule(lr){10-11} \cmidrule(lr){12-13} \cmidrule(l){14-15}
        & MSE & MAE & MSE & MAE & MSE & MAE & MSE & MAE & MSE & MAE & MSE & MAE & MSE & MAE \\
        \midrule
        PatchTST & 0.321 & 0.361 & 0.181 & 0.270 & 0.403 & 0.423 & 0.324 & 0.374 & 0.155 & 0.204 & 0.142 & 0.245 & 0.403 & 0.285 \\
        \rowcolor{blue!7}
        \textbf{\ \ \ \ +TimeFT} & \textbf{0.290} & \textbf{0.339} & \textbf{0.162} & \textbf{0.244} & \textbf{0.376} & \textbf{0.392} & \textbf{0.297} & \textbf{0.339} & \textbf{0.147} & \textbf{0.194} & \textbf{0.139} & \textbf{0.234} & \textbf{0.395} & \textbf{0.279} \\
        $\Delta\%$ Improvement & 9.7\% & 6.1\% & 10.5\% & 9.6\% & 6.7\% & 7.3\% & 8.3\% & 9.4\% & 5.2\% & 4.9\% & 2.1\% & 4.5\% & 2.0\% & 2.1\% \\
        \midrule
        PhaseFormer & 0.312 & 0.352 & 0.174 & 0.261 & 0.389 & 0.411 & 0.309 & 0.361 & 0.152 & 0.201 & 0.145 & 0.248 & 0.418 & 0.289 \\
        \rowcolor{blue!7}
        \textbf{\ \ \ \ +TimeFT} & \textbf{0.293} & \textbf{0.341} & \textbf{0.161} & \textbf{0.249} & \textbf{0.379} & \textbf{0.397} & \textbf{0.299} & \textbf{0.343} & \textbf{0.149} & \textbf{0.193} & \textbf{0.143} & \textbf{0.242} & \textbf{0.406} & \textbf{0.278} \\
        $\Delta\%$ Improvement & 6.1\% & 3.1\% & 7.5\% & 4.6\% & 2.6\% & 3.4\% & 3.2\% & 5.0\% & 1.9\% & 3.9\% & 1.4\% & 2.4\% & 2.9\% & 3.8\% \\
        \bottomrule
    \end{tabular}
    }
\end{table}

\begin{table}[tbp]
    \centering
    \caption{\textbf{Ablation study} on long-term forecasting task on prediction horizon $H=96$ and input length of 320. The “w/o” indicates that the operations of the corresponding stage in Figure~\ref{fig: main_timeft} are skipped in the pipeline, while the rest remains unchanged. Stage 1 is frequency domain decomposition, Stage 2 is spectral shifting and Stage 3 is information-preserving token extraction.}
    \label{tab:ablation}
    
    \renewcommand{\arraystretch}{1.1}
    \small
    \setlength{\tabcolsep}{3pt}

    \resizebox{\textwidth}{!}{
    \begin{tabular}{l cc cc cc cc cc cc cc}
        \toprule
        Datasets & \multicolumn{2}{c}{ETTm1} & \multicolumn{2}{c}{ETTm2} & \multicolumn{2}{c}{ETTh1} & \multicolumn{2}{c}{ETTh2} & \multicolumn{2}{c}{Weather} & \multicolumn{2}{c}{Electricity} & \multicolumn{2}{c}{Traffic} \\
        \cmidrule(r){2-3} \cmidrule(lr){4-5} \cmidrule(lr){6-7} \cmidrule(lr){8-9} \cmidrule(lr){10-11} \cmidrule(lr){12-13} \cmidrule(l){14-15}
        & MSE & MAE & MSE & MAE & MSE & MAE & MSE & MAE & MSE & MAE & MSE & MAE & MSE & MAE \\
        \midrule
        \rowcolor{blue!7}
        \textbf{TimeFT} & \textbf{0.290} & \textbf{0.339} & \textbf{0.162} & \textbf{0.244} & \textbf{0.376} & \textbf{0.392} & \textbf{0.297} & \textbf{0.339} & \textbf{0.147} & \textbf{0.194} & \textbf{0.139} & \textbf{0.234} & \textbf{0.395} & \textbf{0.279} \\
        \ \ \ \ w/o Stage 1 & 0.328 & 0.364 & 0.175 & 0.261 & 0.412 & 0.421 & 0.315 & 0.358 & 0.162 & 0.211 & 0.151 & 0.255 & 0.408 & 0.288 \\
        \ \ \ \ w/o Stage 2 & 0.314 & 0.352 & 0.194 & 0.282 & 0.398 & 0.415 & 0.342 & 0.384 & 0.158 & 0.207 & 0.146 & 0.249 & 0.423 & 0.301 \\
        \ \ \ \ w/o Stage 3 & 0.375 & 0.412 & 0.228 & 0.315 & 0.482 & 0.485 & 0.396 & 0.432 & 0.189 & 0.242 & 0.178 & 0.294 & 0.475 & 0.342 \\
        \bottomrule
    \end{tabular}
    }
\end{table}

\subsection{Plug-in Experiment}
\label{subsec: Plug_in_experiment}

We evaluate the universality of TimeFT using it as a plug-in tokenization module for PatchTST and PhaseFormer. As shown in Table~\ref{tab:plug_in}, integrating TimeFT consistently improves performance in all seven datasets. For PatchTST, performance in MSE and MAE improves from nearly $10\%$ in many cases. For PhaseFormer, the corresponding improvements are more than $5\%$ in many cases. These results demonstrate that TimeFT’s near‑lossless tokenization is readily transferable to different architectures, delivering immediate gains without modifying the base model design.

\subsection{Ablation Study}
\label{subsec: ablation_study}

We conduct the ablation study to demonstrate the contribution of each part in TimeFT. Removing Stage 1 means that only a single block exists, and the subsequent operations artificially introduce aliasing. Removing Stage 2 means that the high-frequency components of each block will individually alias into the low frequencies, and removing Stage 3 is equivalent to using different frequency components directly as tokens. As shown in Table~\ref{tab:ablation}, removing any individual stage significantly degrades the performance, showing that every component is essential to the model.

\subsection{Distortion Comparison of Different Tokenizations}
\label{subsec:Distortion Comparison of Different Tokenizations}

\begin{figure*}[tbp]
    \centering
    \begin{minipage}{0.47\textwidth}    
    \includegraphics[width=\textwidth]{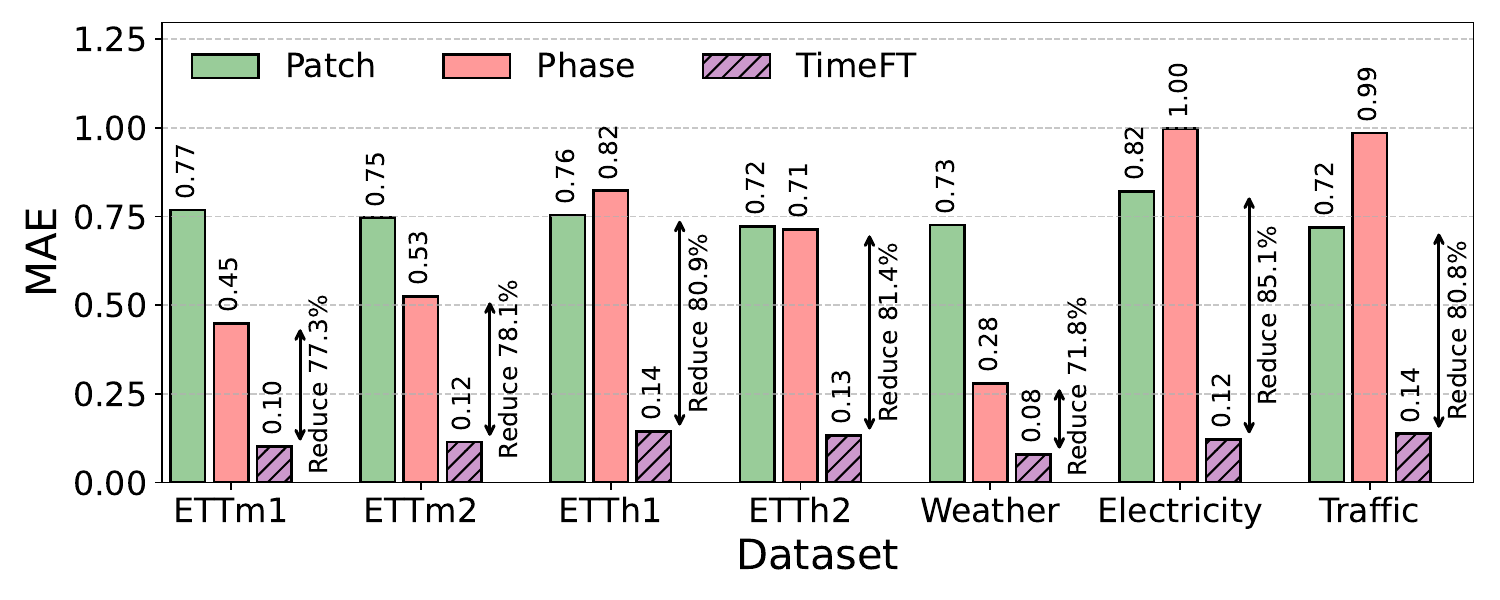} 
    \caption{\textbf{Reconstruction performance of different tokens types} across six commonly used datasets. The details are provided in Table~\ref{tab: Reconstruction Performance}.}
    \label{fig:Reconstruction}
    \end{minipage}
    \quad
    \begin{minipage}{0.49\textwidth}    
    \includegraphics[width=\textwidth]{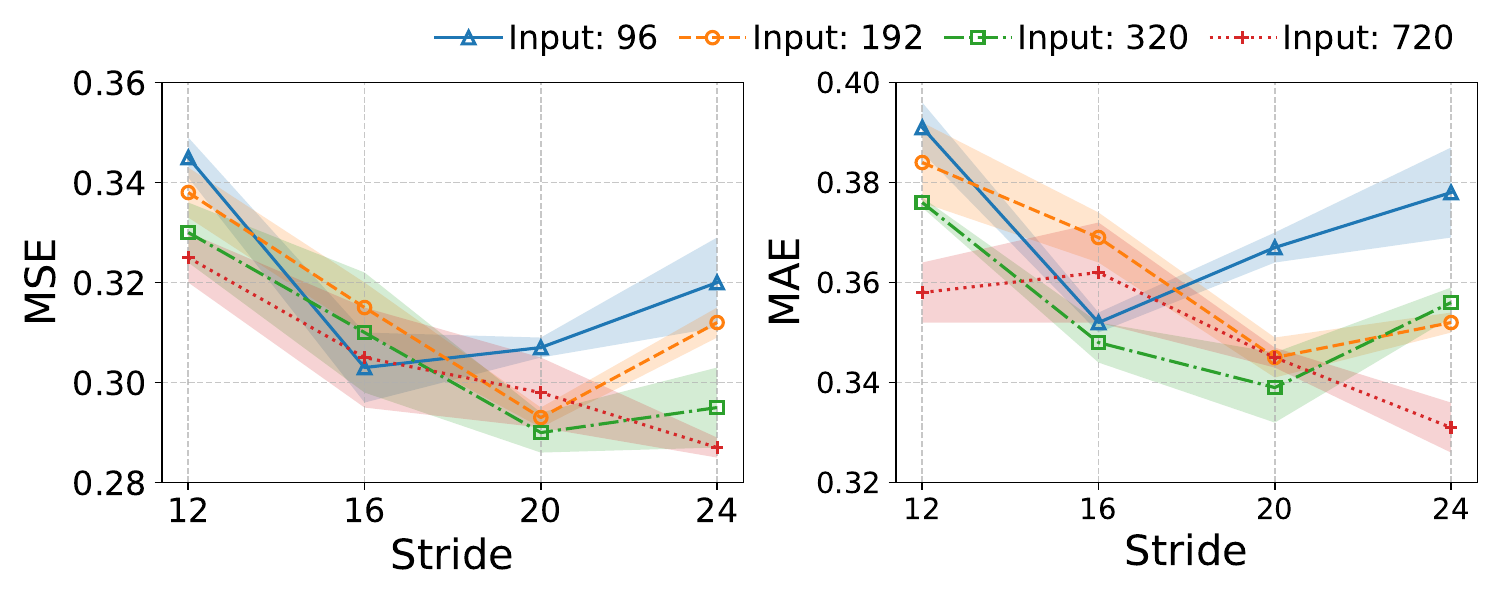} 
    \caption{\textbf{Hyperparameter sensitivity analysis} of tokenization stride $S$ for different input lengths $N$. Metric: \emph{Left:} MSE, \emph{Right:} MAE.}
    \label{fig:hyperparameter}
    \end{minipage}
    \vspace{-10pt}
\end{figure*}

To intuitively illustrate the degree of distortion for different types of tokens, we compare the mean absolute error between the reconstructed sequences and the original sequences across different datasets in the time domain. This experiment can be viewed as a statistical summary of Figure~\ref{fig: figure2} and the visualization section~\ref{app: visualization}. Due to Appendices~\ref{app: Theory} and \ref{app: Reconstruction_method}, the reconstructions from tokens to the reconstructed signals here are theoretically an invertible and lossless transformation. We ensure that all hyperparameters remain identical in this experiment. As shown in Figure~\ref{fig:Reconstruction}, TimeFT reduces more than 80\% reconstruction error in many datasets, which is shown by the percentage of ``Reduce'' on the right side of each bar.. This phenomenon arises because patch and phase tokenization introduce data distortion, causing the reconstructed sequence to differ substantially from the original sequence. In contrast, TimeFT achieves nearly distortion-free tokenization. The details are provided in Table~\ref{tab: Reconstruction Performance} and the reconstruction methods are provided in Appendix~\ref{app: Reconstruction_method}.

\subsection{Sensitivity Analysis}
\label{subsec: hyperparameter sensitivity analysis}

We further examine the impact of the tokenization stride S on model performance across various input lengths in the ETTm1 dataset. As shown in Figure~\ref{fig:hyperparameter}, the results reveal a significant synergy between the stride and the input scale, where the optimal configuration changes according to the input length. For shorter input sequences such as $N=96$, smaller strides ensure sufficient coverage of temporal details. However, as the input length increases to $N=720$, the model demonstrates a clear preference for larger strides. Since the stride determines the Nyquist frequency, a smaller stride results in a higher Nyquist frequency and longer tokens, while also leading to fewer tokens (downsampling). This trade‑off affects the model's performance, making it necessary to choose an appropriate stride for the input. Moreover, longer input sequences generally result in better performance.

\section{Related Work}
\label{sec:Related Work}

Existing time series tokenizers can be classified into two fundamental categories: patch tokens and phase tokens. We consider only the partitioning operation here without the embedding step. Quantization methods such as "bins"~\cite{Chronos1, Apollo} and some frequency methods~\cite{Fredn, FreqTST, DPWmixer, DUALformer} are not discussed because they are irrelevant to our topic~(\S\ref{app:Difference_Quantization}). According to the modeling paradigm, token‑based methods fall into the following three categories. 

For deep learning models, PatchTST\cite{PatchTST} introduces a patch-based approach to time series analysis, establishing patching as a key component of time series modeling. Subsequently, PITS~\cite{PITS}, PatchMLP~\cite{PatchMLP} and xPatch~\cite{xPatch} demonstrate that patch tokens can improve performance. Beyond static patching~\cite{PPT, PatchTSMixer, PathFormer}, some dynamic patching methods~\cite{TimeFilter, TimeSqueeze, APN, FlexTSF, HiPatch} have also been proposed for time series tokenization. SRSNet~\cite{SRSNet} leverages information from contextual time series to improve the forecasting performance of patch-based models. TimeMosaic~\cite{TimeMosaic} employs a lightweight classifier to determine the length of patches. Additionally, PhaseFormer~\cite{PhaseFormer} introduces time series phase tokens, which can be viewed as an approximate transpose of patch tokenization.

For pre-trained tokenizers, TOTEM~\cite{TOTEM} divides the time series into patches and maps them to a discrete embedding using a pre-trained model. EntroPE~\cite{EntroPE} employs a pre-trained entropy model to determine optimal patch boundaries. These methods still belong to the patch token class. 

For pre-trained foundation models, recent state-of-the-art TSFMs~\cite{TSFM_survey}, including Chronos-2~\cite{Chronos2}, Sundial~\cite{Sundial}, TiRex~\cite{TiRex}, Moirai‌~\cite{Moirai}, and TimeFM~\cite{TimeFM}, have adopted patch tokenization as a preprocessing step. Tokenization has become a fundamental component of TSFMs.
\section{Conclusion}
\label{sec: Conclusion}

We identify the inherent spectral distortion in the dominant time series tokenization paradigm and derive its boundary conditions for distortion-free tokenization. We propose TimeFT, an approach obtains time series tokens in a nearly distortion-free manner through frequency-domain partitioning, shifting, and Nyquist sampling. TimeFT is parameter-free, lightweight, and plug-and-play tokenizer for time series data. Extensive experiments on various tasks showcase that TimeFT outperforms existing practices and yields nearly 10\% improvement as a plug-in module in many cases.

\section{Acknowledge}
\label{sec: Acknowledge}

This work was supported by the NVIDIA Academic Grant in Higher Education and Developer program.

\bibliography{nips}
\bibliographystyle{Ref}

\newpage
\appendix

\section{Motivation Explanations and Theoretical Insights}
\label{app: Motivation and Insights}

\subsection{Theoretical Contribution‌ and Insights}

In this work, our theoretical contribution is the first systematic introduction of Digital Signal Processing~(DSP) tools to time series tokenization analysis, to address the inherent distortion problems in previous human intuition approaches such as patch and phase tokenization. The tokenization distortion phenomenon, the distortion boundary conditions, and TimeFT are all proposed for the first time. Specifically:

\begin{itemize}
\item For time-series tokenization, we establish the first frequency-based theoretical framework that precisely derives the frequency-domain effects of tokenization and reveals the root causes of distortions in existing tokenization methods.

\item Based on the above analysis, for the first time we provide the exact distortion-free boundary conditions for patch and phase tokenization, and accordingly design TimeFT to eliminate these distortions.
\end{itemize}

\subsection{Why Distortion in Individual Token Degrades Model Performance}

The distortion occurs during the tokenization step, before the tokens are fed into self-attention. Consequently, every individual token representation is already corrupted at the input level. While self-attention can mix information across tokens, it cannot recover the discarded information because that information was permanently lost prior to the attention layers. It is a matter of irreversible information loss at the representation.

This distortion is an inherent characteristic of tokenization and holds for any token. This leads to an irreversible alteration of the distribution of the input data, as shown in Appendix~\ref{app: visualization}, which causes deep learning models to receive corrupted information. Therefore, distortion in individual tokens degrades model performance.

\subsection{Why Distortion should be Addressed in Tokenization Stage}

While existing deep learning models for time series can capture complex dependencies thanks to the Universal Approximation Theorem, this does not imply that they are immune to input corruption. Distortions such as spectral leakage or aliasing occur during the tokenization process, i.e., before the data is input into the model. This discrepancy directly harms the model's generalization ability, as the learned patterns are tied to the distorted input rather than the underlying data relationship.

General deep learning models are not designed to correct systematic input distortions, and they are function approximators that fit the distribution they are trained on. If the training distribution is already biased, the model will simply converge to a biased solution. As shown in our experiments (Table~\ref{tab:plug_in}), even when the backbone architecture remains unchanged, replacing the tokenizer from patch/phase to TimeFT leads to significant performance improvements.

\section{Frequency Domain Theoretical Analysis and Boundary Conditions}
\label{app: Theory}
\subsection{Patch Tokenization}
\label{app: patch_analysis}

Let $x[n]$, $n \in \{0, 1, \dots, N-1\}$, be a discrete time sequence of finite length $N$. The patch tokenization in Equation \ref{eq:patch token} can be equivalent to the operation of rectangular windowing. A rectangular window $w_L[n; n_0]$ of length $L$ starting at the index $n_0$ is defined as $w_L[n; n_0] = 1$ for $n_0 \leq n \leq n_0 + L - 1$ and $0$ otherwise. The windowed sequence is $x_w[n] = x[n] \cdot w_L[n; n_0]$, and the sliced sequence of length $L$ is obtained by extracting the segment in the window and re-indexing it from zero, i.e., $y[m] = x_w[n_0 + m]$ for $m = 0, \dots, L-1$. We analyze the spectral consequences of these operations within the continuous frequency domain provided by the discrete time Fourier transform (DTFT). The DTFT of $x[n]$ is

\begin{equation}
X(e^{j\omega}) = \sum_{n=-\infty}^{\infty} x[n] e^{-j\omega n} = \sum_{n=0}^{N-1} x[n] e^{-j\omega n}
\end{equation}

where $\omega \in [-\pi, \pi)$ denotes the normalized angular frequency in radians. Note that $\operatorname{DTFT}$ of a sequence $x[n]$ of length $N$ is well-defined, since by treating the values at undefined $n$ as $0$, the sequence naturally extends to an infinite-length sequence, and the infinite summation in $\operatorname{DTFT}$ reduces to a finite sum. The DTFT of the rectangular window $w_L[n; n_0]$ is derived as follows:

\begin{equation}
\begin{aligned}
W_L(e^{j\omega}; n_0) &= \sum_{n=n_0}^{n_0+L-1} e^{-j\omega n} \\
&= e^{-j\omega n_0} \frac{1 - e^{-j\omega L}}{1 - e^{-j\omega}} \\
&= e^{-j\omega n_0} \cdot \frac{2j e^{-j\omega L/2} \sin(\omega L/2)}{2j e^{-j\omega/2} \sin(\omega/2)} \\
&= e^{-j\omega n_0} \, e^{-j\omega (L-1)/2} \, \frac{\sin(\omega L/2)}{\sin(\omega/2)} \\
&= e^{-j\omega \left(n_0 + \frac{L-1}{2}\right)} \cdot \frac{\sin(\omega L/2)}{\sin(\omega/2)}
\end{aligned}
\label{eq:DTFT_conv}
\end{equation}

where the first step expresses the windowed sum as a finite geometric series. Factor out the term corresponding to the starting index by letting \( n = n_0 + k,\ k = 0,1,\dots, L-1 \):

\begin{equation}
\sum_{k=0}^{L-1} e^{-j\omega (n_0 + k)}
   = e^{-j\omega n_0} \sum_{k=0}^{L-1} \left(e^{-j\omega}\right)^k
\end{equation}

The sum on the right is a finite geometric series with the first term \(1\), the common ratio \(r = e^{-j\omega}\), and \(L\) terms. Using the geometric sum formula \(\sum_{k=0}^{L-1} r^k = \frac{1 - r^L}{1 - r}\) (for \(r \neq 1\)), we obtain

\begin{equation}
\sum_{k=0}^{L-1} \left(e^{-j\omega}\right)^k
   = \frac{1 - \left(e^{-j\omega}\right)^L}{1 - e^{-j\omega}}
   = \frac{1 - e^{-j\omega L}}{1 - e^{-j\omega}}
\end{equation}

Substituting this back gives the closed form

\begin{equation}
W_L(e^{j\omega}; n_0) = e^{-j\omega n_0} \frac{1 - e^{-j\omega L}}{1 - e^{-j\omega}}
\end{equation}

In Equation~\ref{eq:DTFT_conv}, the factor $D_L(\omega) \triangleq \frac{\sin(\omega L/2)}{\sin(\omega/2)}$ is the Dirichlet kernel, a periodic sinc-like function whose main lobe has a null-to-null width of $4\pi/L$ and whose highest side lobe is approximately $-13$ dB below the main peak. The complex exponential factor $e^{-j\omega (n_0 + (L-1)/2)}$ imparts a linear phase shift, but since we are concerned with magnitude spectral characteristics, this phase term does not affect the analysis of problems. The windowed sequence $x_w[n] = x[n] \cdot w_L[n; n_0]$ is the pointwise product of the original sequence and the rectangular window. By the DTFT multiplication property, multiplication in the time domain corresponds to \emph{circular convolution} of their DTFTs over a period of length $2\pi$:

\begin{equation}
    X_w(e^{j\omega}) = \frac{1}{2\pi} \int_{-\pi}^{\pi} X(e^{j\theta}) \, W_L(e^{j(\omega-\theta)}; n_0) \, d\theta
    \label{eq:DTFT_W_L}
\end{equation}

where $X_w(e^{j\omega})$ is the DTFT of sequence $x_w[n]$. We then yield the following:

\begin{equation}
\begin{aligned}
    X_w(e^{j\omega}) 
    &= \frac{1}{2\pi} \int_{-\pi}^{\pi} X(e^{j\theta}) \,
       e^{-j(\omega-\theta) \left(n_0 + \frac{L-1}{2}\right)} 
       \frac{\sin\big((\omega-\theta)L/2\big)}{\sin\big((\omega-\theta)/2\big)} \, d\theta \\
    &= e^{-j\omega \left(n_0 + \frac{L-1}{2}\right)} 
       \cdot \frac{1}{2\pi} \int_{-\pi}^{\pi} 
       \Big[ X(e^{j\theta}) \, e^{j\theta \left(n_0 + \frac{L-1}{2}\right)} \Big] \, D_L(\omega-\theta) \, d\theta
\end{aligned}
\label{eq:convolution_expanded}
\end{equation}

where $D_L(\omega-\theta)=\frac{\sin((\omega-\theta)L/2)}{\sin((\omega-\theta)/2)}$ is the Dirichlet kernel shifted to $\theta$. Equation~\ref{eq:convolution_expanded} reveals that, up to a global linear phase shift, $X_w(e^{j\omega})$ is the convolution of a phase‑rotated version of $X(e^{j\omega})$ with the Dirichlet kernel $D_L(\cdot)$. The distortion mainly takes the following three forms: (1) \textbf{spectral leakage:} Slowly decaying sidelobes of $D_L(\cdot)$ spread the energy from each frequency component across the entire spectrum, potentially masking weaker components. (2) \textbf{resolution limit:} The main lobe of $D_L(\cdot)$ has a width of $4\pi/L$, which limits the ability to distinguish adjacent spectral peaks. Components closer than half this interval merge into a single broadened lobe, irrespective of the original sequence length $N$. Moreover, the windowed sequence is further truncated to length L, so it \textbf{discards information} beyond length L while inheriting the identical leakage and resolution penalties. Since discarding information outside the window is intuitive in the time domain, we next analyze the underlying causes of spectral leakage and resolution limit.

\paragraph{Spectral Leakage.} For $X_w(e^{j\omega})$ in Equation~\ref{eq:convolution_expanded}, the linear phase factor $e^{-j\omega (n_0 + (L-1)/2)}$ does not affect the magnitude spectrum, so the magnitude distortion is entirely determined by the convolution with the Dirichlet kernel. This convolution is the origin of \emph{spectral leakage}: each frequency component of the original signal is not confined to its ideal location but spreads to neighboring frequencies due to the side lobes of $D_L(\omega)$, as shown in Figure~\ref{fig:Appendix_spectral_leakage}:

\begin{figure*}[htbp]
    \centering
    \includegraphics[width=0.85\textwidth]{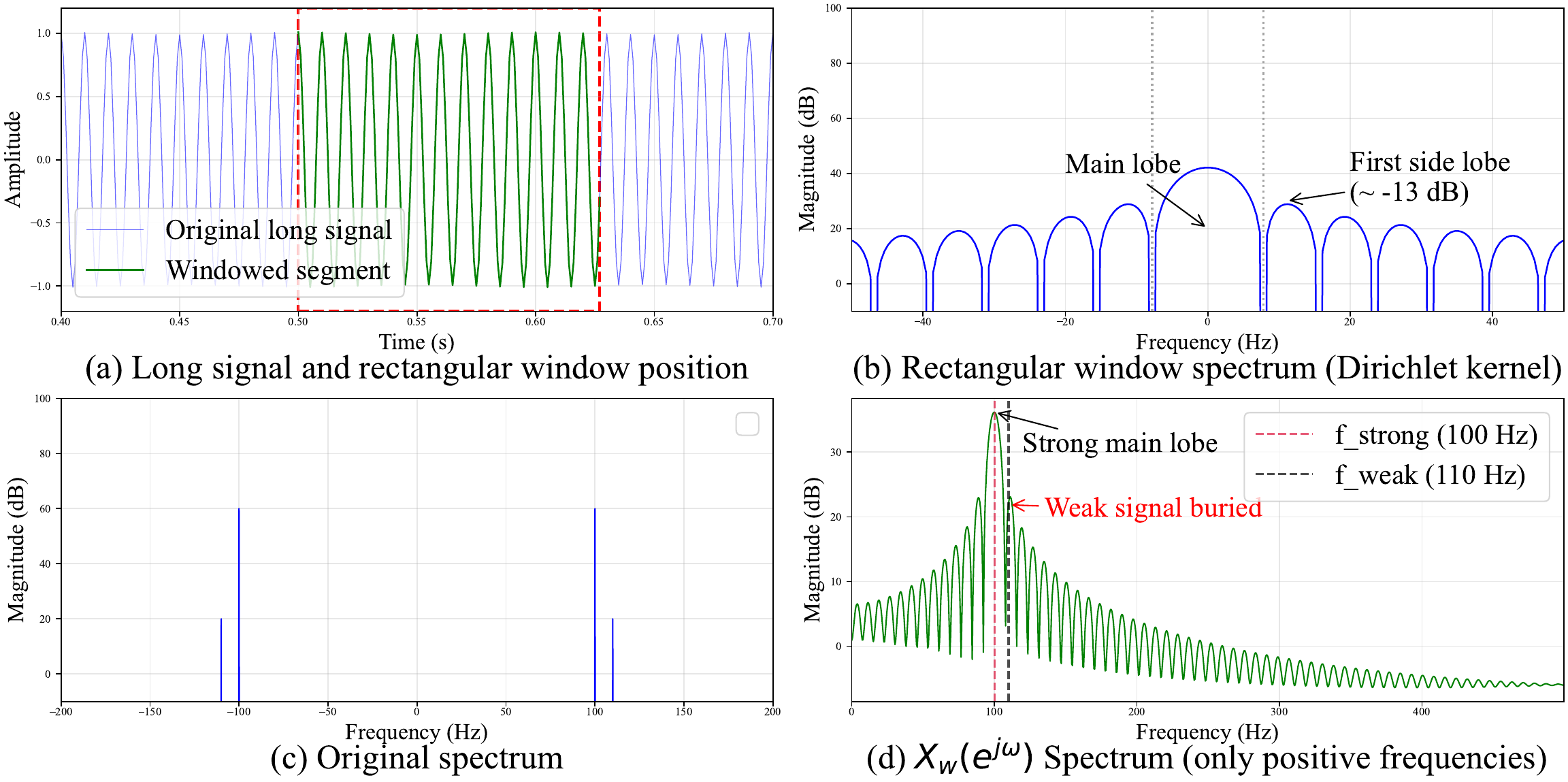} 
    \caption{Illustration of spectral leakage and weak-signal masking caused by rectangular window truncation. (a) Rectangular window truncation of a time-domain signal. (b) Spectral characteristics of the Dirichlet kernel corresponding to the rectangular window. (c) Original spectrum containing adjacent strong and weak signals. (d) Masking of weak signals caused by sidelobe leakage after windowing. The decibel (dB) is a logarithmic unit, so a small difference can represent a large change.}
    \label{fig:Appendix_spectral_leakage}
\end{figure*}

To illustrate the effect concretely with a typical signal from practical time series, consider a real sinusoid infinite-duration sequence $x[n] = \cos(\omega_0 n)$. Using Euler's formula, $x[n] = \frac{1}{2}e^{j\omega_0 n} + \frac{1}{2}e^{-j\omega_0 n}$, whose DTFT consists of two impulses at $\pm\omega_0$. Since the DTFT of the window sequence $x_w[n]$ is the convolution of $X(e^{j\omega})$ with $W_L(e^{j\omega}; n_0)$, linearity allows us to treat each complex exponential. Substituting the rectangular window DTFT from Equation~\ref{eq:DTFT_W_L} yields the following:

\begin{equation}
\begin{aligned}
X_w(e^{j\omega}) 
&= \frac{1}{2} W_L(e^{j(\omega-\omega_0)}; n_0) + \frac{1}{2} W_L(e^{j(\omega+\omega_0)}; n_0) \\[4pt]
&= \frac{1}{2} e^{-j(\omega-\omega_0)\left(n_0 + \frac{L-1}{2}\right)} D_L(\omega-\omega_0) + \frac{1}{2} e^{-j(\omega+\omega_0)\left(n_0 + \frac{L-1}{2}\right)} D_L(\omega+\omega_0)
\end{aligned}
\end{equation}

When $\omega_0$ is sufficiently far from $0$ and $\pi$, the main lobes centered at $\pm\omega_0$ are well separated, and the cross terms between them become negligible. In this case, the magnitude spectrum is approximately:

\begin{equation}
|X_w(e^{j\omega})| \approx \frac{1}{2}\big|D_L(\omega-\omega_0)\big| + \frac{1}{2}\big|D_L(\omega+\omega_0)\big|
\end{equation}

which explicitly shows that each ideal spectral line is replaced by a broadened main lobe accompanied by oscillatory side lobes. The factor $1/2$ accounts for the splitting of the sinusoid's energy between positive and negative frequencies. The side lobes of $|D_L(\cdot)|$ slowly decay, with the highest side lobe only about $13$~dB below the main peak. Consequently, a weak sinusoidal component located near a strong one may be completely masked by the side‑lobe energy of the stronger component.

\paragraph{Resolution Limit.} To understand the resolution limit imposed by the rectangular window, consider two complex exponentials $e^{j\omega_1 n}$ and $e^{j\omega_2 n}$ with $\omega_1 < \omega_2$. After windowing with $w_L[n;n_0]$, the DTFT of each component becomes a shifted Dirichlet kernel $D_L(\omega-\omega_i)$ (up to a phase factor). The magnitude spectrum therefore contains two main lobes centered at $\omega_1$ and $\omega_2$. According to the \textbf{Rayleigh criterion} commonly adopted in spectral analysis, two peaks are resolved when their separation is at least half the null-to-null width of the main lobe. Since the main lobe of $D_L(\omega)$ has a null-to-null width of $4\pi/L$, the critical separation is $2\pi/L$. Hence, two sinusoidal components can be distinguished in the DTFT magnitude spectrum only if

\begin{equation}
    |\omega_1 - \omega_2| \ge \frac{2\pi}{L}
    \label{eq:Rayleigh_criterion}
\end{equation}

If the separation is smaller, the two main lobes merge into a single broadened peak, and the components become indistinguishable regardless of the continuous nature of the DTFT. This is a trade-off relationship, reducing the window length $L$ improves temporal localization but inevitably degrades frequency resolution and exacerbates spectral leakage.

First, we analyze whether the transformation from $x_w[n]$ to $y[n]$ introduces the resolution limit problem. The patch tokenization extracts the windowed segment and re-indexes it to start at zero, yielding $y[m] = x_w[n_0+m]$ for $m=0,1,\dots,L-1$. We now derive its DTFT:

\begin{align}
    Y(e^{j\omega}) 
    &= \sum_{m=0}^{L-1} y[m]\, e^{-j\omega m} \nonumber \\
    &= \sum_{n=n_0}^{n_0+L-1} x_w[n]\, e^{-j\omega (n-n_0)}\\
    &= e^{j\omega n_0} \sum_{n=n_0}^{n_0+L-1} x_w[n]\, e^{-j\omega n} \nonumber \\
    &= e^{j\omega n_0} \, X_w(e^{j\omega})
    \label{eq:slice_dtft}
\end{align}

Equation~\ref{eq:slice_dtft} shows that slicing multiplies the DTFT of the windowed sequence by $e^{j\omega n_0}$, which corresponds to a time advance of $n_0$ samples. This shows that cutting merely applies a pure time advance, leaving the magnitude spectrum unchanged. Substituting the expression for $X_w(e^{j\omega})$ from Equation~\ref{eq:convolution_expanded}:

\begin{equation}
    X_w(e^{j\omega}) = e^{-j\omega\left(n_0+\frac{L-1}{2}\right)}
    \frac{1}{2\pi}\int_{-\pi}^{\pi} X(e^{j\theta})\, e^{j\theta\left(n_0+\frac{L-1}{2}\right)} D_L(\omega-\theta)\, d\theta
\end{equation}

We obtain

\begin{align}
    Y(e^{j\omega}) 
    &= e^{j\omega n_0} \, e^{-j\omega\left(n_0+\frac{L-1}{2}\right)}
    \frac{1}{2\pi}\int_{-\pi}^{\pi} X(e^{j\theta})\, e^{j\theta\left(n_0+\frac{L-1}{2}\right)} D_L(\omega-\theta)\, d\theta \nonumber \\
    &= \frac{1}{2\pi}\int_{-\pi}^{\pi} X(e^{j\theta})\, e^{j\theta n_0}\,
       e^{-j(\omega-\theta)\frac{L-1}{2}} D_L(\omega-\theta)\, d\theta
    \label{eq:slice_conv}
\end{align}

From Equation~\ref{eq:slice_dtft} it immediately follows that the magnitude spectra are identical:

\begin{equation}
    |Y(e^{j\omega})| = |X_w(e^{j\omega})|
\end{equation}

Thus, the step $y[m] = x_w[n_0 + m]$ for $m = 0, \dots, L-1$ does not alter the leak pattern or the frequency resolution; it simply realigns the temporal origin. All subsequent discussion of spectral leakage and resolution applies equally to the sliced sequence $y[m]$ that constitutes the patch token.

\begin{figure*}[htbp]
    \centering
    \includegraphics[width=0.85\textwidth]{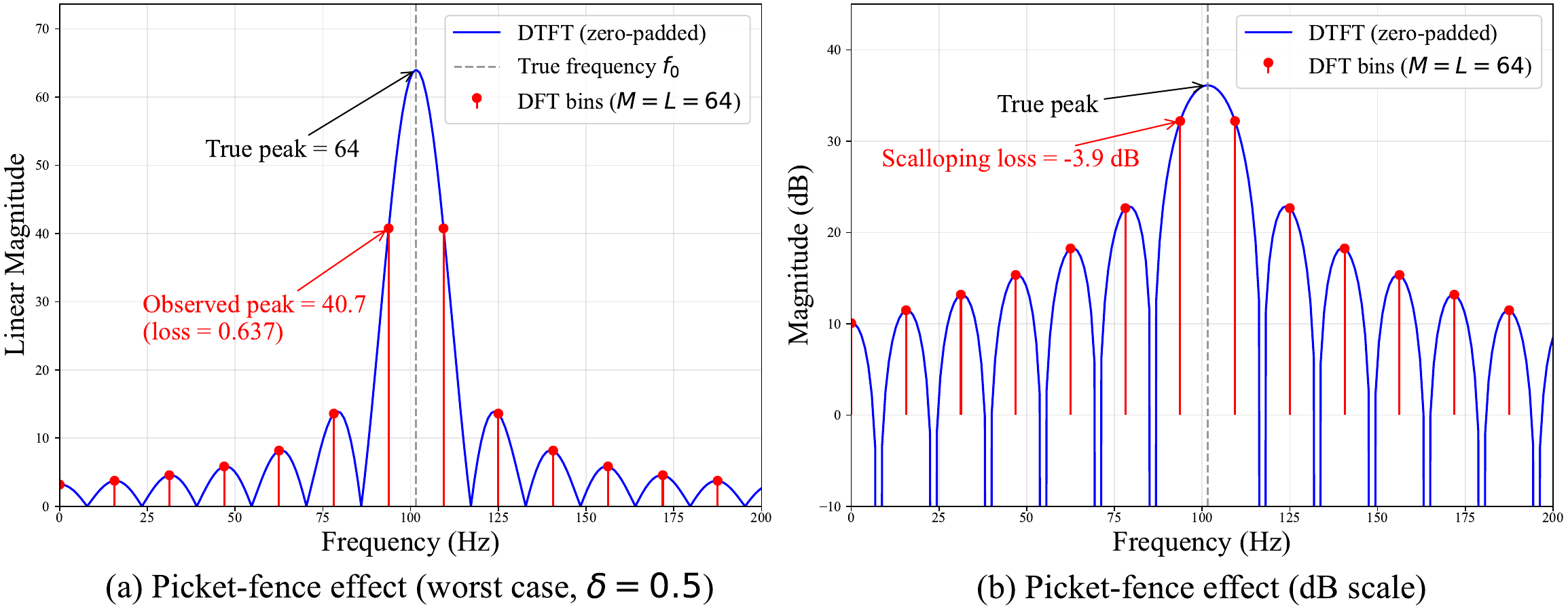} 
    \caption{Illustration of the picket‑fence effect. (a) Picket-fence effect at linear magnitude. (b) Picket-fence effect at dB magnitude.}
    \label{fig:Appendix_resolution_limit}
\end{figure*}

Although DTFT $Y(e^{j\omega})$ is a continuous function of $\omega$, any practical spectral analysis must evaluate it at a finite set of discrete frequencies. When we sample the DTFT at $M$ uniformly spaced frequencies $\omega_k = 2\pi k / M$, $k = 0, \dots, M-1$, we obtain a discrete representation that is subject to the \textbf{picket-fence effect}. When viewing a landscape through a picket fence, one can see only the portions of the scene that align with the gaps between the slats, while the rest remains hidden. In the spectral context, the continuous DTFT is the ``landscape,'' and the $M$ frequency samples $\omega_k$ are the ``gaps.'' If a signal component's true frequency lies exactly on one of these sample points, its full amplitude is observed; if it falls between two points, the observed value at the nearest sample is only an attenuated version of the actual spectral peak. The severity of this attenuation depends on how far the true frequency deviates from the nearest sampling point.

To quantify this effect, consider a single complex sinusoid of frequency $\omega_0$ after rectangular windowing. Its continuous DTFT is a shifted Dirichlet kernel, i.e., $Y(e^{j\omega}) = e^{j\phi(\omega)} D_L(\omega - \omega_0)$, where $\phi(\omega)$ is a linear phase term. The true magnitude peak occurs exactly at $\omega = \omega_0$ and is equal to $|D_L(0)| = L$. Suppose that the frequency $\omega_0$ falls between two discrete frequency grid points, written as $\omega_0 = 2\pi (k_0 + \delta)/M$, where $k_0$ is the index of the nearest grid point and $0 < |\delta| < 0.5$ measures the fractional bin offset. The discrete spectrum is evaluated only on the fixed grid $\omega_k = 2\pi k/M$. Hence, the maximum amplitude that we can observe among the discrete bins is

\begin{equation}
    |Y(e^{j\omega_{k_0}})| = |D_L(2\pi\delta/M)| = \left| \frac{\sin(\pi \delta L / M)}{\sin(\pi \delta / M)} \right|
\end{equation}

which is strictly smaller than $L$ whenever $\delta \neq 0$. The ratio of the observed peak to the true peak,

\begin{equation}
    \frac{|Y(e^{j\omega_{k_0}})|}{|D_L(0)|} = \frac{1}{L} \left| \frac{\sin(\pi \delta L / M)}{\sin(\pi \delta / M)} \right|
\end{equation}

This is known as the scalloping loss or picket-fence loss. For example, as shown in Figure~\ref{fig:Appendix_resolution_limit}, in the worst-case offset $\delta = 0.5$ with $M = L$ (no zero-padding), this ratio is about $|\sin(\pi/2)/(\pi/2)| \approx 0.637$, implying an amplitude reduction of more than $36\%$ and nearly 4 dB compared to the true peak. The effect can be mitigated by increasing $M$, the number of frequency samples over the same interval, which is usually achieved by zero padding the time‑domain sequence $y[m]$ before computing the DFT. Zero‑padding appends zeros to $y[m]$, resulting in a denser set of DTFT evaluation points $\omega_k$ without altering the underlying continuous spectrum $Y(e^{j\omega})$.

\subsection{Phase Tokenization}
\label{app: phase_analysis}

Let $x[n]$ be a finite‑length discrete‑time sequence of length $N$, and let $S\ge 2$ be an integer downsampling factor. The downsampled sequence is $y[m] = x[mS], \ m = 0,1,\dots,\lceil N/S\rceil-1$. We decompose this operation into two stages: (i) zero insertion, and (ii) time compression. To analyze the spectral consequences, it is convenient to introduce an auxiliary zero‑filled sequence:

\begin{equation}
    w[n] = \begin{cases}
        x[n], & n = kS,\; k\in\mathbb{Z},\\
        0,    & \text{otherwise}.
    \end{cases}
    \label{eq:w_def}
\end{equation}

which is simply $x[n]$ sampled on every $S$‑th index and padded with zeros elsewhere. The relation between $w[n]$ and the final downsampled signal $y[m]$ is

\begin{equation}
    y[m] = w[mS],\  m\in\mathbb{Z}
    \label{eq:y_from_w}
\end{equation}

The sequence $w[n]$ can be expressed as the product of $x[n]$ with a periodic impulse sequence

\begin{equation}
    p[n] = \sum_{k=-\infty}^{\infty} \delta[n - kS]
\end{equation}

where $\delta[n]$ denotes the unit sample ($\delta[0]=1$ and $\delta[n]=0$ for $n\neq0$). That is,

\begin{equation}
    w[n] = x[n]\,p[n]
\end{equation}

The DTFT of the impulse sequence $p[n]$ is obtained via the Poisson summation formula. The sequence $p[n] = \sum_{k=-\infty}^{\infty} \delta[n-kS]$ is periodic with period $S$. Within one period $n=0,1,\dots,S-1$, we have $p[0]=1$ and $p[n]=0$ for $n\neq0$. Its discrete Fourier series coefficients are therefore

\begin{equation}
    P_c[k] = \frac{1}{S}\sum_{n=0}^{S-1} p[n]\,e^{-j2\pi k n/S}
         = \frac{1}{S}, \ k=0,1,\dots,S-1
\end{equation}

Therefore,

\begin{equation}
    p[n] = \sum_{k=0}^{S-1} P_c[k]\,e^{j2\pi k n/S}
         = \frac{1}{S}\sum_{k=0}^{S-1} e^{j2\pi k n/S}
\end{equation}

Taking the DTFT of both sides,

\begin{align}
    P(e^{j\omega}) &= \sum_{n=-\infty}^{\infty} p[n]\,e^{-j\omega n}
    = \frac{1}{S}\sum_{k=0}^{S-1}\; \sum_{n=-\infty}^{\infty} e^{-j(\omega-2\pi k/S)n} \nonumber \\
    &= \frac{2\pi}{S}\sum_{k=0}^{S-1}\sum_{r=-\infty}^{\infty} \delta\!\left(\omega - \frac{2\pi(k + rS)}{S}\right)
\end{align}

where we have used $\sum_{n=-\infty}^{\infty} e^{-j\alpha n} = 2\pi\sum_{r=-\infty}^{\infty}\delta(\alpha-2\pi r)$. Letting $m = k + rS$, the double sum collapses to a single sum over all integers $m$:

\begin{equation}
    P(e^{j\omega}) = \frac{2\pi}{S}\sum_{m=-\infty}^{\infty} \delta\!\left(\omega - \frac{2\pi m}{S}\right)
    \label{eq:P_dtft}
\end{equation}

Due to the $2\pi$-periodicity of the DTFT, only $S$ distinct impulses fall within any interval of length $2\pi$. For the principal interval $[-\pi,\pi)$ one may restrict the index set to $m=0,1,\dots,S-1$ when operating inside the convolution integral. Multiplication in the time domain corresponds to circular convolution in the frequency domain:

\begin{equation}
    W(e^{j\omega}) = \frac{1}{2\pi} \int_{-\pi}^{\pi} X(e^{j\theta})\,P\bigl(e^{j(\omega-\theta)}\bigr)\,d\theta
    \label{eq:conv_down}
\end{equation}

Equations~\ref{eq:P_dtft} and \ref{eq:conv_down} give

\begin{align}
    W(e^{j\omega}) 
    &= \frac{1}{2\pi} \int_{-\pi}^{\pi} X(e^{j\theta})\,
       \frac{2\pi}{S}\sum_{m=-\infty}^{\infty} \delta\!\left(\omega-\theta - \frac{2\pi m}{S}\right)d\theta \nonumber \\
    &= \frac{1}{S}\sum_{m=-\infty}^{\infty} \int_{-\pi}^{\pi} X(e^{j\theta})\,
       \delta\!\left(\omega-\theta - \frac{2\pi m}{S}\right)d\theta
\end{align}

For a fixed $\omega\in(-\pi,\pi]$, the argument of the delta function is zero when $\theta = \omega - 2\pi m/S$. The integral is not zero only if this $\theta$ is within the integration limits $[-\pi,\pi)$. Because $X(e^{j\theta})$ is $2\pi$-periodic, exactly $S$ distinct values of $m$ contribute (conventionally $m=0,1,\dots,S-1$ after reindexing). Exploiting the sifting property of the Dirac delta, we obtain:

\begin{equation}
    W(e^{j\omega}) = \frac{1}{S}\sum_{k=0}^{S-1} X\!\left(e^{j(\omega - 2\pi k/S)}\right)
    \label{eq:W_dtft}
\end{equation}

Equation~\ref{eq:W_dtft} shows that the zero‑filled sequence possesses a spectrum consisting of $S$ shifted replicas of the original spectrum $X(e^{j\omega})$ (each scaled by $1/S$), packed into the interval $[-\pi,\pi)$. From Equation~\ref{eq:y_from_w} we have $y[m] = w[mS]$. Its DTFT is

\begin{equation}
    Y(e^{j\omega}) = \sum_{m=-\infty}^{\infty} y[m]\,e^{-j\omega m}
    = \sum_{m=-\infty}^{\infty} w[mS]\,e^{-j\omega m}
\end{equation}

Setting $n = mS$, we see that the sum runs over all integer multiples of $S$. Because $w[n]=0$ for $n$ not a multiple of $S$, we may sum over all integers $n$:

\begin{equation}
    Y(e^{j\omega}) = \sum_{n=-\infty}^{\infty} w[n]\,e^{-j(\omega/S) n}
    = W\!\left(e^{j\omega/S}\right)
    \label{eq:Y_from_W}
\end{equation}

Thus, the time‑compression $S$ in the time domain expands the frequency axis by a factor $S$; i.e., the DTFT of $y[m]$ is the same as that of $w[n]$ but evaluated at $\omega/S$. Substituting Equation~\ref{eq:W_dtft} into Equation~\ref{eq:Y_from_W} yields the DTFT of the downsampled sequence:

\begin{equation}
    Y(e^{j\omega}) = \frac{1}{S}\sum_{k=0}^{S-1} X\!\left(e^{j(\omega/S - 2\pi k/S)}\right)
    \label{eq:Y_final}
\end{equation}

This is a fundamental description of downsampling in the frequency domain. The original spectrum $X(e^{j\omega})$ is first compressed by the factor $S$ (argument $\omega/S$) and then shifted by multiples of $2\pi/S$. If $X(e^{j\omega})$ contains non-zero energy for frequencies $|\omega| \ge \pi/S$, the shifted replicas will overlap, causing high frequency components to irreversibly fold into lower frequencies, which is \textbf{spectral aliasing}. To avoid aliasing, the replicas must remain distinct. 

Equation~\ref{eq:Y_final} shows both spectral stretching and the potential for aliasing. Consider first the term corresponding to $k=0$, namely $\frac{1}{S} X(e^{j\omega/S})$. Here, the original spectrum $X(e^{j\omega})$ undergoes a scaling of the frequency axis by the factor $S$. A frequency component originally located at $\omega_0$ in $x[n]$ will appear at $S\omega_0$ in $Y(e^{j\omega})$. This expansion of the frequency axis is called spectral stretching.  Consequently, a narrowband signal in $x[n]$ spreads over a proportionally wider bandwidth in the downsampled sequence, and its amplitude is reduced by the factor $S$.

The remaining $S-1$ terms of $k=1,\dots,S-1$, have the form $\frac{1}{S} X\bigl(e^{j(\omega/S - 2\pi k/S)}\bigr)$. These arise from the shifted replicas created during the intermediate zero‑insertion step.  In the spectrum of the zero‑filled sequence $W(e^{j\omega})$ in Equation~\ref{eq:W_dtft}, these replicas are centred at $\omega = 2\pi k/S$. The time‑compression step $y[m]=w[mS]$ corresponds to the frequency mapping $Y(e^{j\omega}) = W(e^{j\omega/S})$, which expands the frequency axis by $S$. Hence, a replica originally centered at $2\pi k/S$ is mapped to the new center:

\begin{equation}
    S \cdot \frac{2\pi k}{S} = 2\pi k
\end{equation}

Because the DTFT is $2\pi$-periodic, all these centers fall at integer multiples of $2\pi$, which are exactly the boundaries of the principal interval $[-\pi,\pi)$. Thus, after the downsampling operation, every one of the $S$ replicas is assigned to the same baseband interval.

\begin{figure*}[htbp]
    \centering
    \includegraphics[width=0.85\textwidth]{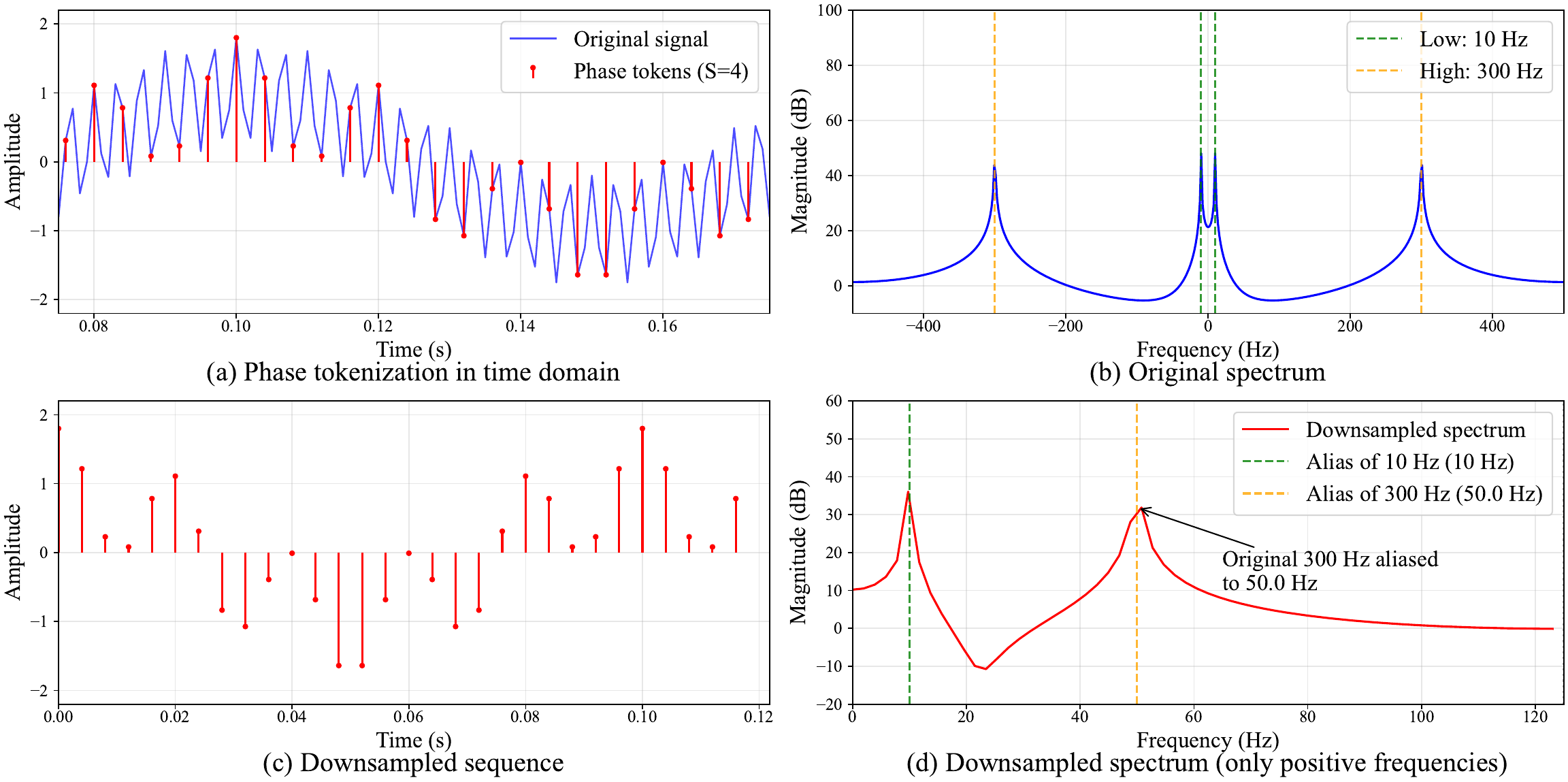} 
    \caption{Illustration of spectral aliasing in the frequency domain. When the original spectrum is not confined to $(-\pi/S,\pi/S)$, high frequency components appear as spurious low frequency components.}
    \label{fig:Appendix_spectral_aliasing}
\end{figure*}

Whether these replicas overlap destructively depends on the bandwidth of $X(e^{j\omega})$. If $X(e^{j\omega})$ contains non-zero energy for $|\omega| \ge \pi/S$, then the $k=0$ baseband replica already extends beyond $|\omega| \leq \pi$ after stretching, and the $k\neq 0$ terms bring additional copies of the spectrum that intrude into the same region. More precisely, a frequency $\omega$ in $Y(e^{j\omega})$ receives contributions from all $k$ for which $\omega/S - 2\pi k/S$ falls within the support of $X(e^{j\omega})$. As shown in Figure~\ref{fig:Appendix_spectral_aliasing}, when the original spectrum is not confined to $(-\pi/S,\pi/S)$, there exist frequencies at which two or more replicas contribute, causing high frequency components to appear as spurious low frequency components. This irreversible superposition is spectral aliasing. Because the overlapping terms all share the same frequency in $Y(e^{j\omega})$, the aliased components cannot be separated after downsampling, leading to permanent distortion.

To avoid aliasing entirely, the support of $X(e^{j\omega})$ must be a subset of $(-\pi/S,\pi/S)$. In other words, the maximum normalized frequency $\omega_{\max}$ present in $x[n]$ must satisfy

\begin{equation}
    \omega_{\max} \leq \frac{\pi}{S}
    \label{eq:nyquist_normalized}
\end{equation}

When this condition is satisfied, each frequency $\omega$ in the principal interval receives a contribution from exactly one value of $k$ (the one for which $\omega/S - 2\pi k/S$ lies within the support of $X$). The replicas are then placed side‑by‑side without overlapping, and the original signal can in principle be recovered from the downsampled version.

The condition can be expressed in terms of physical frequencies by relating $\omega$ to the underlying continuous‑time signal from which $x[n]$ was obtained. Assume $x[n] = x_c(nT)$, with the sampling period $T = 1/f_s$ and $f_s$ the original sampling rate. The normalized frequency is $\omega = 2\pi f / f_s$. If the analog signal $x_c(t)$ is bandlimited, say $|f| \leq B$, then $X(e^{j\omega})$ is non‑zero only for $|\omega| \leq 2\pi B/f_s$. Imposing the no‑aliasing requirement Equation~\ref{eq:nyquist_normalized} gives

\begin{equation}
    \frac{2\pi B}{f_s} \leq \frac{\pi}{S}
    \quad\Longrightarrow\quad
    B \leq \frac{f_s}{2S}
    \label{eq:nyquist_physical}
\end{equation}

After downsampling, the effective sampling rate becomes $f_s' = f_s / S$, and its associated Nyquist frequency is $f_s' / 2 = f_s/(2S)$. Therefore, alias‑free downsampling is possible if and only if the original continuous‑time signal is bandlimited to less than or equal to the new Nyquist frequency.

\subsection{Boundary Conditions for Distortion-Free Tokenization}
\label{app: Boundary Conditions}

The preceding analyses establish that patch tokenization and phase tokenization are governed by distinct mathematical constraints. We now formalize the precise boundary conditions to avoid irreversible distortion.

\subsubsection{Boundary Condition for Patch Tokenization}
\label{app: patch_boundary}

Patch tokenization, as defined in Equation \ref{eq:patch token}, is mathematically equivalent to multiplying the original sequence $x[n]$ by a rectangular window of length $L$ starting at index $n_0$, and then re-indexing the non‑zero segment from zero. The window sequence is $x_w[n] = x[n] \, w_L[n; n_0]$, and the final patch token is $y[m] = x_w[n_0 + m]$ for $m = 0, \dots, L-1$.

For an exact reconstruction of the original finite length sequence $x[n]$ of length $N$ from its patch token, it is necessary and sufficient that the window covers all non‑zero samples of $x[n]$. This requires $n_0 \le 0$ and $n_0 + L - 1 \ge N - 1$. When the window starts at $n_0 = 0$, the condition reduces to $L \ge N$. Under this condition, the token $y[m] = x[m]$ for $m = 0,\dots,N-1$ (and zero for $m \ge N$), so the original $x[n]$ can be recovered losslessly by discarding the zero‑padded tail.

If $x[n]$ is periodic with a fundamental period $T \in \mathbb{N}^+$, i.e., $x[n+T] = x[n]$ for all integers $n$, then the entire signal is fully determined by any contiguous block of $T$ samples. Hence, choosing the window length as an integer multiple of the period, $L = m T$ ($m \in \mathbb{N}^+$), guarantees that the token contains exactly the information needed to reconstruct the whole periodic sequence by periodic extension, irrespective of the starting index $n_0$.

Although time domain preservation is straightforward, the windowing operation inherently alters the frequency‑domain representation. As derived in Section~\ref{app: patch_analysis}, multiplication by the rectangular window convolves the true spectrum with a Dirichlet kernel, whose main‑lobe width ($4\pi/L$) limits frequency resolution and whose sidelobes cause spectral leakage.

For periodic signals with $L = mT$, an important special case arises: the discrete Fourier transform (or the DTFT evaluated at the harmonic frequencies) samples the Dirichlet kernel exactly at its zero crossings between harmonics. This coherent sampling completely eliminates spectral leakage at those discrete frequency bins because the contribution from any harmonic to a different harmonic's bin is identically zero. However, this cancellation occurs only at the sampled frequencies; the continuous DTFT still exhibits sidelobes between the bins, and the fundamental resolution limit imposed by the main‑lobe width persists. For non‑periodic signals or when $L \neq mT$, the Dirichlet kernel's nulls no longer align with the analysis frequencies, and spectral leakage becomes inevitable.

The resolution limit states that two sinusoids can be separated in the magnitude spectrum only if their frequency separation exceeds half the main‑lobe width, $|\omega_1 - \omega_2| \ge \frac{2\pi}{L}$. This limit is entirely independent of whether the lossless condition in the time domain is satisfied. This follows only from the finite observation length $L$.

In summary, the time‑domain lossless condition for patch tokenization is $L \ge N$ for finite signals, or $L = mT$ for periodic signals. In the frequency domain, spectral leakage is an unavoidable consequence of the finite rectangular window, except that for coherently sampled periodic signals it is null at the DFT bins.

\subsubsection{Boundary Condition for Phase Tokenization}
\label{app: phase_boundary}

Phase tokenization, equivalent to downsampling the original sequence by an integer factor $S \ge 2$, imposes a fundamentally different and more restrictive boundary condition. Compared to patch tokenization, the boundary condition of phase tokenization is much simpler. As derived from Equation~\ref{eq:Y_final}, the alias-free condition for the discrete time signal is given by Equation~\ref{eq:nyquist_physical}. This constitutes the necessary and sufficient condition for distortion-free phase tokenization. If the original signal violates this bandlimit, high-frequency energy residing outside the interval $|\omega| \leq \pi/S$ will be folded into the baseband, causing irreversible distortion.

\subsection{Analysis of TimeFT Methodology}
\label{app: Analysis of TimeFT}

\subsubsection{Analysis of Distortion}
\label{app: Analysis of Distortion}

Consider a real sequence $\mathbf{X}\in\mathbb{R}^{N}$. Its RFFT $\mathbf{X}_{\mathrm{freq}}\in\mathbb{C}^{N_f}$ collects the non‑negative frequency components, with the $f$-th bin corresponding to the angular frequency $\omega_f = 2\pi f/N$, $f=0,\dots,N_f-1$. The block size is chosen as $B_N = \bigl\lfloor N/(2S) \bigr\rfloor$, satisfying the boundary condition~\ref{eq:bound}.

The $k-th$ spectral block $\mathbf{X}_{\mathrm{block}}^{(k)}$ retains only the frequency bins $f\in[kB_N,\,kB_N+B_N-1]$ and zeros out the remaining ones. After the shifting step, i.e., reshaping into blocks of size $B_N$ and zero‑padding back to length $N_f$, each block becomes a spectrum $\widetilde{\mathbf{X}}^{(k)}$ concentrated in the low‑frequency interval $f=0,\dots,B_N-1$:

\begin{equation}
    \bigl[\widetilde{\mathbf{X}}^{(k)}\bigr]_{b} =
    \begin{cases}
        [\mathbf{X}_{\mathrm{freq}}]_{kB_N+b}, & 0 \le b \le B_N-1,\\
        0, & \text{otherwise}.
    \end{cases}
\end{equation}

The inverse RFFT of $\widetilde{\mathbf{X}}^{(k)}$ yields a real‑valued time‑domain sequence $\mathbf{x}_{\mathrm{time}}^{(k)}\in\mathbb{R}^{N}$.  By the properties of the DFT, the spectral support of $\mathbf{x}_{\mathrm{time}}^{(k)}$ is confined to (normalized) frequencies

\begin{equation}
    |\omega| \le \frac{2\pi (B_N-1)}{N} < \frac{2\pi B_N}{N}
    \label{eq:bw_time}
\end{equation}

We then obtain the strict band limit:

\begin{equation}
    |\omega| < \frac{2\pi}{N}\cdot\frac{N}{2S} = \frac{\pi}{S}
    \label{eq:bandlimit_pi_S}
\end{equation}

Equation~\ref{eq:bandlimit_pi_S} is exactly the alias‑free condition for downsampling by a factor $S$~(\S\ref{app: phase_analysis}). Hence, every sub‑band signal $\mathbf{x}_{\mathrm{time}}^{(k)}$ obeys the Nyquist criterion for the subsequent decimation step. The operator $\mathcal{D}_S$ defined in Equation~\ref{eq:D_s} extracts the polyphase components of $\mathbf{x}_{\mathrm{time}}^{(k)}$ with offsets $\lambda=0,\dots,S-1$. Because the signal is bandlimited to $|\omega|<\pi/S$, each decimated polyphase stream is free from spectral aliasing. Therefore, the entire information contained in the $k$-th frequency sub‑band is almost perfectly transferred to the tokens.

The simplified spectral shifting of Equation~\ref{eq: torch_shift} moves a complex Fourier coefficient originally located at frequency bin $kB_N+b$ directly to position $b$. When $b=0$, this coefficient becomes the DC term of the shifted block. For a real signal, the DC component of the RFFT must be purely real. The IRFFT therefore discards any imaginary part at that point. Consequently, a single complex value per block loses its imaginary part, a negligible fraction of the total coefficients. As shown in the visualization in Appendix~\ref{app: visualization}, TimeFT tokens can be well reconstructed into the original sequence, and thus this discarding is negligible. All other frequencies ($b\ne 0$) are almost perfectly preserved. Hence, TimeFT is almost distortion‑free.

\subsubsection{Linear Phase Induced by Frequency Shifting}
\label{app:linear_phase_rfft}

Let $x[n]$, $n=0,\dots,N-1$, be a real sequence and $X_r[f]$, $f=0,\dots,N_f-1$, its RFFT coefficients, where $N_f=\lfloor N/2\rfloor+1$. We isolate the $k$-th sub-band of width $B_N$, starting at index $k B_N$ ($B_N = \lfloor N/(2S)\rfloor$, $S$ is the downsampling factor) $\mathcal{I}_k = \{\,k B_N,\, k B_N+1,\dots,\, k B_N+B_N-1\,\}$. The spectral shifting stage moves the content of $\mathcal{I}_k$ to the baseband $0,\dots,B_N-1$. A natural implementation copies the coefficients without modification:

\begin{equation}
\widetilde{X}_r[b] = X_r[k B_N + b], \quad b=0,\dots,B_N-1
\label{eq:direct_copy}
\end{equation}

and zeros the remaining bins. We then perform a $N$-point inverse RFFT to obtain a real sequence $\tilde{x}^{(k)}[n]$ of length $N$. We will show that this simple copy operation already produces the correct baseband signal, no extra phase rotation is required when the inverse transform retains the original length $N$.

\paragraph{Why Direct Copy Preserves the Phase Relations.}
The $N$-point DFT of the original signal satisfies $X[\ell] = \sum_{n=0}^{N-1} x[n]\,e^{-j\frac{2\pi \ell n}{N}}$ and for real $x[n]$ we have $X_r[f] = X[f]$ for $0\le f\le N/2$. Consider the contribution of the $k$-th sub‑band to $x[n]$:

\begin{equation}
x_{\text{band},k}[n] = \frac{2}{N}\sum_{b=0}^{B_N-1} \operatorname{Re}\!\Bigl( X[k B_N+b]\,e^{j\frac{2\pi (k B_N+b) n}{N}} \Bigr)
\label{eq:orig_band}
\end{equation}

Now, take the DFT of the sequence $\tilde{x}^{(k)}$ obtained from the copied spectrum~\ref{eq:direct_copy}. By construction, its DFT is non‑zero only at indices $b$ and $N-b$ with coefficients $X[k B_N+b]$ and $X^*[k B_N+b]$, respectively. Therefore,
\begin{align}
\tilde{x}^{(k)}[n] &= \frac{1}{N}\sum_{b=0}^{B_N-1}\Bigl( X[k B_N+b]\,e^{j\frac{2\pi b n}{N}} + X^*[k B_N+b]\,e^{-j\frac{2\pi b n}{N}} \Bigr) \nonumber\\
&= \frac{2}{N}\sum_{b=0}^{B_N-1} \operatorname{Re}\!\Bigl( X[k B_N+b]\,e^{j\frac{2\pi b n}{N}} \Bigr)
\label{eq:direct_time}
\end{align}

Comparing Equation~\ref{eq:direct_time} with the demodulated version of \ref{eq:orig_band}, i.e., multiplying $x_{\text{band},k}[n]$ by $e^{-j\frac{2\pi (k B_N)}{N}n}$ and taking the real part, reveals that the two expressions are identical after canceling the complex exponential factor. Hence $\tilde{x}^{(k)}[n]$ is exactly the real baseband signal obtained by demodulating the $k$-th band to zero frequency and low‑pass filtering. All sinusoids maintain the same relative phases as they had in the original band.

\paragraph{When Does a Phase Factor Become Necessary?}
If we were to directly obtain a critically downsampled baseband signal of length $L = N/S$ without going through the $N$-point grid, we would need to perform an inverse transform of length $L$. In that case, the time‑axis compression changes the time origin: the new sample index $m=0$ corresponds to the original time $n=0$, but the frequency grid changes from $2\pi b/N$ to $2\pi b/L$. To correctly align the phases of the complex exponentials, each coefficient must be multiplied by $\exp(-j\frac{2\pi (k B_N)}{N}b)$, compensating for the shift of the frequency center $k B_N$ relative to the new sampling grid. Explicitly, if we form a length‑$L$ DFT as

\begin{equation}
Z[b] = \frac{1}{S}\,X[k B_N+b]\,e^{-j\frac{2\pi (k B_N)}{N}b}, \quad b=0,\dots,B_N-1
\end{equation}

and apply a length‑$M$ inverse DFT, the resulting sequence is exactly the downsampled version of $\tilde{x}^{(k)}[n]$ without any aliasing. This is the context in which the phase rotation $\exp(-j2\pi (k B_N)b/N)$ appears; it accounts for the different time origin of the shortened sequence.

\paragraph{TimeFT Avoids the Explicit Phase Factor.}

In the TimeFT pipeline we deliberately stay on the original $N$-point grid. After copying the coefficients, we apply an $N$-point inverse RFFT and then employ the dense sampling operator $\mathcal{D}_S$ to extract the tokens. Because the downsampling is performed after recovering the full‑length baseband signal, the phase relations are automatically correct because the phase factor is implicitly absorbed by the subsequent decimation step. Had we shortened the inverse transform to $B_N$ or $L$ points, the phase compensation would have been indispensable.

\subsubsection{Analysis of Multiple Tokens}
\label{app: Analysis of Multiple Tokens}

A natural question that arises is whether the distortions introduced by individual tokens can be mitigated when multiple tokens are processed jointly. One might argue that since a collection of tokens together covers the entire original sequence, no information is lost in principle. However, in the case of patch tokenization, this argument does not hold when tokens are processed independently, which is the standard step in models such as PatchTST~\cite{PatchTST}. Consequently, the patterns that the model learns are inherently distorted.

Several works have proposed cross‑token learning mechanisms that allow information to flow between tokens~\cite{XCTFormer, Token_merging}. These methods can partially alleviate the distortion by enabling tokens to share contextual information, but their cross‑token capacity is inherently bounded, which cannot cover all tokens simultaneously and therefore cannot fully recover the missing structure. Although these methods demonstrate the existence of the problem, they cannot fundamentally solve it. 

In contrast, our TimeFT tokenization solves this problem. The token can almost losslessly reconstruct the original signal. At first glance, this may appear to introduce redundant information. However, this can be approximately viewed as a phase decomposition in the frequency domain. Extensive experiments on forecasting, classification, and anomaly detection confirm that this representation consistently improves model performance rather than introducing harmful redundancy. Furthermore, based on Appendix~\ref{app: phase_analysis}, Phase Tokens also contain only the low-frequency information of the original sequence, resulting in similar information redundancy and even worse because of frequency aliasing. However, phase tokens have also been shown to be an effective tokenization method, such as PhaseFormer~\cite{PhaseFormer}. Therefore, this phenomenon does not pose a risk.

\subsection{Difference between Quantization Methods and Our Main Topic}
\label{app:Difference_Quantization}

Time series quantization methods originate from attempts to process time series in a text‑like fashion inspired by large language models~\cite{Chronos1}. They discretize individual time points from continuous values into a finite set of bins, leaving the input shape unchanged. This paradigm is fundamentally different from the tokenization framework that dominates current time series modeling. A large number of current TSFMs~\cite{Chronos2, Sundial, TimeFM} and deep learning time series models~\cite{PatchTST, PhaseFormer, SRSNet} adopt patch or phase tokenizations, whereas quantization remains a marginal approach.

For instance, Apollo~\cite{Apollo} views quantization as a sampling procedure and proposes to first remove high‑frequency content before discretizing the continuous data points, a scheme analogous to the ``bins'' used in Chronos~\cite{Chronos1}. Because such methods do not involve partitioning the time series into subsequences, they fall outside the scope of our tokenization focused analysis. We therefore do not discuss them further.

Our work is the first to provide a rigorous frequency domain theoretical foundation for the distortions introduced by the prevailing tokenization paradigm, and to derive precise boundary conditions for distortion‑free tokenization. Building on this novel perspective, we propose TimeFT, a general, parameter‑free, and highly effective tokenizer that consistently improves performance in diverse tasks and models.

\section{Reconstruction Methods}
\label{app: Reconstruction_method}

\subsection{The further explanation of Figure~\ref{fig: figure2}}

In Figure~\ref{fig: figure2}, the further explanation of spectrum is as follows:

(a) Patch Token Spectrum: The rectangular window truncation causes the original spectrum to be convolved with a Dirichlet kernel, producing a broadened main lobe and oscillatory sidelobes. The true frequency components are indistinguishable due to leakage. The reconstruction (c) shows severe deviation from the original.

(b) Phase Token Spectrum: Downsampling by factor S creates S-1 aliased replicas that fold into the baseband. The Nyquist frequency is marked with a black vertical line; components above this threshold alias into lower frequencies, creating false peaks.

(c) TimeFT: The spectrum is partitioned into alias-free bands, shifted to baseband, and Nyquist-sampled. The reconstruction (i) is nearly identical to the original, as shown in Table~\ref{tab: Reconstruction Performance}.

\subsection{Reconstruction from Patch Tokens}

A patch token is obtained by multiplying the original sequence $x[n]$ of length $N$ by a rectangular window of length $L$ starting at $n_0$, and then discarding the zero entries. Reconstruction therefore amounts to padding the token with zeros to the original length and shifting it back to its original temporal position.

\begin{equation}
\hat{x}[n] = \begin{cases}
y[n-n_0], & n_0 \le n \le n_0+L-1,\\
0, & \text{otherwise}.
\end{cases}
\end{equation}

Equivalently, in the frequency domain,

\begin{equation}
\hat{X}(e^{j\omega}) = Y(e^{j\omega})\,e^{-j\omega n_0} = X_w(e^{j\omega})
\end{equation}

where $X_w$ is the DTFT of the windowed sequence. As shown in Appendix~\ref{app: patch_analysis}, it inherits these distortions without any possibility of recovery, the discarded information is permanently lost.

\subsection{Reconstruction from Phase Tokens}

A phase token is obtained by downsampling the original sequence $x[n]$ of length $N$ by an integer factor $S$, which yields $y[m] = x[mS]$ for $m=0,\dots,M_f-1$, where $M_f = \lfloor N/S\rfloor$. Reconstruction consists in upsampling $y[m]$ back to the original length $N$. The practical method used in our experiments is Fourier‑based sinc interpolation. In terms of RFFT, the algorithm computes the $M_f$-point RFFT of the real sequence $y$,

\begin{equation}
Y_r[k] = \sum_{m=0}^{M_f-1} y[m]\,e^{-j2\pi km/M_f}, \qquad k = 0,\dots,\lfloor M_f/2\rfloor
\end{equation}

and then zero‑pads it to the length required by the original signal. The RFFT of the reconstructed sequence $\hat{x}[n]$ thus becomes

\begin{equation}
\hat{X}_r[k] = \begin{cases}
Y_r[k], & 0 \le k \le \lfloor M_f/2\rfloor,\\[6pt]
0, & \lfloor M_f/2\rfloor < k \le \lfloor N_f/2\rfloor.
\end{cases}
\label{eq:rfft_zeropad}
\end{equation}

The time domain reconstruction is finally obtained by IRFFT, $\hat{x} = \mathcal{F}_{\mathbb{R}}^{-1}(\hat{X}_r)$. We denote the upsampling operator here as $\mathcal{R}(\cdot)$. Since $\lfloor M_f/2\rfloor = \lfloor N/(2S)\rfloor$, zero‑padding in Equation~\ref{eq:rfft_zeropad} is equivalent to applying an ideal low‑pass filter with cutoff frequency $\pi/S$ to the upsampled sequence. The quality of reconstruction critically depends on whether the original signal is bandlimited to $|\omega|\leq\pi/S$. If $x[n]$ contains energy at frequencies $|\omega| > \pi/S$, the downsampling step produces spectral aliasing that maps high‑frequency components to the low‑frequency region~(\S\ref{app: phase_analysis}). The upsampling process cannot distinguish between a true low‑frequency component and an aliased high‑frequency one, so the reconstructed signal will still contain wrong patterns.

\subsection{Reconstruction from TimeFT Tokens}
\label{app:Reconstruction from TimeFT Tokens}

The TimeFT tokenizer produces $S$ tokens per variate. Each token is a vector of length $K \cdot L$, where $K=\lceil N_f/B\rceil$ is the number of frequency blocks, $B=\lfloor N/(2S)\rfloor$ is the bandwidth, and $L=\lceil N/S\rceil$ is an intermediate variable. Reconstruction proceeds by inverting the two stages of the tokenizer: (i) polyphase recomposition and (ii) inverse spectral shifting.

\paragraph{Polyphase Recomposition.} For a fixed variate and frequency block index $k$, the $S$ tokens with offsets $\lambda=0,\dots,S-1$ represent the $S$ polyphase components of the bandlimited time domain sequence $\mathbf{x}_{\mathrm{time}}^{(k)}$ obtained after spectral shifting and IRFFT~\ref{eq:IRFFT}. The reconstruction first upsamples every sequence back to length $N$ using the same Fourier-based sinc interpolation described in the previous subsection:

\begin{equation}
\hat{\mathbf{x}}_{\mathrm{time}}^{(k)} = \mathcal{R}(\mathbf{x}_{\mathrm{time}}^{(k)})
\end{equation}

Equivalently, this can be expressed as the output of a $S$-channel synthesis filter bank with ideal interpolation filters.

\paragraph{Inverse Spectral Shifting.} The reconstructed bandlimited signal $\hat{\mathbf{x}}_{\mathrm{time}}^{(k)}$ occupies the low-frequency region of the $k$-th spectral block. To recover the contribution of this block to the full spectrum, we apply $N$-point RFFT, obtaining $\hat{\mathbf{X}}_{\mathrm{block}}^{(k)}\in\mathbb{C}^{N_f}$. Only the first $B$ frequency bins are retained and the remainder are set to zero.

\begin{equation}
\bigl[\hat{\mathbf{X}}_{\mathrm{block}}^{(k)}\bigr]_{f} = \begin{cases}
[\operatorname{RFFT}(\hat{\mathbf{x}}_{\mathrm{time}}^{(k)})]_{f}, & 0\le f < B,\\
0, & \text{otherwise}.
\end{cases}
\end{equation}

The original spectrum was partitioned into contiguous blocks of size $B$~\ref{eq:band_extraction}). The inverse operation simply places the retained frequency bins of the $k$-th block back into their original positions:

\begin{equation}
\hat{\mathbf{X}}_{\mathrm{freq}}[kB + b] = \bigl[\hat{\mathbf{X}}_{\mathrm{block}}^{(k)}\bigr]_{b}, \qquad k=0,\dots,K-1,\;\; b=0,\dots,B-1
\end{equation}

All bins of $\hat{\mathbf{X}}_{\mathrm{freq}}$ that lie beyond the last block (if any) are obtained from the zero-padding that was applied during the forward decomposition. Finally, an IRFFT produces the reconstructed time domain sequence:

\begin{equation}
\hat{\mathbf{x}} = \mathcal{F}_{\mathbb{R}}^{-1}\bigl(\hat{\mathbf{X}}_{\mathrm{freq}}\bigr) \in \mathbb{R}^N
\end{equation}

\paragraph{Fidelity of the Reconstruction.}

Spectral shifting does not distort the raw time series. For the frequency, spectral shifting is merely a change of basis. Moving a frequency band from to can be viewed as a signal modulation rather than a distortion. Moreover, since our method satisfies the boundary condition in Eq. 4, no aliasing issues arise. For the magnitude, the phase factor has a modulus of 1, which preserves all amplitude.

The only deviation from perfect reconstruction comes from the simplified spectral shifting of Equation~\ref{eq: torch_shift}, causing the DC coefficient of each block to lose its imaginary part. Apart from this single complex value per block, the entire process is algebraically equivalent to passing the signal through a bank of ideal bandpass filters followed by perfect reconstruction synthesis. Because the block bandwidth $B$ is chosen to respect the alias-free condition $B\le N/(2S)$, no aliasing occurs during downsampling, and polyphase decomposition captures all samples without loss. Consequently, the reconstructed signal $\hat{\mathbf{x}}$ differs from the original $\mathbf{x}$ only in the imaginary parts of the zero-frequency bins of each spectral block. This distortion is negligible in magnitude and that leaves the amplitude spectrum essentially identical to that of the original. This explains why TimeFT tokens are almost distortion‑free, instead of information loss of patch tokens and aliasing artifacts from phase tokens.

\section{Full Experiment Results}
\label{app: Experiments}

In this section, we present detailed experiments of TimeFT, including three major tasks of long-term forecasting~\ref{app:Long_Term_Forecasting_all}, classification~\ref{app:Classification_all}
and anomaly detection~\ref{app:anomaly_detection_all}. Moreover, we provide settings for other experiments~\ref{app:Setting in Other Experiments}, variation in model parameters~\ref{app:Variation in Model Parameter Count} and reconstruction performance of different tokenizations~\ref{app:Reconstruction Performance across Different Token Types}. The computational efficiency~(\S\ref{subsec: time_complexity}), the plug-in experiment~(\S\ref{subsec: Plug_in_experiment}), the ablation study~(\S\ref{subsec: ablation_study}) and the hyperparameter sensitivity~(\S\ref{subsec: hyperparameter sensitivity analysis}) please see Section~\ref{sec: Experiments}. All experiments were performed using Intel Core i7-13700H CPU and NVIDIA GeForce RTX 4060 GPU. The training and testing procedures in the experiments follow the Time Series Library~\cite{tslibrary}. 

In the experiments, we select datasets following the standard protocol established by prior works. \textbf{Long‑term forecasting} is performed on seven real‑world datasets: ETTh1, ETTh2, ETTm1, ETTm2, Weather, Electricity and Traffic~\cite{AutoFormer, TimesNet, iTransformer}. \textbf{Classification} uses all UEA datasets~\cite{EntroPE, Classification_survey, UEA}. \textbf{Anomaly detection} is conducted on five widely used benchmarks~\cite{EntroPE, Anomaly_detection_survey, ADbench}: SMD, MSL, SMAP, SWaT, and PSM.

\subsection{Long-Term Forecasting Task}
\label{app:Long_Term_Forecasting_all}

We evaluate the long-term forecasting performance of TimeFT against state-of-the-art tokenization baselines across diverse benchmarks. The experiments adopt the same settings as TimeMosaic~\cite{TimeMosaic}. We evaluate our model on widely used real-world time series forecasting datasets, including Electricity Transformer Temperature (ETT), Weather, Electricity, and Traffic. The details of the datasets for this task are as follows:

\begin{itemize}
    \item \textbf{ETT~\footnote{\url{https://github.com/zhouhaoyi/ETDataset}}} comprises four subsets: ETTh1 and ETTh2 (hourly), and ETTm1 and ETTm2 (every 15 minutes), collected from two electricity transformers.
    \item \textbf{Weather~\footnote{\url{https://www.bgc-jena.mpg.de/wetter}}} contains 21 meteorological variables recorded every 10 minutes in 2020 at the Max Planck Institute for Biogeochemistry.
    \item \textbf{Electricity~\footnote{\url{https://archive.ics.uci.edu/dataset/321/electricityloaddiagrams20112014}}} provides hourly power consumption data of 321 clients from 2012 to 2014.
    \item \textbf{Traffic\footnote{\url{https://pems.dot.ca.gov}}} consists of hourly road occupancy rates from 862 freeway sensors in the San Francisco Bay Area between January 2015 and December 2016.
\end{itemize}

Table~\ref{tab:long_term_forecasting_datasets} summarizes the key characteristics of the datasets used in our evaluation. The collection encompasses various sampling rates, ranging from 10-minute and 15-minute intervals to hourly and daily recordings. And they exhibits considerable diversity in dimensionality, spanning from the 7-dimensional ETT subsets to the 321-dimensional ECL dataset. For each dataset, we report the partition sizes in the format (Training, Validation, Test) to clearly specify the data splitting strategy adopted across all experiments.

\begin{table}[htbp]
    \centering
    \caption{Summary of datasets for long-term forecasting task. Dim denotes dimension, which is the variate number of each dataset. Dataset sizes represent the number of temporal observations in each partition (Training, Validation, Test). Frequency denotes the sampling interval of time points.}
    \label{tab:long_term_forecasting_datasets}
    \small
    \begin{tabular}{lcccl}
        \toprule
        Dataset & Dim & \makecell{Dataset Size \\ (Train, Val, Test)} & Frequency & Domain Information \\
        \midrule
        ETTm1      & 7   & (34465, 11521, 11521)  & 15min  & Energy Infrastructure \\
        ETTm2      & 7   & (34465, 11521, 11521)  & 15min  & Energy Infrastructure \\
        ETTh1      & 7   & (8545, 2881, 2881)    & Hourly & Energy Infrastructure \\
        ETTh2      & 7   & (8545, 2881, 2881)    & Hourly & Energy Infrastructure \\
        Weather    & 21  & (36792, 5271, 10540)  & 10min  & Meteorological Monitoring \\
        Electricity & 321 & (18317, 2633, 5261)  & Hourly & Electricity Consumption \\
        Traffic & 862 & (12185,1757,3509) & Hourly & Traffic Flow\\
        \bottomrule
    \end{tabular}
\end{table}

As demonstrated in Table \ref{tab:long_term_forecasting_all}, TimeFT achieves the best performance in the majority of datasets and metrics. By outperforming both classic and recent baselines, TimeFT establishes an excellent paradigm for obtaining tokens of time series.

\begin{table*}[htbp]
    \centering
    \caption{For long-term forecasting, we evaluate prediction horizons $H\in\{96,192,336,720\}$. All baselines use a fixed input sequence length of $320$. The term “Avg.” refers to the average metric over the four horizons. The best and second-best results are marked as \best{best} and \second{second}, respectively. The column “$1^{st}$ Count” reports the number of times each method ranks first among all competitors.}
    \label{tab:long_term_forecasting_all}
    \renewcommand{\arraystretch}{1.4}
    \setlength{\tabcolsep}{2pt}

    \resizebox{\textwidth}{!}{
    \begin{tabular}{c c | cc | cc | cc | cc | cc | cc | cc | cc | cc | cc | cc | cc}
        \toprule
        \multicolumn{2}{c|}{\multirow{2}{*}{Models}} & \multicolumn{2}{c|}{TimeFT} & \multicolumn{2}{c|}{EntroPE} & \multicolumn{2}{c|}{TimeMosaic} & \multicolumn{2}{c|}{PhaseFormer} & \multicolumn{2}{c|}{TimeFilter} & \multicolumn{2}{c|}{xPatch} & \multicolumn{2}{c|}{PatchMLP} & \multicolumn{2}{c|}{LightTS} & \multicolumn{2}{c|}{iTransformer} & \multicolumn{2}{c|}{TimeMixer} & \multicolumn{2}{c|}{PatchTST} & \multicolumn{2}{c}{DLinear} \\
        \multicolumn{2}{c|}{} & \multicolumn{2}{c|}{(\textbf{Ours})} & \multicolumn{2}{c|}{(2026)} & \multicolumn{2}{c|}{(2026)} & \multicolumn{2}{c|}{(2026)} & \multicolumn{2}{c|}{(2025)} & \multicolumn{2}{c|}{(2025)} & \multicolumn{2}{c|}{(2025)} & \multicolumn{2}{c|}{(2024)} & \multicolumn{2}{c|}{(2024)} & \multicolumn{2}{c|}{(2024)} & \multicolumn{2}{c|}{(2023)} & \multicolumn{2}{c}{(2023)} \\
        \cmidrule{3-26}
        \multicolumn{2}{c|}{Metric} & MSE & MAE & MSE & MAE & MSE & MAE & MSE & MAE & MSE & MAE & MSE & MAE & MSE & MAE & MSE & MAE & MSE & MAE & MSE & MAE & MSE & MAE & MSE & MAE \\
        \midrule
        \multirow{5}{*}{\rotatebox{90}{ETTm1}} 
        & 96  & \best{0.290} & \best{0.339} & 0.316 & 0.349 & 0.304 & \second{0.344} & 0.312 & 0.352 & 0.355 & 0.373 & \second{0.293} & 0.347 & 0.305 & 0.357 & 0.313 & 0.363 & 0.327 & 0.371 & 0.307 & 0.352 & 0.321 & 0.361 & 0.299 & \second{0.344} \\
        & 192 & \best{0.325} & \best{0.357} & 0.344 & 0.371 & 0.339 & \second{0.364} & 0.339 & 0.381 & 0.376 & 0.396 & \second{0.332} & 0.371 & 0.351 & 0.384 & 0.350 & 0.387 & 0.357 & 0.388 & \second{0.332} & 0.373 & 0.342 & 0.385 & 0.337 & 0.370 \\
        & 336 & \best{0.360} & \best{0.381} & 0.386 & 0.394 & 0.375 & \second{0.385} & 0.375 & 0.398 & 0.401 & 0.413 & 0.364 & 0.395 & 0.387 & 0.403 & 0.399 & 0.418 & 0.393 & 0.412 & \second{0.362} & 0.390 & 0.386 & 0.410 & 0.369 & \second{0.385} \\
        & 720 & \best{0.418} & \second{0.415} & 0.431 & 0.419 & \second{0.422} & \best{0.413} & 0.428 & 0.437 & 0.465 & 0.449 & 0.428 & 0.432 & 0.454 & 0.441 & 0.489 & 0.479 & 0.467 & 0.452 & \second{0.422} & 0.426 & 0.439 & 0.441 & 0.424 & 0.419 \\
        \cmidrule{2-26}
        & Avg. & \best{0.348} & \best{0.373} & 0.369 & 0.383 & 0.360 & \second{0.377} & 0.364 & 0.392 & 0.399 & 0.408 & \second{0.354} & 0.386 & 0.374 & 0.396 & 0.388 & 0.412 & 0.386 & 0.406 & 0.356 & 0.385 & 0.372 & 0.399 & 0.357 & 0.380 \\
        \midrule
        \multirow{5}{*}{\rotatebox{90}{ETTm2}}
        & 96  & \best{0.162} & \best{0.244} & 0.173 & 0.256 & \second{0.164} & \second{0.247} & 0.174 & 0.261 & 0.174 & 0.261 & 0.168 & 0.259 & 0.179 & 0.265 & 0.187 & 0.286 & 0.187 & 0.271 & 0.169 & 0.258 & 0.181 & 0.270 & 0.172 & 0.270 \\
        & 192 & \best{0.223} & \best{0.287} & 0.231 & 0.301 & \second{0.226} & \second{0.291} & 0.231 & 0.306 & 0.229 & 0.298 & \second{0.226} & 0.299 & 0.232 & 0.301 & 0.311 & 0.384 & 0.251 & 0.314 & 0.240 & 0.308 & 0.241 & 0.311 & 0.239 & 0.322 \\
        & 336 & \best{0.275} & \best{0.321} & 0.288 & 0.334 & \second{0.279} & \second{0.324} & 0.291 & 0.342 & 0.312 & 0.351 & 0.281 & 0.337 & 0.285 & 0.334 & 0.399 & 0.418 & 0.319 & 0.355 & 0.282 & 0.334 & 0.303 & 0.348 & 0.305 & 0.367 \\
        & 720 & \second{0.361} & \best{0.375} & 0.364 & 0.388 & \best{0.353} & \second{0.376} & 0.374 & 0.388 & 0.396 & 0.408 & 0.372 & 0.391 & 0.380 & 0.394 & 0.489 & 0.479 & 0.413 & 0.412 & 0.369 & 0.391 & 0.385 & 0.399 & 0.412 & 0.432 \\
        \cmidrule{2-26}
        & Avg. & \best{0.255} & \best{0.307} & 0.264 & 0.320 & \second{0.256} & \second{0.310} & 0.268 & 0.324 & 0.278 & 0.330 & 0.262 & 0.322 & 0.269 & 0.324 & 0.347 & 0.392 & 0.293 & 0.338 & 0.265 & 0.323 & 0.278 & 0.332 & 0.282 & 0.348 \\
        \midrule
        \multirow{5}{*}{\rotatebox{90}{ETTh1}}
        & 96  & \second{0.376} & \second{0.392} & 0.378 & 0.396 & \best{0.362} & \best{0.387} & 0.389 & 0.411 & 0.460 & 0.457 & 0.380 & 0.403 & 0.411 & 0.426 & 0.432 & 0.443 & 0.403 & 0.417 & \second{0.376} & 0.398 & 0.403 & 0.423 & 0.377 & 0.400 \\
        & 192 & 0.417 & \second{0.416} & \second{0.409} & 0.418 & \best{0.400} & \best{0.411} & 0.461 & 0.472 & 0.439 & 0.443 & 0.425 & 0.430 & 0.443 & 0.445 & 0.474 & 0.472 & 0.443 & 0.444 & 0.412 & 0.420 & 0.479 & 0.478 & 0.410 & 0.420 \\
        & 336 & \best{0.434} & \best{0.429} & 0.447 & 0.446 & \second{0.436} & \second{0.435} & 0.485 & 0.476 & 0.474 & 0.463 & 0.444 & 0.439 & 0.468 & 0.461 & 0.515 & 0.501 & 0.470 & 0.464 & 0.445 & 0.446 & 0.490 & 0.481 & 0.452 & 0.458 \\
        & 720 & \best{0.446} & \best{0.458} & 0.462 & 0.474 & \second{0.451} & \second{0.460} & 0.619 & 0.542 & 0.670 & 0.564 & 0.534 & 0.509 & 0.525 & 0.508 & 0.592 & 0.567 & 0.632 & 0.573 & 0.488 & 0.489 & 0.641 & 0.559 & 0.477 & 0.497 \\
        \cmidrule{2-26}
        & Avg. & \second{0.417} & \best{0.423} & 0.424 & \second{0.433} & \best{0.412} & \best{0.423} & 0.489 & 0.475 & 0.511 & 0.482 & 0.446 & 0.445 & 0.462 & 0.460 & 0.503 & 0.496 & 0.487 & 0.475 & 0.430 & 0.438 & 0.503 & 0.485 & 0.429 & 0.444 \\
        \midrule
        \multirow{5}{*}{\rotatebox{90}{ETTh2}}
        & 96  & 0.297 & \best{0.339} & 0.308 & 0.354 & 0.300 & \second{0.342} & 0.309 & 0.361 & 0.367 & 0.389 & \best{0.280} & 0.343 & 0.311 & 0.371 & 0.324 & 0.383 & 0.323 & 0.369 & \second{0.285} & 0.346 & 0.324 & 0.374 & 0.305 & 0.368 \\
        & 192 & \second{0.349} & \best{0.383} & 0.366 & 0.392 & 0.354 & 0.387 & 0.364 & 0.403 & 0.372 & 0.404 & \best{0.345} & \second{0.386} & 0.381 & 0.413 & 0.438 & 0.455 & 0.421 & 0.422 & 0.358 & 0.391 & 0.378 & 0.416 & 0.407 & 0.435 \\
        & 336 & \best{0.375} & \best{0.401} & 0.387 & 0.411 & \second{0.378} & 0.403 & 0.392 & 0.425 & 0.402 & 0.426 & 0.382 & 0.415 & 0.401 & 0.427 & 0.496 & 0.490 & 0.440 & 0.441 & 0.380 & 0.409 & 0.405 & 0.432 & 0.493 & 0.490 \\
        & 720 & \second{0.410} & \best{0.437} & 0.434 & 0.447 & 0.421 & \second{0.439} & 0.469 & 0.477 & 0.465 & 0.464 & \best{0.408} & 0.443 & 0.426 & 0.451 & 1.151 & 0.749 & 0.459 & 0.466 & \second{0.410} & \second{0.439} & 0.484 & 0.484 & 0.762 & 0.620 \\
        \cmidrule{2-26}
        & Avg. & \second{0.358} & \best{0.390} & 0.374 & 0.401 & 0.363 & \second{0.393} & 0.384 & 0.417 & 0.402 & 0.421 & \best{0.354} & 0.397 & 0.380 & 0.416 & 0.602 & 0.520 & 0.411 & 0.425 & \second{0.358} & 0.396 & 0.398 & 0.427 & 0.492 & 0.478 \\
        \midrule
        \multirow{5}{*}{\rotatebox{90}{Weather}} 
        & 96  & \second{0.147} & \second{0.194} & 0.153 & 0.197 & \best{0.144} & \best{0.189} & 0.152 & 0.201 & 0.156 & 0.209 & 0.148 & 0.196 & 0.153 & 0.206 & 0.152 & 0.212 & 0.160 & 0.209 & 0.150 & 0.199 & 0.155 & 0.204 & 0.177 & 0.242 \\
        & 192 & 0.199 & \second{0.236} & 0.201 & 0.239 & \second{0.197} & \best{0.232} & 0.201 & 0.249 & 0.195 & 0.246 & \best{0.190} & 0.238 & 0.197 & 0.245 & 0.197 & 0.255 & 0.207 & 0.251 & 0.193 & 0.244 & 0.201 & 0.249 & 0.217 & 0.276 \\
        & 336 & 0.248 & \second{0.279} & 0.254 & 0.283 & 0.245 & \best{0.276} & 0.249 & 0.282 & 0.251 & 0.286 & \best{0.241} & 0.280 & 0.247 & 0.282 & 0.249 & 0.299 & 0.256 & 0.289 & \second{0.244} & 0.283 & 0.255 & 0.289 & 0.262 & 0.312 \\
        & 720 & 0.332 & 0.335 & 0.339 & 0.336 & 0.327 & \best{0.327} & 0.331 & 0.340 & 0.322 & 0.339 & \best{0.318} & \second{0.333} & 0.324 & 0.334 & 0.323 & 0.357 & 0.325 & 0.338 & \second{0.319} & 0.336 & 0.333 & 0.340 & 0.326 & 0.367 \\
        \cmidrule{2-26}
        & Avg. & 0.232 & \second{0.261} & 0.237 & 0.264 & 0.228 & \best{0.256} & 0.233 & 0.268 & 0.231 & 0.270 & \best{0.224} & 0.262 & 0.230 & 0.267 & 0.230 & 0.281 & 0.237 & 0.272 & \second{0.227} & 0.266 & 0.236 & 0.271 & 0.246 & 0.299 \\
        \midrule
        \multirow{5}{*}{\rotatebox{90}{Electricity}}
        & 96  & \second{0.139} & \second{0.234} & 0.148 & 0.241 & 0.140 & 0.236 & 0.145 & 0.248 & \best{0.137} & \best{0.233} & 0.143 & 0.241 & 0.165 & 0.271 & 0.171 & 0.285 & 0.145 & 0.244 & 0.141 & 0.238 & 0.142 & 0.245 & 0.143 & 0.242 \\
        & 192 & \best{0.151} & \best{0.245} & 0.160 & 0.258 & 0.156 & 0.250 & 0.160 & 0.261 & \second{0.153} & \second{0.246} & 0.157 & 0.252 & 0.183 & 0.288 & 0.192 & 0.305 & 0.165 & 0.263 & 0.158 & 0.254 & 0.159 & 0.259 & 0.157 & 0.255 \\
        & 336 & \best{0.166} & \best{0.259} & 0.181 & 0.275 & 0.173 & 0.267 & 0.178 & 0.275 & \second{0.169} & \second{0.263} & 0.172 & 0.266 & 0.201 & 0.304 & 0.214 & 0.324 & 0.184 & 0.282 & 0.175 & 0.270 & 0.176 & 0.276 & 0.172 & 0.272 \\
        & 720 & \best{0.206} & \best{0.293} & 0.219 & 0.309 & 0.212 & 0.298 & 0.224 & 0.309 & 0.208 & \second{0.296} & 0.209 & 0.297 & 0.250 & 0.343 & 0.251 & 0.349 & 0.227 & 0.317 & 0.217 & 0.307 & 0.215 & 0.308 & \second{0.207} & 0.305 \\
        \cmidrule{2-26}
        & Avg. & \best{0.165} & \best{0.258} & 0.177 & 0.271 & 0.170 & 0.263 & 0.177 & 0.273 & \second{0.167} & \second{0.260} & 0.170 & 0.264 & 0.200 & 0.302 & 0.207 & 0.316 & 0.180 & 0.277 & 0.173 & 0.267 & 0.173 & 0.272 & 0.170 & 0.269 \\
        \midrule
        \multirow{5}{*}{\rotatebox{90}{Traffic}}
        & 96  & \best{0.395} & \second{0.279} & 0.421 & 0.304 & 0.413 & 0.297 & 0.418 & 0.289 & \second{0.397} & \best{0.274} & 0.413 & 0.297 & 0.486 & 0.361 & 0.493 & 0.367 & 0.422 & 0.310 & 0.421 & 0.297 & 0.403 & 0.285 & 0.422 & 0.296 \\
        & 192 & \second{0.412} & \second{0.281} & 0.433 & {0.288} & 0.424 & {0.281} & 0.426 & 0.301 & \best{0.411} & \best{0.278} & 0.425 & 0.298 & 0.515 & 0.378 & 0.511 & 0.373 & 0.449 & 0.327 & 0.441 & 0.311 & {0.419} & {0.292} & 0.435 & 0.302 \\
        & 336 & \best{0.421} & \second{0.291} & 0.446 & {0.295} & 0.435 & {0.287} & 0.443 & 0.304 & \second{0.423} & \best{0.284} & 0.430 & 0.297 & 0.533 & 0.386 & 0.521 & 0.378 & 0.469 & 0.339 & {0.425} & 0.317 & 0.432 & 0.298 & 0.447 & 0.308 \\
        & 720 & \second{0.452} & \second{0.309} & 0.477 & {0.314} & 0.464 & {0.302} & 0.478 & 0.324 & \best{0.449} & \best{0.300} & 0.450 & 0.306 & 0.564 & 0.401 & 0.504 & 0.360 & 0.512 & 0.364 & 0.489 & 0.340 & {0.459} & {0.313} & 0.476 & 0.327 \\
        \cmidrule{2-26}
        & Avg. & \best{0.420} & 0.290 & 0.444 & {0.300} & 0.434 & \second{0.287} & 0.441 & 0.305 & \best{0.420} & \best{0.284} & 0.430 & 0.300 & 0.525 & 0.382 & 0.507 & 0.370 & 0.463 & 0.335 & 0.444 & 0.316 & \second{0.428} & {0.297} & 0.445 & 0.308 \\
        \midrule
        \multicolumn{2}{c|}{$1^{st}$ Count} & \multicolumn{2}{c|}{\best{40}} & \multicolumn{2}{c|}{0} & \multicolumn{2}{c|}{\second{14}} & \multicolumn{2}{c|}{0} & \multicolumn{2}{c|}{10} & \multicolumn{2}{c|}{8} & \multicolumn{2}{c|}{0} & \multicolumn{2}{c|}{0} & \multicolumn{2}{c|}{0} & \multicolumn{2}{c|}{0} & \multicolumn{2}{c|}{0} & \multicolumn{2}{c}{0} \\
        \bottomrule
    \end{tabular}
    }
\end{table*}

\subsection{Classification}
\label{app:Classification_all}

For the classification task, we employ publicly available multivariate time series datasets from the University of East Anglia (UEA~\footnote{\url{https://www.timeseriesclassification.com/}}) Time Series Classification repository, which are widely recognized benchmarks in the field. These datasets capture a diverse range of real-world scenarios,including human activity recognition, sensor data interpretation, and complex system monitoring, and thus provide a multidimensional foundation for evaluating time series models.

We evaluate the classification performance of TimeFT against state-of-the-art tokenization methods on all UEA datasets. We run and evaluate several recent tokenization methods for comparison. As shown in Table~\ref{tab:classification_all_uea}, TimeFT achieves the highest average accuracy and achieves the best or second-best result in a majority of datasets. This strong performance improvement is due to TimeFT's almost distortion‑free tokenization ability, which faithfully preserves discriminative temporal patterns that are often corrupted by tokenizers.

\begin{table*}[htbp]
    \centering
    \caption{We evaluate classification tasks on all UEA datasets. Best results are in \best{bold} and second best are \second{underlined}.}
    \label{tab:classification_all_uea}
    
    \renewcommand{\arraystretch}{1.1}
    \small
    \setlength{\tabcolsep}{5pt}

    \resizebox{\textwidth}{!}{
    \begin{tabular}{l ccccccccc}
        \toprule
        Datasets & \textbf{TimeFT} & EntroPE & TimeMosaic & PhaseFormer & PatchMLP & GPT4TS & PatchTST & TimesNet & MLP \\
        & (\textbf{Ours}) & (2026) & (2026) & (2026) & (2025) & (2024) & (2023) & (2023) & - \\
        \midrule
        ArticularyWordRecognition & \second{97.50} & 97.00 & 96.50 & 95.00 & 97.20 & 93.33 & \best{97.67} & 96.18 & 97.33 \\
        AtrialFibrillation      & \second{53.00} & 52.00 & 50.50 & 48.00 & \second{53.00} & 33.33 & \best{53.33} & 33.33 & 46.66 \\
        BasicMotions            & \second{98.00} & \second{98.00} & 97.00 & 95.50 & \second{98.00} & 92.50 & 92.50 & \best{100.00} & 85.00 \\
        Cricket                 & \second{91.00} & 90.00 & 89.00 & 87.00 & 90.50 & 8.33 & 84.72 & 87.50 & \best{91.67} \\
        Epilepsy                & \second{85.00} & 84.00 & 82.50 & 80.00 & 84.50 & \best{85.51} & 65.94 & 78.13 & 60.14 \\
        EthanolConcentration    & \best{31.94} & \second{29.27} & 28.90 & 26.13 & 29.65 & 25.48 & 28.90 & 27.73 & 28.09 \\
        FaceDetection           & \best{69.48} & \second{69.03} & 67.95 & 64.65 & 67.14 & 65.58 & 68.96 & 67.47 & 67.42 \\
        FingerMovements         & \second{63.00} & 62.50 & 61.00 & 59.50 & \second{63.00} & 57.00 & 62.00 & 59.38 & \best{64.00} \\
        HandMovementDirection   & \second{58.00} & 56.00 & 54.50 & 52.00 & 57.00 & 18.92 & \best{58.11} & 50.00 & \best{58.11} \\
        Handwriting             & 24.70 & 23.31 & \second{24.80} & 23.16 & 23.87 & 3.76 & \second{26.00} & \best{26.18} & 22.47 \\
        Heartbeat               & \best{77.06} & \second{76.64} & 73.34 & 76.10 & 73.54 & 36.59 & 76.59 & 74.48 & 73.17 \\
        InsectWingbeat          & \best{11.00} & \second{10.50} & 10.00 & 9.50 & 10.20 & 10.00 & 10.00 & 10.00 & 10.00 \\
        JapaneseVowels          & \best{98.87} & \second{98.76} & 97.23 & 95.57 & 97.49 & 98.11 & 98.65 & 97.83 & 97.84 \\
        Libras                  & \best{82.00} & \second{80.00} & 78.50 & 76.00 & 80.50 & 79.44 & 81.11 & 77.84 & 73.33 \\
        LSST                    & \best{68.50} & 66.00 & 64.00 & 61.50 & \second{66.50} & 46.39 & 67.80 & 59.21 & 35.77 \\
        MotorImagery            & \best{62.00} & 60.00 & 58.50 & 56.00 & 60.50 & 50.00 & 61.00 & 51.04 & 61.00 \\
        NATOPS                  & \best{97.50} & 95.50 & 94.00 & 92.00 & 96.00 & 91.67 & \second{96.67} & 81.82 & 93.89 \\
        PEMS-SF                 & \best{88.86} & 87.12 & 87.15 & 86.80 & 78.85 & 87.28 & \second{88.44} & 88.13 & 82.08 \\
        PenDigits               & \best{99.80} & 98.50 & 98.00 & 97.00 & 99.00 & 97.74 & \second{99.23} & 98.19 & 92.94 \\
        PhonemeSpectra          & \best{19.00} & 17.00 & 15.50 & 14.00 & \second{17.50} & 3.01 & 11.69 & 18.24 & 7.10 \\
        RacketSports            & \best{85.00} & 83.00 & 81.50 & 79.50 & \second{83.50} & 76.97 & 84.21 & 82.64 & 78.95 \\
        SelfRegulationSCP1      & 89.05 & 88.58 & 89.35 & 86.03 & 89.36 & \best{91.47} & \second{89.76} & 77.43 & 88.40 \\
        SelfRegulationSCP2      & \best{59.44} & \second{57.42} & 52.07 & 56.46 & 53.62 & 51.67 & 54.44 & 52.84 & 51.67 \\
        SpokenArabicDigits      & \best{99.71} & 98.75 & 98.73 & 97.60 & 96.99 & 99.36 & \second{99.68} & 98.36 & 96.68 \\
        StandWalkJump           & \best{61.00} & 58.50 & 57.00 & 55.00 & 59.00 & 33.33 & \second{60.00} & 53.33 & \second{60.00} \\
        UWaveGestureLibrary     & 83.75 & \best{84.38} & 81.88 & 76.81 & 78.63 & \best{84.38} & 80.00 & 83.13 & 81.88 \\
        \midrule
        \textbf{Average}        & \best{71.31} & \second{70.07} & 68.82 & 67.19 & 69.42 & 58.51 & 69.13 & 66.55 & 65.60 \\
        \textbf{1st}            & \best{16} & \second{1} & 0 & 0 & 0 & 3 & 3 & 2 & 3 \\
        \bottomrule
    \end{tabular}
    }
\end{table*}

\begin{table}[htbp]
    \centering
    \caption{Summary of datasets used for anomaly detection evaluation. Dim denotes the variate number of each dataset. Dataset sizes represent the number of temporal observations in each partition (Training, Validation, Test). Frequency denotes the sampling interval of time points.}
    \label{tab:anomaly_datasets}
    \small
    \begin{tabular}{lcccl}
        \toprule
        Dataset & Dim & \makecell{Dataset Size \\ (Train, Val, Test)} & Length & Domain Information \\
        \midrule
        SMD   & 38 & (566724, 141681, 708420)  & 100 & Server Machine \\
        MSL   & 55 & (44653, 11664, 73729)    & 100 & Spacecraft \\
        SMAP  & 25 & (108146, 27037, 427617)  & 100 & Spacecraft \\
        SWaT  & 51 & (396000, 99000, 449919)  & 100 & Infrastructure \\
        PSM   & 25 & (105984, 26497, 87841)   & 100 & Server Machine \\
        \bottomrule
    \end{tabular}
\end{table}

\begin{table*}[htbp]
    \centering
    \caption{For anomaly detection task, we evaluate F1-score (\%) on five benchmark datasets. Best results are in \best{bold} and second best are \second{underlined}.}
    \label{tab:anomaly_detection_all}
    
    \renewcommand{\arraystretch}{1.1}
    \small
    \setlength{\tabcolsep}{5pt}

    \resizebox{\textwidth}{!}{
    \begin{tabular}{l cccccccccc}
        \toprule
        Models & \textbf{TimeFT} & EntroPE & TimeMosaic & Phaseformer & LightTS & TimesNet & PatchTST & DLinear & ETSformer & FEDformer \\
        & (\textbf{Ours}) & (2026) & (2026) & (2026) & (2024) & (2023) & (2023) & (2023) & (2023) & (2022) \\
        \midrule
        SMD  & \best{85.52} & 84.75 & 84.25 & 84.85 & 82.53 & 84.61 & 84.62 & 77.10 & 83.13 & \second{85.08} \\
        MSL  & 82.95 & 81.95 & 78.42 & 79.15 & 78.95 & 81.84 & 78.70 & \second{84.88} & \best{85.03} & 78.57 \\
        SMAP & \best{71.80} & 69.80 & 68.10 & 68.75 & 69.21 & 69.39 & 68.82 & 69.26 & 69.50 & \second{70.76} \\
        SWaT & \best{93.65} & 93.10 & 85.90 & 86.12 & \second{93.33} & 93.02 & 85.72 & 87.52 & 84.91 & 93.19 \\
        PSM  & 96.68 & \second{97.28} & 95.85 & 96.20 & 97.15 & \best{97.34} & 96.08 & 93.55 & 91.76 & 97.23 \\
        \midrule
        \textbf{Average} & \best{86.12} & \second{85.38} & 82.50 & 83.01 & 84.23 & 85.24 & 82.79 & 82.46 & 82.87 & 84.97 \\
        \bottomrule
    \end{tabular}
    }
\end{table*}

\subsection{Anomaly Detection}
\label{app:anomaly_detection_all}

For anomaly detection, we adopt five widely recognized datasets that span diverse application domains, allowing a thorough evaluation of detection capabilities under varied conditions. The detailed characteristics of these datasets are presented in Table~\ref{tab:anomaly_datasets}.

\begin{itemize}
    \item \textbf{SMD (Server Machine Dataset)}: Comprising multivariate server telemetry, SMD targets the identification of anomalous server behavior indicative of failures or security breaches.
    \item \textbf{MSL (Mars Science Laboratory)}: Originating from the Mars rover, this telemetry dataset focuses on space exploration, where detecting anomalies is critical for diagnosing spacecraft subsystem faults.
    \item \textbf{SMAP (Soil Moisture Active Passive)}: Containing satellite soil moisture measurements, SMAP supports environmental monitoring; detected irregularities can reveal critical changes in environmental and agricultural conditions.
    \item \textbf{SWaT (Secure Water Treatment)}: Derived from a water treatment testbed, SWaT simulates plant operations and is essential for ensuring the safety and integrity of water treatment processes through anomaly detection.
    \item \textbf{PSM (Pump Sensor Monitoring)}: Collected from industrial pump sensors, PSM enables anomaly detection to flag equipment malfunctions and maintenance needs, thereby helping to prevent industrial incidents.
\end{itemize}

We evaluate the anomaly detection capability of TimeFT against state-of-the-art tokenization baselines and classic time series models on five standard benchmarks. We run and evaluate several recent tokenization methods for comparison. As shown in Table~\ref{tab:anomaly_detection_all}, TimeFT achieves the highest average F1‑score of 86.12\%. TimeFT preserves the integrity of anomalous patterns and shows excellent detection performance. Experiments on diverse tasks and extensive benchmarks demonstrate that TimeFT exhibits superior performance compared to existing tokenization methods.

\subsection{Robustness Analysis under Noisy Conditions}
\label{app:Robustness Analysis under Noisy Conditions}

\paragraph{Bandwidth Estimation.} Since deep learning models typically require fixed-dimensional input, time-series models often employ an input window to perform tasks such as forecasting, specifically a hyperparameter of length $N$, which provides a fixed number of data points for DSP analysis. Furthermore, the stride $S$ is usually a tokenization hyperparameter used to control the number of tokens. Therefore, the critical bandwidth $B_N$ is independent of noise. It is entirely derived from the distortion conditions of the existing tokenization paradigm.

\paragraph{Experiments under Noisy Conditions.} The results of robustness analysis under noisy conditions are as shown in Table~\ref{tab: noise_robustness}. TimeFT strictly enforces the alias-free boundary condition. The frequency-domain blocking operation applies ideal bandpass filters, which mathematically zero out all out-of-band noise components before any downsampling. Each sub-band is then independently shifted to baseband and satisfies the Nyquist criterion. Consequently, the subsequent sampling step will not incur aliasing. High-frequency noise is strictly confined to the corresponding high frequency. This gives TimeFT excellent noise robustness.

\begin{table*}[htbp]
    \centering
    \caption{Long-term forecasting results on ETTm1 under Gaussian noise. $\sigma$ is the standard deviation of the noise.}
    \label{tab: noise_robustness}
    
    \renewcommand{\arraystretch}{1.1}
    \small
    \setlength{\tabcolsep}{7pt}
    
    \resizebox{\textwidth}{!}{
    \begin{tabular}{l cc cc cc cc cc cc}
        \toprule
        Noise ($\sigma$) & \multicolumn{2}{c}{0} & \multicolumn{2}{c}{0.2} & \multicolumn{2}{c}{0.4} & \multicolumn{2}{c}{0.6} & \multicolumn{2}{c}{0.8} & \multicolumn{2}{c}{1.0} \\
        \cmidrule(lr){2-3} \cmidrule(lr){4-5} \cmidrule(lr){6-7} \cmidrule(lr){8-9} \cmidrule(lr){10-11} \cmidrule(lr){12-13}
        Metric & MSE & MAE & MSE & MAE & MSE & MAE & MSE & MAE & MSE & MAE & MSE & MAE \\
        \midrule
        PatchTST        & 0.321 & 0.361 & 0.343 & 0.377 & 0.368 & 0.394 & 0.397 & 0.419 & 0.440 & 0.447 & 0.483 & 0.485 \\
        PhaseFormer     & 0.312 & 0.352 & 0.330 & 0.364 & 0.350 & 0.381 & 0.382 & 0.395 & 0.414 & 0.426 & 0.458 & 0.449 \\
        \rowcolor{blue!7}
        \textbf{TimeFT (ours)} & \textbf{0.290} & \textbf{0.339} & \textbf{0.302} & \textbf{0.347} & \textbf{0.318} & \textbf{0.355} & \textbf{0.335} & \textbf{0.374} & \textbf{0.364} & \textbf{0.387} & \textbf{0.391} & \textbf{0.412} \\
        \bottomrule
    \end{tabular}
    }
\end{table*}

\begin{table*}[htbp]
    \centering
    \caption{Degradation Rate of the long-term forecasting task under Gaussian noise. Metric: MSE.}
    \label{tab: noise_degradation}
    
    \renewcommand{\arraystretch}{1.1}
    \small
    \setlength{\tabcolsep}{7pt}
    
    \resizebox{0.8\textwidth}{!}{
    \begin{tabular}{l cccccc}
        \toprule
        Noise ($\sigma$) & 0 & 0.2 & 0.4 & 0.6 & 0.8 & 1.0 \\
        \midrule
        PatchTST        & 0 & 6.85\% & 14.64\% & 23.68\% & 37.07\% & 50.47\% \\
        PhaseFormer     & 0 & 5.77\% & 12.18\% & 22.44\% & 32.69\% & 46.79\% \\
        \rowcolor{blue!7}
        \textbf{TimeFT (ours)} & \textbf{0} & \textbf{4.14\%} & \textbf{9.66\%} & \textbf{15.52\%} & \textbf{25.52\%} & \textbf{34.83\%} \\
        \bottomrule
    \end{tabular}
    }
\end{table*}

\paragraph{Qualitative Analysis.} It can be observed that there is a positive correlation between the level of distortion and the degradation in downstreaming tasks performance, which validates the argument presented in our paper. For TimeFT, the frequency-domain blocking operation applies ideal bandpass filters, which mathematically zero out all out-of-band noise components before any downsampling. Each sub-band is then independently shifted to baseband and satisfies the Nyquist criterion.

Consequently, the subsequent sampling step incurs zero aliasing. High-frequency noise is strictly confined to the corresponding high frequency, while the low frequency remains completely immune to contamination. This band-decoupled robustness is absent in both patch and phase tokenization.

\paragraph{Quantitative Analysis.} To measure the sensitivity of each tokenization method to spectral distortion, we define the Degradation Rate $D(\sigma)$ as the relative increase in MSE from the noise-free baseline $\sigma=0$.

$$D(\sigma)=(MSE(\sigma)-MSE(0))/MSE(0)\times 100 \%$$

As shown in Table~\ref{tab: noise_degradation}, the degradation rates reveal a clear hierarchy. TimeFT exhibits the lowest performance deterioration across all noise levels, with a substantially smaller average slope compared to PatchTST and PhaseFormer. This quantitatively demonstrates that TimeFT's spectral-distortion-free design confers greater robustness to input corruption. Moreover, the performance gap widens slowly with increasing noise, suggesting that TimeFT exhibits superior anti-distortion performance compared to prior tokenization methods.

\subsection{Experiments on More Architectures}
\label{app:Experiments on More Architectures}

As shown in Table~\ref{tab: Experiments on More Architectures}, the experimental results of more diverse and state-of-the-art architectures demonstrate that TimeFT’s near-lossless tokenization is readily transferable to different architectures, delivering immediate gains without modifying the base model design.

\begin{table*}[htbp]
    \centering
    \caption{Plug-in experiments of the long-term forecasting task to show the general capability. All settings are the same as Table~\ref{tab:plug_in}.}
    \label{tab: Experiments on More Architectures}
    
    \renewcommand{\arraystretch}{1.1}
    \small
    \setlength{\tabcolsep}{5pt}
    
    \resizebox{\textwidth}{!}{
    \begin{tabular}{l cc cc cc cc cc cc cc}
        \toprule
        Dataset & \multicolumn{2}{c}{ETTm1} & \multicolumn{2}{c}{ETTm2} & \multicolumn{2}{c}{ETTh1} & \multicolumn{2}{c}{ETTh2} & \multicolumn{2}{c}{Weather} & \multicolumn{2}{c}{Electricity} & \multicolumn{2}{c}{Traffic} \\
        \cmidrule(lr){2-3} \cmidrule(lr){4-5} \cmidrule(lr){6-7} \cmidrule(lr){8-9} \cmidrule(lr){10-11} \cmidrule(lr){12-13} \cmidrule(lr){14-15}
        Metric & MSE & MAE & MSE & MAE & MSE & MAE & MSE & MAE & MSE & MAE & MSE & MAE & MSE & MAE \\
        \midrule
        PatchTST            & 0.321 & 0.361 & 0.181 & 0.270 & 0.403 & 0.423 & 0.324 & 0.374 & 0.155 & 0.204 & 0.142 & 0.245 & 0.403 & 0.285 \\
        \rowcolor{blue!7}
        \textbf{+TimeFT}    & \textbf{0.290} & \textbf{0.339} & \textbf{0.162} & \textbf{0.244} & \textbf{0.376} & \textbf{0.392} & \textbf{0.297} & \textbf{0.339} & \textbf{0.147} & \textbf{0.194} & \textbf{0.139} & \textbf{0.234} & \textbf{0.395} & \textbf{0.279} \\
        $\Delta$\% Improvement & 9.7\% & 6.1\% & 10.5\% & 9.6\% & 6.7\% & 7.3\% & 8.3\% & 9.4\% & 5.2\% & 4.9\% & 2.1\% & 4.5\% & 2.0\% & 2.1\% \\
        \midrule
        PhaseFormer         & 0.312 & 0.352 & 0.174 & 0.261 & 0.389 & 0.411 & 0.309 & 0.361 & 0.152 & 0.201 & 0.145 & 0.248 & 0.418 & 0.289 \\
        \rowcolor{blue!7}
        \textbf{+TimeFT}    & \textbf{0.293} & \textbf{0.341} & \textbf{0.161} & \textbf{0.249} & \textbf{0.379} & \textbf{0.397} & \textbf{0.299} & \textbf{0.343} & \textbf{0.149} & \textbf{0.193} & \textbf{0.143} & \textbf{0.242} & \textbf{0.406} & \textbf{0.278} \\
        $\Delta$\% Improvement & 6.1\% & 3.1\% & 7.5\% & 4.6\% & 2.6\% & 3.4\% & 3.2\% & 5.0\% & 1.9\% & 3.9\% & 1.4\% & 2.4\% & 2.9\% & 3.8\% \\
        \midrule
        PatchMLP            & 0.305 & 0.357 & 0.179 & 0.265 & 0.411 & 0.426 & 0.311 & 0.371 & 0.153 & 0.206 & 0.165 & 0.271 & 0.486 & 0.361 \\
        \rowcolor{blue!7}
        \textbf{+TimeFT}    & \textbf{0.286} & \textbf{0.339} & \textbf{0.165} & \textbf{0.250} & \textbf{0.390} & \textbf{0.405} & \textbf{0.294} & \textbf{0.348} & \textbf{0.147} & \textbf{0.197} & \textbf{0.160} & \textbf{0.264} & \textbf{0.472} & \textbf{0.349} \\
        $\Delta$\% Improvement & 6.2\% & 5.0\% & 7.8\% & 5.7\% & 5.1\% & 4.9\% & 5.5\% & 6.2\% & 3.9\% & 4.4\% & 3.0\% & 2.6\% & 2.9\% & 3.3\% \\
        \midrule
        iTransformer        & 0.327 & 0.371 & 0.187 & 0.271 & 0.403 & 0.417 & 0.323 & 0.369 & 0.160 & 0.209 & 0.145 & 0.244 & 0.422 & 0.310 \\
        \rowcolor{blue!7}
        \textbf{+TimeFT}    & \textbf{0.309} & \textbf{0.355} & \textbf{0.174} & \textbf{0.257} & \textbf{0.390} & \textbf{0.403} & \textbf{0.312} & \textbf{0.354} & \textbf{0.156} & \textbf{0.202} & \textbf{0.142} & \textbf{0.239} & \textbf{0.412} & \textbf{0.302} \\
        $\Delta$\% Improvement & 5.5\% & 4.3\% & 7.0\% & 5.2\% & 3.2\% & 3.4\% & 3.4\% & 4.1\% & 2.5\% & 3.3\% & 2.1\% & 2.0\% & 2.4\% & 2.6\% \\
        \bottomrule
    \end{tabular}
    }
\end{table*}

\subsection{Experiments on Different Random Seeds}
\label{app:Experiments on Different Random Seeds}

As shown in Table~\ref{tab: experiments on random_seeds}, it can be observed that the robustness of different methods under multiple random seeds is basically the same.

\begin{table*}[htbp]
    \centering
    \caption{Long-term forecasting results under different random seeds to show the effect of random seeds.}
    \label{tab: experiments on random_seeds}
    
    \renewcommand{\arraystretch}{1.1}
    \small
    \setlength{\tabcolsep}{5pt}
    
    \resizebox{\textwidth}{!}{
    \begin{tabular}{l cc cc cc cc}
        \toprule
        Models & \multicolumn{2}{c}{TimeFT (ours)} & \multicolumn{2}{c}{EntroPE} & \multicolumn{2}{c}{TimeMosaic} & \multicolumn{2}{c}{PhaseFormer} \\
        \cmidrule(lr){2-3} \cmidrule(lr){4-5} \cmidrule(lr){6-7} \cmidrule(lr){8-9}
        Metric & MSE & MAE & MSE & MAE & MSE & MAE & MSE & MAE \\
        \midrule
        ETTm1       & \textbf{0.348$\pm$0.005} & \textbf{0.373$\pm$0.004} & 0.369$\pm$0.007 & 0.383$\pm$0.006 & 0.360$\pm$0.004 & 0.377$\pm$0.008 & 0.364$\pm$0.006 & 0.392$\pm$0.005 \\
        ETTm2       & 0.255$\pm$0.003 & \textbf{0.307$\pm$0.004} & 0.264$\pm$0.006 & 0.320$\pm$0.005 & \textbf{0.256$\pm$0.003} & 0.310$\pm$0.007 & 0.268$\pm$0.009 & 0.324$\pm$0.006 \\
        ETTh1       & 0.417$\pm$0.006 & \textbf{0.423$\pm$0.005} & 0.424$\pm$0.008 & 0.433$\pm$0.004 & \textbf{0.412$\pm$0.005} & \textbf{0.423$\pm$0.007} & 0.489$\pm$0.009 & 0.475$\pm$0.004 \\
        ETTh2       & \textbf{0.358$\pm$0.004} & \textbf{0.390$\pm$0.007} & 0.374$\pm$0.008 & 0.401$\pm$0.005 & 0.363$\pm$0.003 & 0.393$\pm$0.006 & 0.384$\pm$0.006 & 0.417$\pm$0.005 \\
        Weather     & 0.232$\pm$0.003 & 0.261$\pm$0.005 & 0.237$\pm$0.005 & 0.264$\pm$0.007 & \textbf{0.228$\pm$0.004} & \textbf{0.256$\pm$0.008} & 0.233$\pm$0.005 & 0.268$\pm$0.009 \\
        Electricity & \textbf{0.165$\pm$0.004} & \textbf{0.258$\pm$0.006} & 0.177$\pm$0.007 & 0.271$\pm$0.005 & 0.170$\pm$0.003 & 0.263$\pm$0.008 & 0.177$\pm$0.007 & 0.273$\pm$0.008 \\
        Traffic     & \textbf{0.420$\pm$0.008} & 0.290$\pm$0.006 & 0.444$\pm$0.009 & 0.300$\pm$0.005 & 0.434$\pm$0.006 & \textbf{0.287$\pm$0.004} & 0.441$\pm$0.003 & 0.305$\pm$0.006 \\
        \bottomrule
    \end{tabular}
    }
\end{table*}

\subsection{Non-stationary Data Analysis}
\label{app:Non-stationary Data Analysis}

TimeFT is orthogonal to non-stationary property, just as RevIN normalization is orthogonal to PatchTST.

RFFT still holds for non-stationary data. As one of the foundations of TimeFT, RFFT and the Fourier transform family are indeed affected by abrupt shifts. This is due to the fact that the Fourier transform operates as a convolution across the entire time domain on a time series of length N. But as basis transformations, the FFT family does not need the stationarity assumption. For non-stationary signals or abrupt shifts, the Fourier transform will just simply produce a broadband spectrum containing high-frequency components associated with the sharp transitions.

However, our TimeFT addresses the distortion problem in tokenization, which is an issue independent of the abrupt drifts in the raw data. This orthogonality arises from the fact that the characteristics of the raw data and the tokenization process are two distinct types of problems. This is similar to how PatchTST's patch tokenization also cannot resolve abrupt shifts in the original data. For data types that suffer from abrupt shift issues, researchers typically adopt additional methods, such as RevIN normalization~\cite{Revin}, to suppress such abrupt changes.

Therefore, TimeFT is not designed exclusively for periodic or stationary data. Rather, it addresses the distortion problem inherent in the general tokenization paradigm, which is orthogonal to the non-stationarity of the raw data. If handling highly non-stationary data types is required, we can introduce dedicated modules, such as RevIN normalization, as many prior works have done.

\subsection{Settings for Other Experiments}
\label{app:Setting in Other Experiments}

\paragraph{Time Complexity Analysis.} In the time complexity analysis experiment, we ensure that all experimental parameters are identical. The test is conducted on the ETTm1 dataset, with an input length of 320 and a prediction length of 336.

\paragraph{Plu-in Experiments.} In the plug-in experiments, we replace the tokenization modules of different methods with TimeFT, leaving the other components unchanged. For each method, we carefully tune the hyperparameters both before and after the substitution to ensure strong performance.

\paragraph{Ablation Study.} In the ablation study, removing a Stage means removing that module from the pipeline while keeping the rest unchanged. The evaluation is conducted on the long-term forecasting task, with an input length of 320 and a prediction length of 96.

\paragraph{Hyperparameter Sensitive Analysis.} Generally speaking, the only hyperparameter of TimeFT is the stride $S$, and all other parameters are mathematically derived. Moreover, as a tokenizer, TimeFT is closely related to input length $N$. Therefore, in the hyperparameter sensitivity experiment, we compare the effects of different strides $S$ and input lengths $N$ on model performance. For each setting, we tune other hyperparameters because the optimal configuration changes accordingly.

\subsection{Variation in Model Parameters}
\label{app:Variation in Model Parameter Count}

When analyzing the impact of TimeFT on the model parameters, we observed that hyperparameters significantly affect this metric. Therefore, we keep all hyperparameters identical across experiments and compare only the variation in the embedding module parameters and the total parameter size, since the tokenizer mainly influences the number of embedding parameters. "Total of Model" denotes the total parameter count of the model under identical hyperparameter settings. As shown in Table~\ref{tab:Variation in Model Parameter Count}, although TimeFT increases the number of embedding parameters due to the change in input dimensionality, this change is negligible compared to the total number of parameters.

\begin{table}[htbp]
    \centering
    \caption{Variation in Model Parameter Count.}
    \label{tab:Variation in Model Parameter Count}
    \small
    \begin{tabular}{lcccl}
        \toprule
        Layers & TimeFT Embedding Layer & PatchTST Embedding Layer & Total of Model\\
        \midrule
        Params (MB) & 0.699 & 0.031 & 50.337\\
        \bottomrule
    \end{tabular}
\end{table}

\subsection{Reconstruction Performance of Different Token Types}
\label{app:Reconstruction Performance across Different Token Types}

To intuitively illustrate the degree of distortion for different types of tokens, we compare the mean absolute error between the reconstructed sequences and the original sequences across different datasets. This experiment can be viewed as a statistical summary of Figure~\ref{fig: figure2} and the visualization section~\ref{app: visualization}. We ensure that all hyperparameters remain identical in the experiments. As shown in Table~\ref{tab: Reconstruction Performance}, TimeFT tokens are almost distortion-free and reduce more than 80\% reconstruction error in many datasets. The reconstruction methods are provided in Section~\ref{app: Reconstruction_method}.

\begin{table*}[htbp]
    \centering
    \caption{Reconstruction performance of different token types across different datasets. The percentage represents the MAE reduction of TimeFT tokens relative to the best one of others. Metric: MAE.}
    \label{tab: Reconstruction Performance}
    
    \renewcommand{\arraystretch}{1.1}
    \small
    \setlength{\tabcolsep}{7pt}
    \resizebox{0.9\textwidth}{!}{
    \begin{tabular}{l ccccccc}
        \toprule
        Datasets & ETTm1 & ETTm2 & ETTh1 & ETTh2 & Weather & Electricity & Traffic \\
        \midrule
        Patch   &  0.768 & 0.747 & 0.755 & 0.722 & 0.726 & 0.821 & 0.719\\
        Phase   &  0.449 & 0.525 & 0.823 & 0.714 & 0.280 & 0.998 & 0.986\\
        \rowcolor{blue!7}
        \textbf{TimeFT}  & \textbf{0.102} & \textbf{0.115} & \textbf{0.144} & \textbf{0.133} & \textbf{0.079} & \textbf{0.122} & \textbf{0.138} \\
        \midrule
        Reduction  &  77.3\% & 78.1\% & 80.9\% & 81.4\% & 71.8\% & 85.1\% & 80.8\% \\
        \bottomrule
    \end{tabular}
    }
\end{table*}

\section{Limitations and Social Impacts}
\label{app: limitations}

\paragraph{Limitations.} Our work carries two limitations. Owing to the substantial computational expense involved in pre-training large-scale time series foundation models, our experimental evaluation does not encompass pre-trained TSFMs. Consequently, the behavior and potential benefits of TimeFT in a pre-training and fine-tuning pipeline remain unverified, and investigating its effectiveness under TSFMs regimes is left to future work. Furthermore, although tokenization has become a prevailing paradigm in modern time series analysis, our method intrinsically assumes a tokenization-based processing flow and therefore yields no direct improvement for models that operate directly on raw sequences without any tokenization step.

\paragraph{Social Impacts.} As a fundamental component in time series modeling, our work can enhance the performance of a wide range of time series methods on various real-world tasks, such as forecasting, anomaly detection, and classification. It can serve as a foundational module in improving the application of time series models in fields such as healthcare, transportation, and economics.

\section{Visualization of TimeFT}
\label{app: visualization}
\begin{figure*}[htbp]
    \centering
    \includegraphics[width=\textwidth]{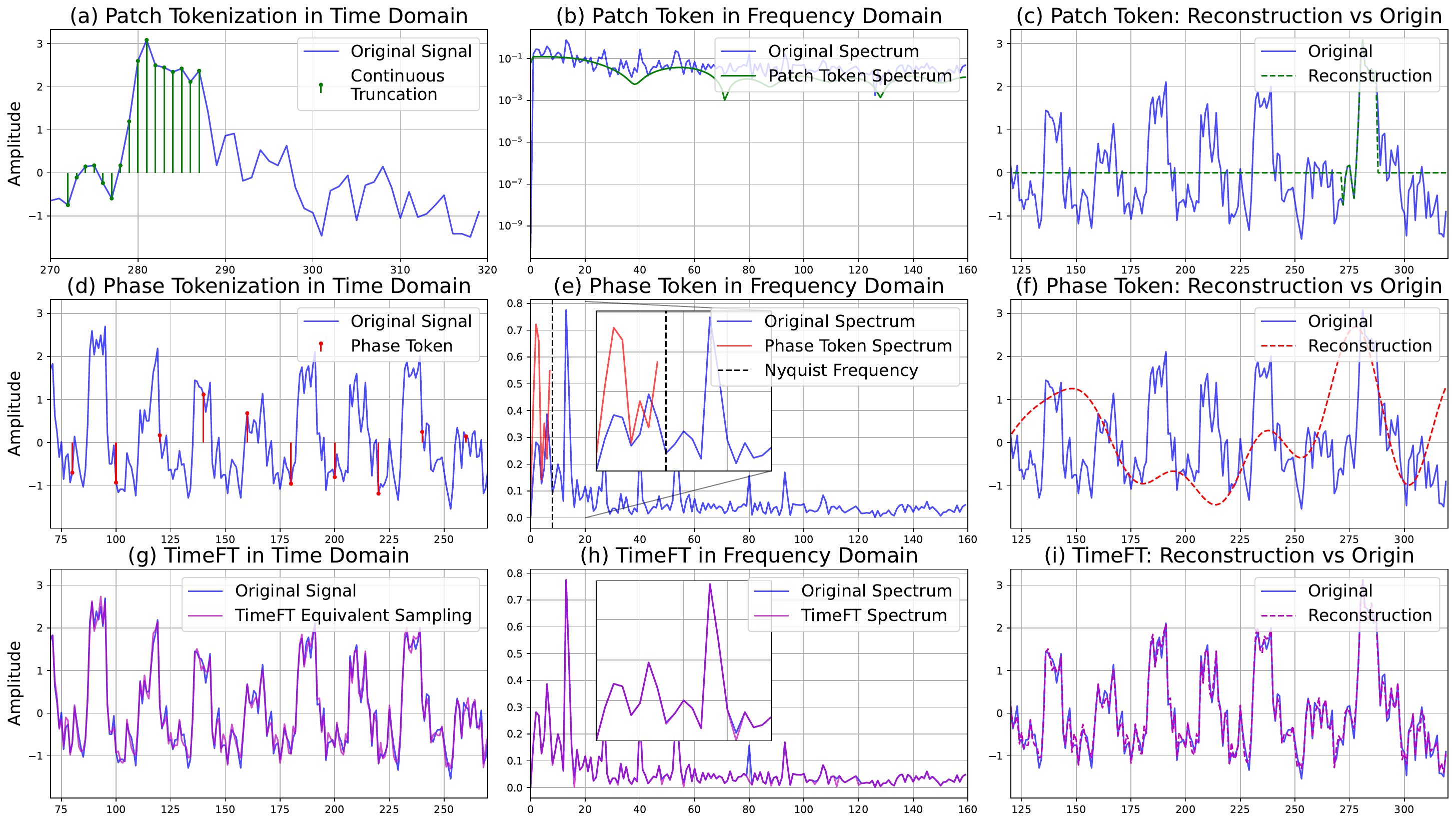} 
    \caption{The visualization of tokenization methods across the time domain, frequency domain and reconstruction vs origin in ETTh1 dataset.}
    \label{fig: visualization_ETTh1}
    \vspace{-10pt}
\end{figure*}

\begin{figure*}[htbp]
    \centering
    \includegraphics[width=\textwidth]{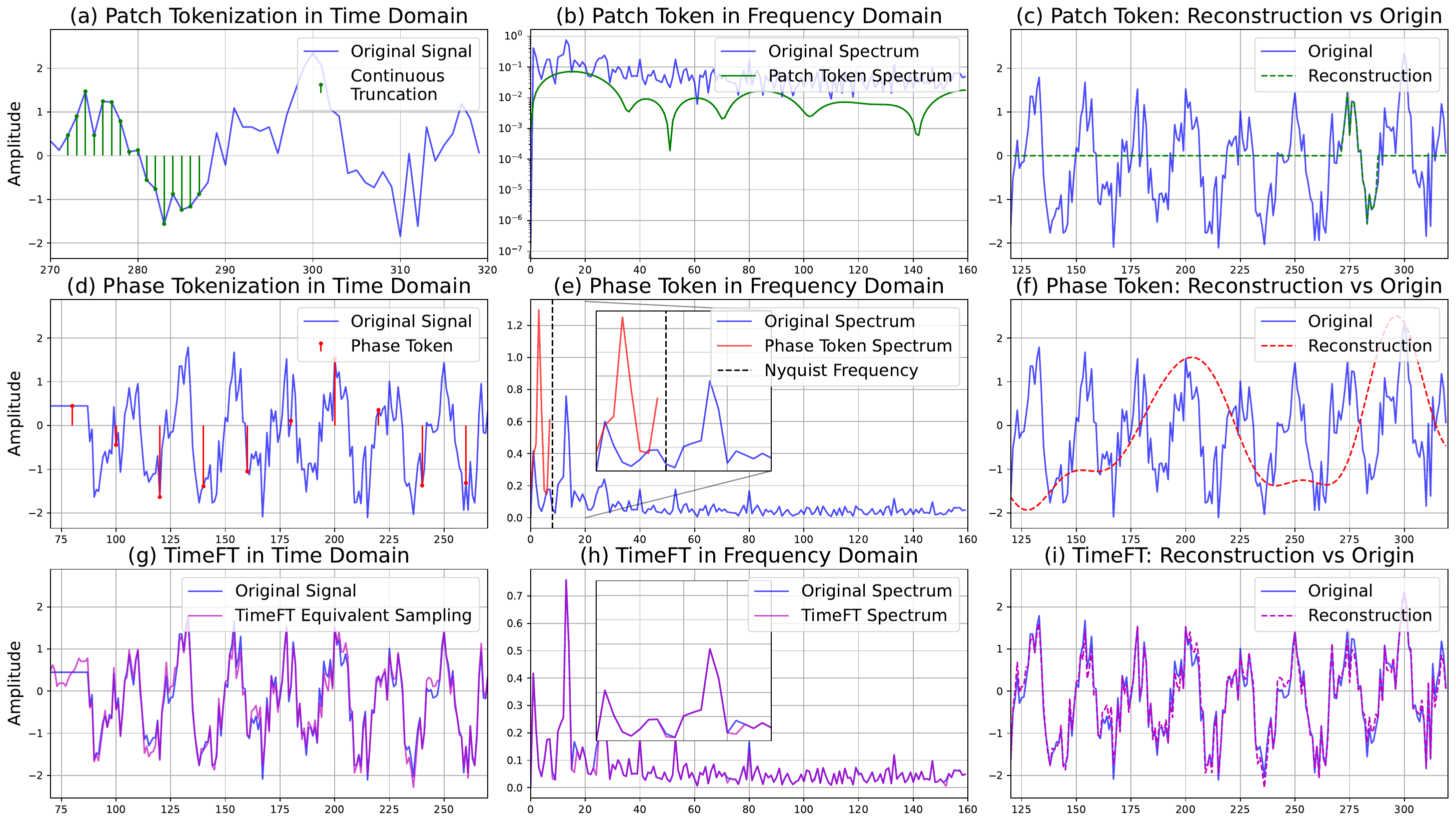} 
    \caption{The visualization of tokenization methods across the time domain, frequency domain and reconstruction vs origin in ETTh2 dataset.}
    \label{fig: visualization_ETTh2}
    \vspace{-10pt}
\end{figure*}

\begin{figure*}[htbp]
    \centering
    \includegraphics[width=\textwidth]{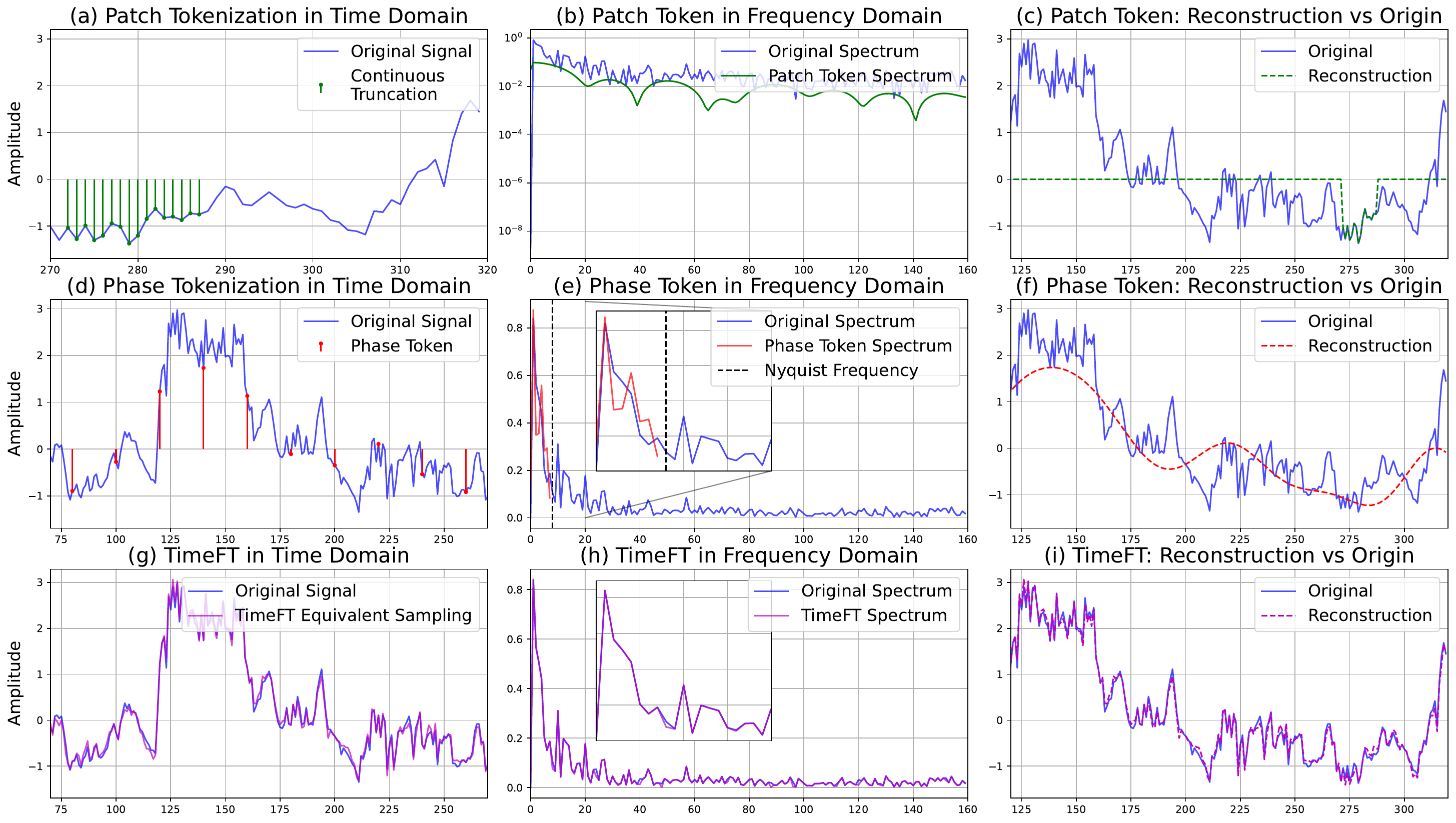} 
    \caption{The visualization of tokenization methods across the time domain, frequency domain and reconstruction vs origin in ETTm1 dataset.}
    \label{fig: visualization_ETTm1}
    \vspace{-10pt}
\end{figure*}

\begin{figure*}[htbp]
    \centering
    \includegraphics[width=\textwidth]{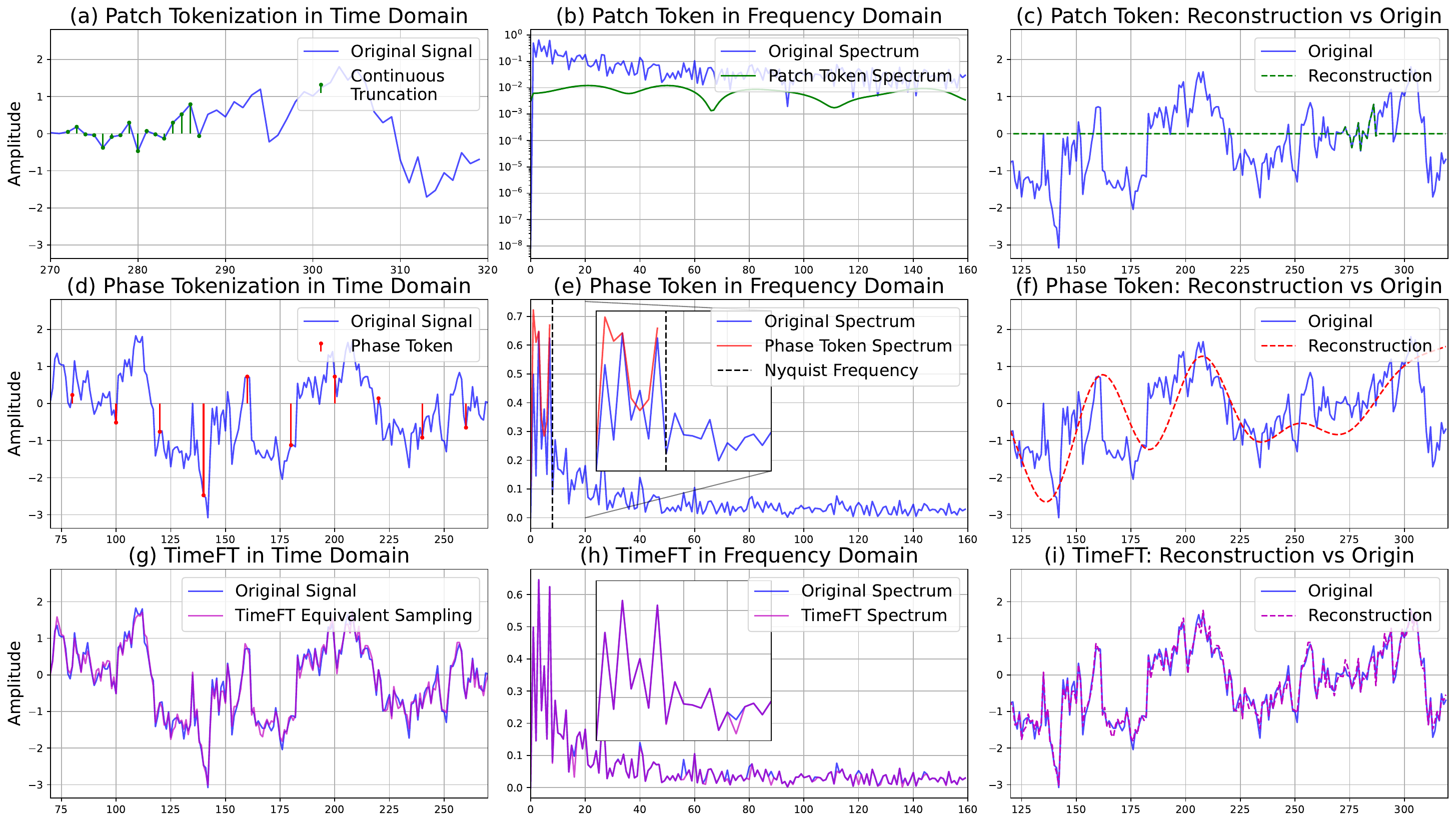} 
    \caption{The visualization of tokenization methods across the time domain, frequency domain and reconstruction vs origin in ETTm2 dataset.}
    \label{fig: visualization_ETTm2}
    \vspace{-10pt}
\end{figure*}

\begin{figure*}[htbp]
    \centering
    \includegraphics[width=\textwidth]{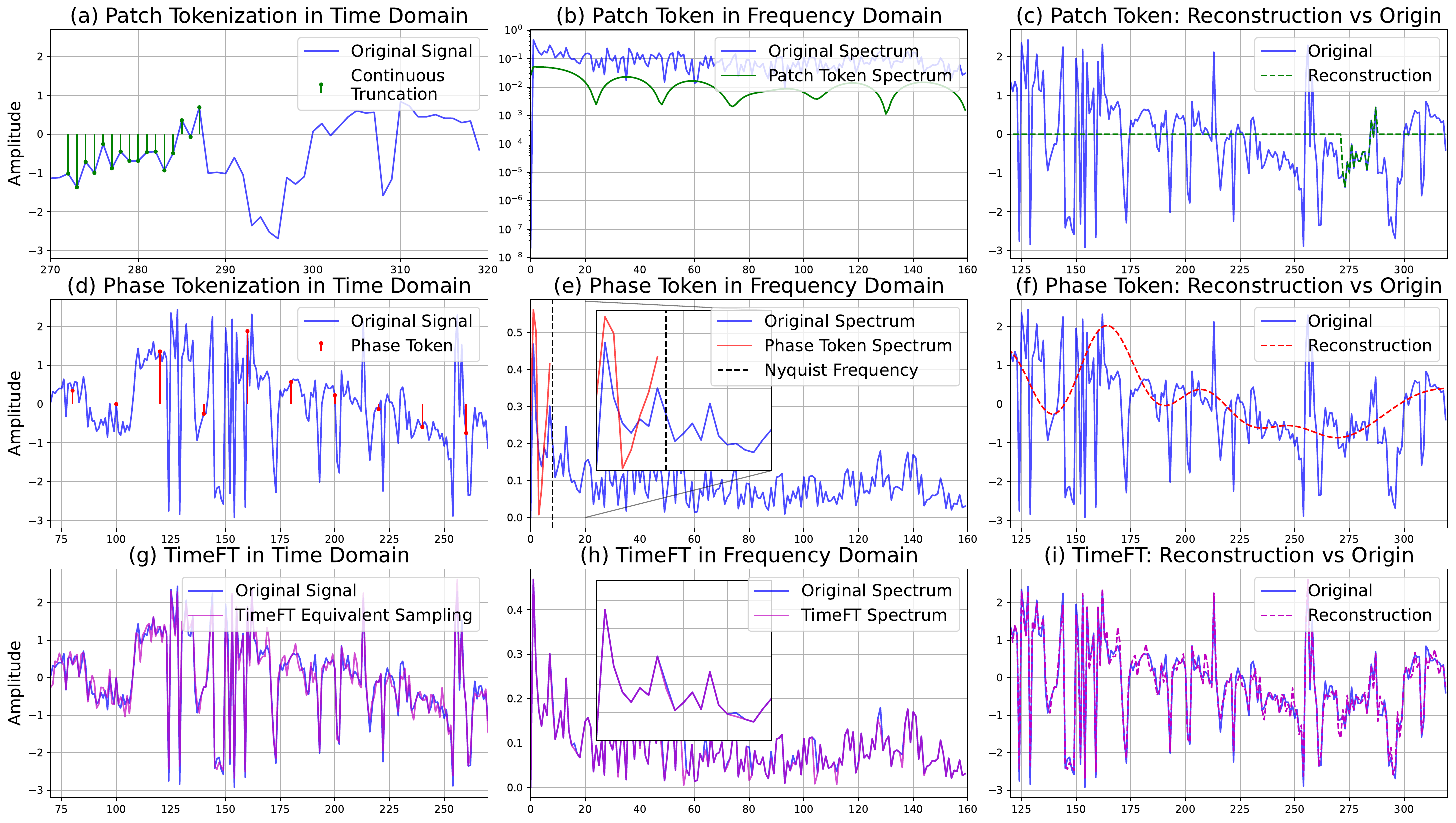} 
    \caption{The visualization of tokenization methods across the time domain, frequency domain and reconstruction vs origin in Weather dataset.}
    \label{fig: visualization_Weather}
    \vspace{-10pt}
\end{figure*}

\begin{figure*}[htbp]
    \centering
    \includegraphics[width=\textwidth]{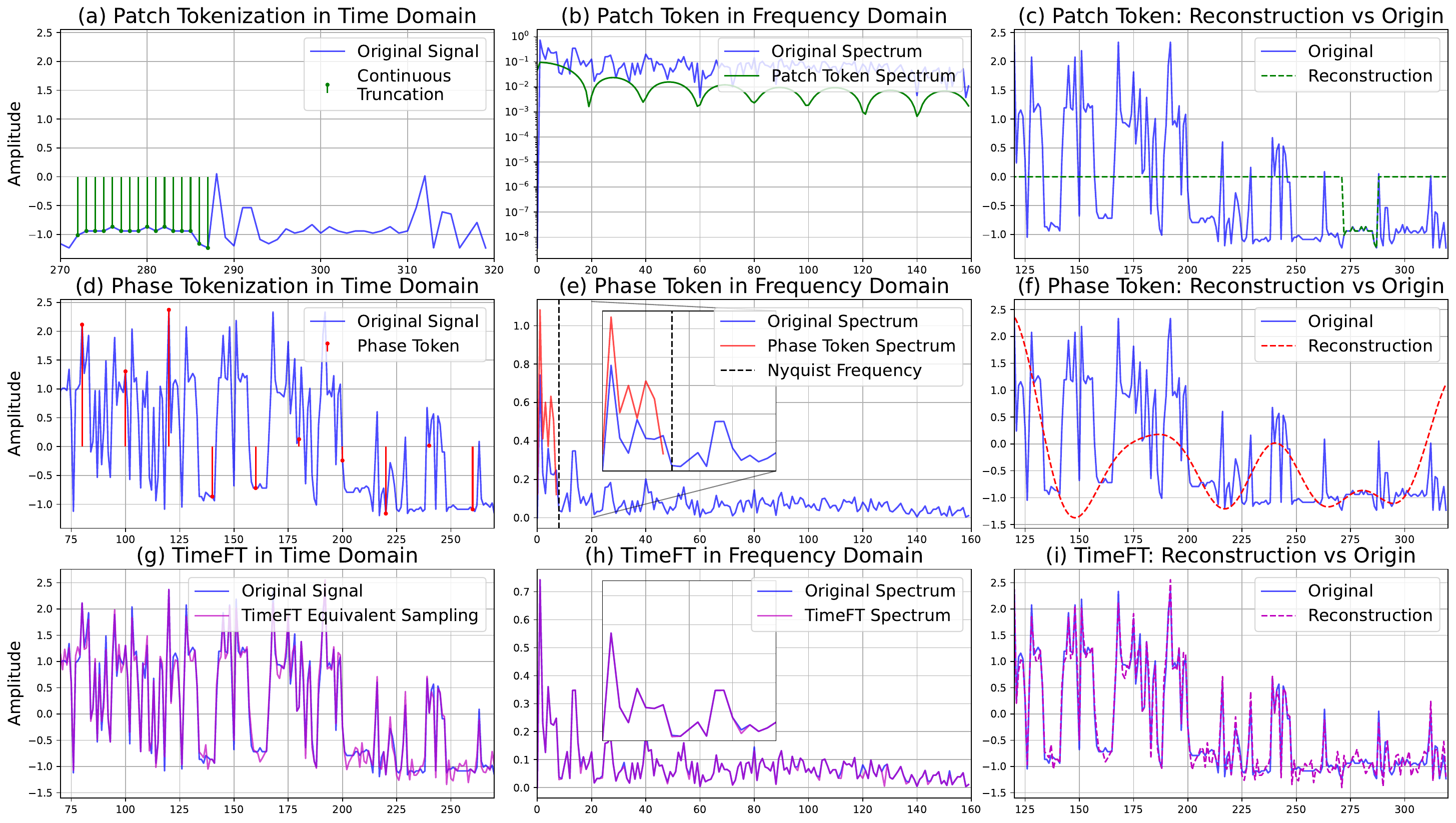} 
    \caption{The visualization of tokenization methods across the time domain, frequency domain and reconstruction vs origin in Electricity dataset.}
    \label{fig: visualization_Electricity}
    \vspace{-10pt}
\end{figure*}

\begin{figure*}[t]
    \centering
    \includegraphics[width=\textwidth]{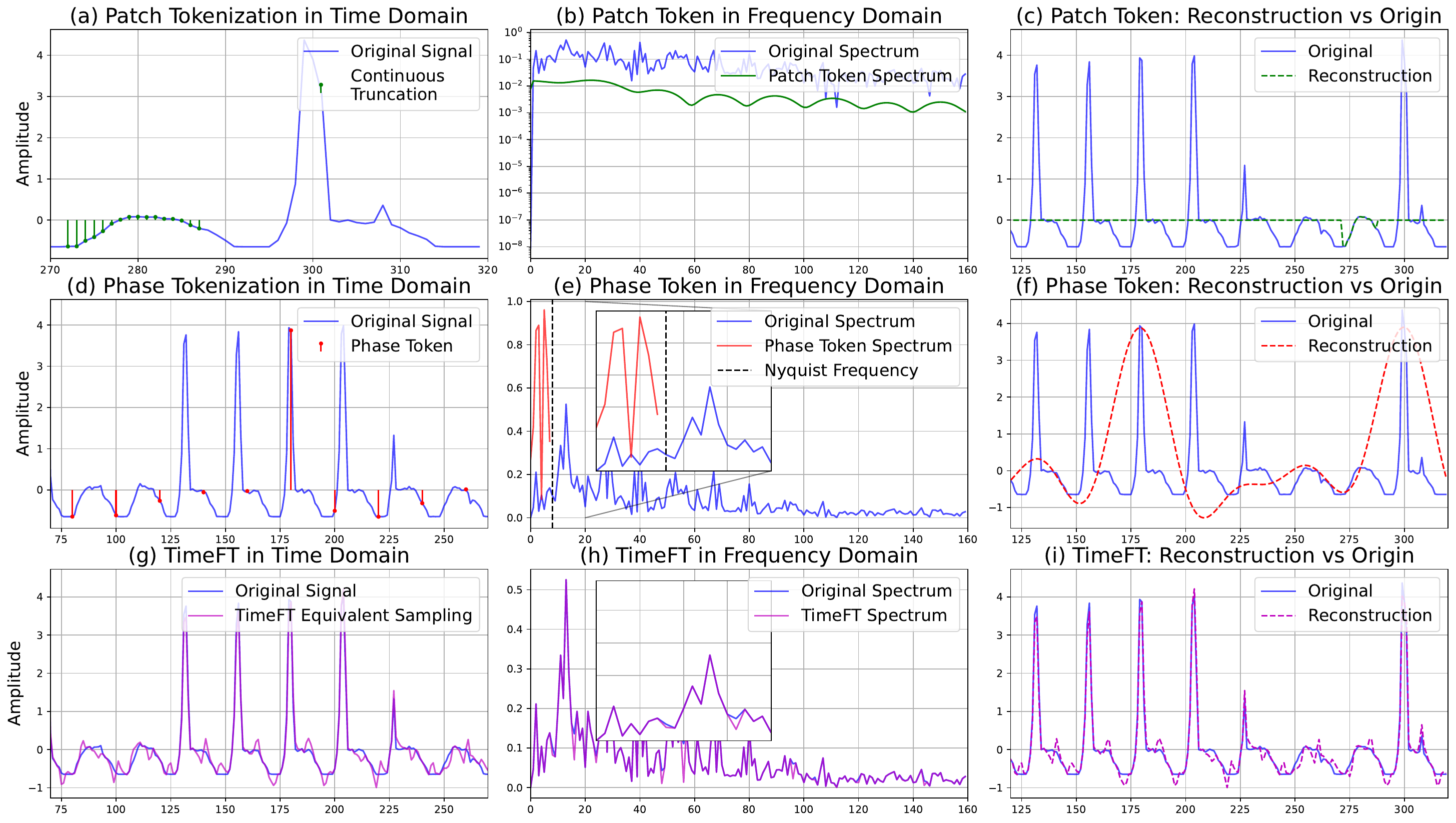} 
    \caption{The visualization of tokenization methods across the time domain, frequency domain and reconstruction vs origin in Traffic dataset.}
    \label{fig: visualization_traffic}
    \vspace{-10pt}
\end{figure*}

Since TimeFT is a tokenization method, we present visualizations comparing different classes of tokens in the time domain, frequency domain, and after equivalent reconstruction. As shown in \Cref{fig: visualization_ETTh1,fig: visualization_ETTh2,fig: visualization_ETTm1,fig: visualization_ETTm2,fig: visualization_Weather,fig: visualization_Electricity,fig: visualization_traffic}, patch tokens lead to (b) a completely different frequency spectrum and (c) cause the loss of all information outside the rectangular window. Phase tokens cause (e) frequency components above the Nyquist frequency to completely alias into the low frequency, thereby (f) exhibiting erroneous patterns. Our TimeFT (h) effectively solves the above issues and (i) achieves almost distortion‑free tokenization.

\newpage

\newpage

\newpage

\newpage
\ \ 

\section*{NeurIPS Paper Checklist}

The checklist is designed to encourage best practices for responsible machine learning research, addressing issues of reproducibility, transparency, research ethics, and societal impact. Do not remove the checklist: {\bf The papers not including the checklist will be desk rejected.} The checklist should follow the references and follow the (optional) supplemental material.  The checklist does NOT count towards the page
limit. 

Please read the checklist guidelines carefully for information on how to answer these questions. For each question in the checklist:
\begin{itemize}
    \item You should answer \answerYes{}, \answerNo{}, or \answerNA{}.
    \item \answerNA{} means either that the question is Not Applicable for that particular paper or the relevant information is Not Available.
    \item Please provide a short (1--2 sentence) justification right after your answer (even for \answerNA). 
\end{itemize}

{\bf The checklist answers are an integral part of your paper submission.} They are visible to the reviewers, area chairs, senior area chairs, and ethics reviewers. You will also be asked to include it (after eventual revisions) with the final version of your paper, and its final version will be published with the paper.

The reviewers of your paper will be asked to use the checklist as one of the factors in their evaluation. While \answerYes{} is generally preferable to \answerNo{}, it is perfectly acceptable to answer \answerNo{} provided a proper justification is given (e.g., error bars are not reported because it would be too computationally expensive'' or ``we were unable to find the license for the dataset we used''). In general, answering \answerNo{} or \answerNA{} is not grounds for rejection. While the questions are phrased in a binary way, we acknowledge that the true answer is often more nuanced, so please just use your best judgment and write a justification to elaborate. All supporting evidence can appear either in the main paper or the supplemental material, provided in appendix. If you answer \answerYes{} to a question, in the justification please point to the section(s) where related material for the question can be found.

IMPORTANT, please:
\begin{itemize}
    \item {\bf Delete this instruction block, but keep the section heading ``NeurIPS Paper Checklist"},
    \item  {\bf Keep the checklist subsection headings, questions/answers and guidelines below.}
    \item {\bf Do not modify the questions and only use the provided macros for your answers}.
\end{itemize}


\begin{enumerate}

\item {\bf Claims}
    \item[] Question: Do the main claims made in the abstract and introduction accurately reflect the paper's contributions and scope?
    \item[] Answer: \answerYes{} 
    \item[] Justification: The paper’s contributions and scope are well-explained throughout the abstract and introduction, and the entire paper is written to provide a consistent and comprehensive conclusion.

\item {\bf Limitations}
    \item[] Question: Does the paper discuss the limitations of the work performed by the authors?
    \item[] Answer: \answerYes{} 
    \item[] Justification: Our paper discusses the limitations in the Limitation section~\ref{app: limitations}.

\item {\bf Theory assumptions and proofs}
    \item[] Question: For each theoretical result, does the paper provide the full set of assumptions and a complete (and correct) proof?
    \item[] Answer: \answerYes{} 
    \item[] Justification: Our paper provides the full set of assumptions and a complete and correct proof in Section~\ref{app: Theory}.

    \item {\bf Experimental result reproducibility}
    \item[] Question: Does the paper fully disclose all the information needed to reproduce the main experimental results of the paper to the extent that it affects the main claims and/or conclusions of the paper (regardless of whether the code and data are provided or not)?
    \item[] Answer: \answerYes{} 
    \item[] Justification: Our paper provides detailed information to reproduce all experiments and the complete experiment code.

\item {\bf Open access to data and code}
    \item[] Question: Does the paper provide open access to the data and code, with sufficient instructions to faithfully reproduce the main experimental results, as described in supplemental material?
    \item[] Answer: \answerYes{} 
    \item[] Justification: Please see the code link in Abstract.

\item {\bf Experimental setting/details}
    \item[] Question: Does the paper specify all the training and test details (e.g., data splits, hyperparameters, how they were chosen, type of optimizer) necessary to understand the results?
    \item[] Answer: \answerYes{} 
    \item[] Justification: Our paper provides detailed information about the experiments, including data, hyperparameters, how they were chosen, and the type of optimizer, in the main text and the appendix.

\item {\bf Experiment statistical significance}
    \item[] Question: Does the paper report error bars suitably and correctly defined or other appropriate information about the statistical significance of the experiments?
    \item[] Answer: \answerYes{} 
    \item[] Justification: Experiments are repeated three times and the results are reported in Tables. We also evaluate the sensitivity of hyperparameter in~\S\ref{subsec: hyperparameter sensitivity analysis}.

\item {\bf Experiments compute resources}
    \item[] Question: For each experiment, does the paper provide sufficient information on the computer resources (type of compute workers, memory, time of execution) needed to reproduce the experiments?
    \item[] Answer: \answerYes{} 
    \item[] Justification: We reported the computer resources used in the experiments in Appendix.
    
\item {\bf Code of ethics}
    \item[] Question: Does the research conducted in the paper conform, in every respect, with the NeurIPS Code of Ethics \url{https://neurips.cc/public/EthicsGuidelines}?
    \item[] Answer: \answerYes{} 
    \item[] Justification: The research conducted in the paper fully conforms to the NeurIPS Code of Ethics in every respect.

\item {\bf Broader impacts}
    \item[] Question: Does the paper discuss both potential positive societal impacts and negative societal impacts of the work performed?
    \item[] Answer: \answerYes{}{} 
    \item[] Justification: Our paper discusses the potential broader impacts of our research in the Appendix~\ref{app: limitations}.
    
\item {\bf Safeguards}
    \item[] Question: Does the paper describe safeguards that have been put in place for responsible release of data or models that have a high risk for misuse (e.g., pre-trained language models, image generators, or scraped datasets)?
    \item[] Answer: \answerNA{} 
    \item[] Justification: Our model does not pose such risks.

\item {\bf Licenses for existing assets}
    \item[] Question: Are the creators or original owners of assets (e.g., code, data, models), used in the paper, properly credited and are the license and terms of use explicitly mentioned and properly respected?
    \item[] Answer: \answerYes{} 
    \item[] Justification: We have properly credited the creators or original owners of assets used in the paper.

\item {\bf New assets}
    \item[] Question: Are new assets introduced in the paper well documented and is the documentation provided alongside the assets?
    \item[] Answer: \answerNA{} 
    \item[] Justification: This paper does not include new assets.

\item {\bf Crowdsourcing and research with human subjects}
    \item[] Question: For crowdsourcing experiments and research with human subjects, does the paper include the full text of instructions given to participants and screenshots, if applicable, as well as details about compensation (if any)? 
    \item[] Answer: \answerNA{} 
    \item[] Justification: This paper does not involve crowdsourcing and research with human subjects.

\item {\bf Institutional review board (IRB) approvals or equivalent for research with human subjects}
    \item[] Question: Does the paper describe potential risks incurred by study participants, whether such risks were disclosed to the subjects, and whether Institutional Review Board (IRB) approvals (or an equivalent approval/review based on the requirements of your country or institution) were obtained?
    \item[] Answer: \answerNA{} 
    \item[] Justification: This paper does not involve research with human subjects.

\item {\bf Declaration of LLM usage}
    \item[] Question: Does the paper describe the usage of LLMs if it is an important, original, or non-standard component of the core methods in this research? Note that if the LLM is used only for writing, editing, or formatting purposes and does \emph{not} impact the core methodology, scientific rigor, or originality of the research, declaration is not required.
    \item[] Answer: \answerNA{} 
    \item[] Justification: LLM is not an important, original, or non-standard component of the core methods in this research.

\end{enumerate}

\end{document}